\documentclass{article}
\usepackage[margin=1in]{geometry}

\def\colorful{1}

\usepackage[utf8]{inputenc} %
\usepackage[T1]{fontenc}    %
\usepackage[colorlinks, backref]{hyperref}         %
\usepackage{url}            %
\usepackage{booktabs}       %
\usepackage{amsthm, amsmath, mathrsfs, mathtools}
\usepackage{amsfonts}       %
\usepackage{nicefrac}       %
\usepackage{microtype}      %
\usepackage{xcolor}         %

\usepackage{thm-restate}
\usepackage[algo2e,ruled,noend,resetcount,linesnumbered]{algorithm2e}
\usepackage{graphicx}
\usepackage{enumitem}
\usepackage{thmtools}
\usepackage{algorithm, algorithmicx, algpseudocode} 
\usepackage{cleveref}
\usepackage{xspace}
\usepackage{caption}
\usepackage{subcaption}

\usepackage{comment}
\newtheorem{fact}{Fact}
\usepackage[draft,margin=false,inline=true]{fixme}
\fxsetup{mode=multiuser,theme=color, layout=inline}

\setlist[enumerate]{nosep, topsep=1ex}
\setlist[itemize]{nosep, topsep=1ex}
\setlist[description]{nosep}

\usepackage{dsfont}
\usepackage{algpseudocode}
\usepackage{xcolor}
\usepackage{comment}

\ifnum\colorful=0
\newcommand{\todo}[1]{\textcolor{red}{TODO: #1}}
\newcommand{\chris}[1]{\footnote{\textcolor{blue}{[CY: #1]}}}

\newcommand{\snew}[1]{\textcolor{orange}{#1}}
\newcommand{\snote}[1]{\footnote{\textcolor{orange}{\bf [[Sihan: {#1}\bf ]] }}}
\newcommand{\jnote}[1]{\footnote{\textcolor{olive}{\bf [[Jingyi: {#1}\bf ]] }}}
\newcommand{\inote}[1]{\footnote{{\bf [[Ilias: {#1}\bf ]] }}}
\newcommand{\jnew}[1]{\textcolor{cyan}{#1}}
\else
\newcommand{\todo}[1]{}
\newcommand{\chris}[1]{}
\newcommand{\sihan}[1]{}

\newcommand{\snew}[1]{#1}
\newcommand{\jnew}[1]{#1}

\newcommand{\jnote}[1]{}
\newcommand{\inote}[1]{}
\newcommand{\snote}[1]{}
\fi

\newlist{encase}{enumerate}{1}
\setlist[encase]{label=Case \arabic*:, leftmargin=0.1cm,align=left}

\newcommand{\innerAlg}{\mathcal{A}}

\newcommand{\p}{\mathbf p}
\newcommand{\q}{\mathbf q}
\algrenewcommand\algorithmicrequire{\textbf{Input:}}
\algrenewcommand\algorithmicensure{\textbf{Output:}}

\newcommand{\bigO}[1]{O \left( #1 \right)}
\newcommand{\litO}[1]{o \left( #1 \right)}

\newcommand{\poly}[1]{{\rm poly} \left( #1 \right)}

\newcommand{\E}[1]{\mathbb{E}\left[#1\right]}
\newcommand{\Ep}{\mathbb{E}}
\newcommand{\Var}{\mathrm{Var}}

\newcommand{\dtv}{d_{\text{TV}}}

\newcommand{\PoiD}[1]{\mathrm{Poi} \left(#1\right)}

\newcommand{\iid}{{i.i.d.}\ }

\newcommand{\tvd}[1]{\dtv \left(#1\right)}

\newtheorem{theorem}{Theorem}[section]
\newtheorem{corollary}[theorem]{Corollary}
\newtheorem{claim}[theorem]{Claim}
\newtheorem{definition}[theorem]{Definition}
\newtheorem{lemma}[theorem]{Lemma}
\newtheorem{proposition}[theorem]{Proposition}

\newcommand{\N}{\mathbb{N}}
\newcommand{\Z}{\mathbb{Z}}
\newcommand{\R}{\mathbb{R}}

\newcommand{\distribution}{\mathbf p}

\newcommand{\eps}{\varepsilon}
\newcommand{\polylog}{\mathrm{polylog}}

\newcommand{\given}{\textrm{\xspace s.t.\ \xspace}}

\newcommand{\set}[1]{\{#1\}}
\newcommand{\bigset}[1]{\left\{#1\right\}}

\newcommand{\norm}[2]{\left|\left|#1\right|\right|_{#2}}

\newcommand{\lp}{\left}
\newcommand{\rp}{\right}

\newcommand{\KL}{\text{KL}}

\newcommand{\0}{\mathbf{0}}

\newcommand{\RW}{\mathbf{RW}}

\newcommand{\rel}{{\rm rel}}

\newcommand{\calM}{\mathcal{M}}
\newcommand{\calH}{\mathcal{H}}
\newcommand{\calN}{\mathcal{N}}
\newcommand{\calT}{\mathcal{T}}
\newcommand{\calB}{\mathcal{B}}
\setlist[encase]{label=Case \arabic*:, leftmargin=2cm}

\newcommand{\Poi}{\text{Poi}}
\newcommand{\PoiS}{\text{PoiS}}

\newcommand{\pq}[1]{(\p_{#1}, \q_{#1})}

\newcommand{\Acc}{\text{Acc}}
\newcommand{\Accept}{\text{Accept}}

\newcommand{\indcomplexity}{
n_1^{2/3}n_2^{1/3} 
\rho^{-2/3} 
\eps^{-4/3}
+ \sqrt{n_1 n_2} \rho^{-1} \eps^{-2}
+ \rho^{-2} \eps^{-2}
}

\newcommand{\zexpgap}{{\min\lp(\eps m, m^2 \eps^2 / (n_1 n_2), m^{3/2} \eps^2 / \sqrt{n_1 n_2} \rp)}}

\newcommand{ \B }{\text{Bernoulli}}

\title{Replicable Distribution Testing}

\author{%
Ilias Diakonikolas\thanks{Supported by NSF Medium Award CCF-2107079 and an H.I. Romnes Faculty Fellowship.}\\
University of Wisconsin-Madison\\
{\tt ilias@cs.wisc.edu}
\and
Jingyi Gao\\
University of Wisconsin-Madison\\
{\tt jingyig@cs.wisc.edu}
\and
Daniel M. Kane\thanks{Supported by NSF Medium Award CCF-2107547 and NSF Award CCF-1553288 (CAREER).}\\
University of California, San Diego\\
{\tt dakane@cs.ucsd.edu}
\and
  Sihan Liu\thanks{Supported by NSF Medium Award CCF-2107547 and NSF Award CCF-1553288 (CAREER).}\\
  University of California, San Diego\\
  La Jolla, CA 
  \\
  \texttt{sil046@ucsd.edu}
  \and
  Christopher Ye\thanks{Supported by NSF Award AF: Medium 2212136, NSF grants 1652303, 1909046, 2112533, and HDR TRIPODS Phase II grant 2217058.}\\
  University of California, San Diego\\
  La Jolla, CA 
  \\
  \texttt{czye@ucsd.edu} \\
}

\begin{document}

\maketitle

\setcounter{page}{0}
\thispagestyle{empty}

\begin{abstract}
We initiate a systematic investigation of distribution testing in 
the framework of algorithmic replicability. Specifically, given 
independent samples from a collection of probability distributions, the goal is to characterize 
the sample complexity of replicably testing 
natural properties of the underlying distributions. On the algorithmic front, we develop new replicable algorithms 
for testing closeness and independence of discrete distributions. 
On the lower bound front, we develop a new methodology  
for proving sample complexity lower bounds for 
replicable testing that may be of broader interest. As an application of our technique, 
we establish near-optimal sample complexity lower bounds for replicable uniformity testing---answering an open question from prior work---and closeness testing.
\end{abstract}

\newpage 

\section{Introduction}

Algorithmic replicability has emerged as a fundamental notion in modern statistics and machine learning to ensure consistency of algorithm outputs in the presence of randomness in input datasets. 
The formal notion of replicability, proposed in \cite{impagliazzo2022reproducibility}, is as follows.
\begin{definition}[Replicability \cite{impagliazzo2022reproducibility}]
    \label{def:replicability}
    A randomized algorithm $\innerAlg: \mathcal X^n \mapsto \mathcal Y$ is $\rho$-replicable if for all distributions $\distribution$ on $\mathcal X$,
    $
        \Pr_{r, T, T'} \left( \innerAlg(T; r) = \innerAlg(T'; r) \right) \geq 1 - \rho,
    $
    where $T, T'$ are \iid samples drawn from $\distribution$, and
    $r$ denotes the internal randomness of the algorithm $\innerAlg$.
\end{definition}
Since its introduction, replicability has been studied in the context of a wide range of machine learning tasks, including 
multi-arm bandits \cite{esfandiari2022replicable},
clustering \cite{esfandiari2023replicable}, 
reinforcement learning \cite{karbasi2023replicability,eaton2024replicable},
halfspace learning \cite{kalavasis2024replicable}, and high-dimension statistics \cite{hopkins2024replicability}.
A related line of work explored the connection between replicability and other algorithmic stability notions 
such as differential privacy \cite{bun2023stability,MSS23}, total variation indistinguishability \cite{DBLP:conf/icml/KalavasisKMV23},
global stability \cite{CCMY23}, and one-way perfect generalization~\cite{bun2023stability,MSS23}.

In this work, we initiate a systematic study of replicability in distribution testing, a central area in property testing and statistics that aims to ascertain whether an unknown distribution satisfies a certain property or is ``far'' from satisfying that property.
Specifically, we focus on replicable testing of discrete distributions in total variation distance, which encompasses canonical problems such as uniformity/identity testing, closeness testing, and independence testing. Formally, we have:
\begin{definition}[$(\epsilon,\rho)$-replicable testing of property $\mathcal{P}$]
\label{def:nepsrho-rep-testing}
Let $\mathcal{P}$ be a property consisting of $k$-tuples of distributions, and $\epsilon,\rho \in (0,1/4)$. 
Given sample access to a collection of distributions $\p^{(1)}, \cdots, \p^{(k)}$, 
we say a randomized algorithm $\innerAlg$ solves $(\epsilon,\rho)$-replicable $\mathcal{P}$-testing if $\innerAlg$ is $\rho$-replicable and can distinguish between the following cases with probability at least $2/3$ \footnote{
\snew{Even though here we only require the success probability to be $2/3$ (as is standard), 
one can see that any $\rho$-replicable algorithm which is correct with probability $2/3$ should also be correct with probability at least $1 - \rho$.}
}:
completeness case
$(\p^{(1)}, \cdots, \p^{(k)})\in \mathcal{P}$ or
soundness case
$ k^{-1} \; \sum_{i=1}^k \dtv\lp( \p^{(i)},  \q^{(i)} \rp) \geq \eps$ for all tuples $(\q^{(1)}, \cdots, \q^{(k)}) \in \mathcal P$. 
In particular, (a) uniformity testing over the  
domain $[n]$ corresponds to the case that $k=1$, and 
$\mathcal P$ consists of only the uniform distribution over 
$[n]$; (b) closeness testing over the domain $[n]$
corresponds to the case that $k=2$, 
and $\mathcal P$ consists of all pairs of distributions 
$(\p, \p)$ over $[n]$; 
(c) independence testing over 
domain $[n_1] \times [n_2]$ corresponds 
to the case that $k=1$, 
and $\mathcal P$ consists of all product distributions 
over that domain.
\end{definition}

 After the pioneering works formulating this 
 field~\cite{batu2000testing,goldreich2011testing} 
 from a TCS perspective, 
a line of work has given efficient testers (without 
the replicability requirement) achieving 
information-theoretically optimal sample complexities 
for the aforementioned problems; see \cite{paninski2008coincidence,DiakonikolasKN15, valiant2017automatic} 
for uniformity/identity testing, 
\cite{batu2000testing,valiant2008testing,batu2013testing,chan2014optimal}
for closeness testing, and \cite{batu2001testing,acharya2015optimal,diakonikolas2016new} for independence testing.
More broadly, 
substantial progress has been made on testing a wide range of natural properties; 
see, e.g.,~\cite{BDKR:02, BKR:04,
DJOP11, LRR11, diakonikolas2015optimal, CDKS17, DiakonikolasKN17, DaskalakisDK18, CanonneDKS18, CanonneJKL22, CDKL22,DiakonikolasKL24}
for a sample of works, and~\cite{rubinfeld2012taming,canonne2020survey,canonne2022topics} for surveys on the topic.

Despite the maturity of the field of distribution testing, the relevant literature contains only a single prior work addressing replicability: a recent paper by \cite{liu2024replicable} designs replicable testers for the task of uniformity testing (and identity testing via a standard reduction technique), and demonstrates that the sample complexity is nearly tight within a {\em restricted} class of algorithms. Specifically,  
their techniques are insufficient to establish lower bounds against \emph{general} uniformity testers, or to design sample-optimal replicable testers for other distribution testing problems.

An instructive parallel arises in the context of differentially private (DP) distribution testing, where similar challenges in designing testers with additional stability constraints 
have been addressed; see \cite{cai2017priv,aliakbarpour2018differentially,acharya2018differentially,aliakbarpour2019private}. 
Although generic reductions from replicable to DP algorithms 
exist, see, e.g.,~\cite{bun2023stability}, they 
incur polynomial overheads in sample complexity. 
This motivates our goal: to develop a principled and fine-grained 
understanding of the sample complexity cost of
replicability in distribution testing.

\subsection{Our Results}

Our first main contribution is a new lower bound framework 
for replicable distribution testing 
that yields unconditional lower bounds against all testers, without 
additional assumptions.
As a first application of our framework, we show that the replicable 
uniformity tester proposed in \cite{liu2024replicable} is indeed 
nearly optimal, thus settling the main open problem posed in their work.

\begin{theorem}[Sample Complexity of Replicable Uniformity Testing]
\label{thm:unif-test-lower-bound}
The sample complexity of $(\eps, \rho)$-replicable uniformity testing over $[n]$ is
$\tilde \Theta(\sqrt{n} \eps^{-2} \rho^{-1}+ \eps^{-2} \rho^{-2} )$.
\end{theorem}
We believe that the novel lower bounds framework that we develop 
is broadly applicable to establishing lower bounds for other replicable distribution testing problems. 
In particular, as an additional application of our framework, 
we derive a tight lower bound 
for replicable closeness testing. 

\begin{theorem}[Lower Bound of Replicable Closeness Testing]
    \label{thm:closensess-lb}
    The sample complexity of $(\eps, \rho)$-replicable closeness testing over $[n]$ is at least $\tilde \Omega(
    n^{2/3} \eps^{-4/3} \rho^{-2/3} + \sqrt{n} \eps^{-2} \rho^{-1} + \eps^{-2} \rho^{-2} )$.
\end{theorem}

On the algorithmic front, 
we provide new replicable testers for closeness and independence 
testing. For closeness testing, we give a tester 
whose sample complexity nearly matches 
our aforementioned lower bound: 
\begin{theorem}[Replicable Closeness Tester]
\label{thm:closeness}
    The sample complexity of $(\eps, \rho)$-replicable closeness testing over $[n]$ is at most $O(
    n^{2/3} \eps^{-4/3} \rho^{-2/3} + \sqrt{n} \eps^{-2} \rho^{-1} + \eps^{-2} \rho^{-2} )$.
\end{theorem}

\Cref{thm:closeness}, together with \Cref{thm:closensess-lb}, 
give a tight characterization of the sample complexity of replicable closeness testing up to polylogarithmic factors. 

For independence testing, we show that:
\begin{theorem}[Replicable Independence Tester]
\label{thm:independence}
The sample complexity of $(\eps, \rho)$-replicable independence testing over $[n_1] \times [n_2]$ for $n_1\ge n_2$ is at most $\tilde O\lp(\frac{n_1^{2/3}n_2^{1/3}}{\rho^{2/3}\eps^{4/3}} + \frac{\sqrt{n_1 n_2}}{\rho\eps^2} + \frac{1}{\eps^2\rho^2}  \rp)$. 
\end{theorem}

We note that both our testers are computationally efficient. 
While our replicable closeness tester runs in sample 
linear time, our replicable independence tester takes 
sample-polynomial time. This is due to an extra ``averaging'' 
operation applied to make the statistic more stable (see 
\Cref{ssec:techniques}). We leave it for future work to 
explore whether its runtime can be further improved.

Perhaps surprisingly, our upper bounds 
point to an intriguing conceptual connection 
between replicability and distribution testing 
in the \emph{high success probability regime}. 
In particular, the functional forms of the sample complexities of 
(non-replicable) uniformity, 
closeness, and independence testing up to error probability $\delta$ 
have been characterized in \cite{diakonikolas2018sample,diakonikolas2021optimal} to be 
precisely the sample complexity upper bounds 
of the corresponding replicable testing problems 
after replacing $\rho^{-1}$ with $\sqrt{ \log(1/\delta)}$~\footnote{For example, 
the sample complexity of high probability closeness testing has been shown to be $\Theta \lp(  n^{2/3} \eps^{-4/3}  \log^{1/3}(1/\delta) 
+ \sqrt{n} \eps^{-2} \sqrt{\log(1/\delta)}
+ \eps^{-2} \log(1/\delta)
\rp)$ in \cite{diakonikolas2021optimal}.}.
Moreover, all known replicable distribution testers (including ours) 
leverage the statistics developed in the context 
of high probability distribution testers.
We leave it as an interesting open problem whether 
there exists a generic reduction from high-probability testers 
to replicable ones or vice versa.
With this connection in mind, it is a plausible conjecture that our sample complexity 
upper bound for replicable independence testing 
is nearly optimal. While we believe that our lower bounds framework can be used to establish a nearly tight sample complexity lower bound for replicable independence testing, we leave this as an open question for future work.

\subsection{Technical Overview} \label{ssec:techniques}

{We start with a description of our lower bound framework, which 
is the main technical contribution of this work, 
followed by our upper bounds.} 

\paragraph{Replicable Testing Lower Bounds}
 In what follows, we will sketch the overall framework for showing 
 lower bounds against replicable uniformity and closeness testing.  
 For concreteness, we will point to the specific lemmas we 
 establish in deriving our uniformity testing lower bound.
Let $\innerAlg$ be a randomized algorithm for the testing problem 
that uses significantly fewer samples than the target lower bound. 
Our end goal is to show that if $\innerAlg$ satisfies 
the correctness requirements of the corresponding testing problem, 
then $\innerAlg$ cannot be replicable.
Towards this goal, we begin with the same reasoning steps as the ones in \cite{liu2024replicable}.
In particular, we construct a meta-distribution $\mathcal M_\xi$, 
parametrized by $\xi \in [0, \eps]$, over potential hard instances of the 
testing problem such that (i) $\mathcal M_0$ and $\mathcal M_\eps$ 
correspond to instances that should be respectively accepted and rejected by 
the tester, and (ii) it should be hard to distinguish a random instance from 
$\mathcal M_{\xi}$ versus a random instance from $\mathcal M_{\xi + \eps \rho}$.

After that, using the same argument as \cite{liu2024replicable}, 
we can deduce that if we sample $\xi \sim \mathcal U([0, \eps])$, 
then the average acceptance probability of the tester 
under $\mathcal M_\xi$, i.e., 
$ \Ep_{  \p \sim \mathcal M_\xi } \lp[ 
\Pr_{ S \sim \p }\lp[  \innerAlg \text{ accepts } S \rp] \rp] $, 
will be close to $1/2$ with probability at least $\Omega(\rho)$ 
over the randomness of $\xi$. 
See \Cref{lem:avg-acceptance-prob-unif} for the formal statement.
If $\mathcal M_\xi$ were to contain just a single distribution 
instance $\p_\xi$, then the statement would directly imply 
that $\innerAlg$ is not replicable under $\p_\xi$, 
and this would conclude the lower bound argument.
Of course, $\mathcal M_\xi$ is in reality 
a meta-distribution over (exponentially) many different 
instances by design (see \Cref{def:uniform-hard-instance}).
To overcome this issue, \cite{liu2024replicable} takes 
advantage of the fact that the instances from the meta-distribution are identical 
up to permutation of the domain elements. 
As such, if one makes the {\em additional assumption} 
that the output of the tester is invariant 
up to domain relabeling 
(in other words, the tester is \emph{symmetric}), 
then it is not hard to show that the acceptance probability 
of the tester under each individual distribution 
must be the same as the overall averaged acceptance probability 
under $\mathcal M_\xi$, and the proof is complete.

Our proof circumvents this difficulty  
with a fundamentally different approach 
that allows us to avoid making {\em any} assumptions on the tester.
As one of our main technical contributions, 
we show that even when the tester is {\em not} symmetric, 
the acceptance probability under a random choice of 
$\p \sim \mathcal M_\xi$ must nonetheless concentrate 
around its expectation, as long as the tester 
is still moderately replicable under $\p$, 
i.e., replicable with probability $1 - 1 / \polylog(n)$.
See \Cref{lemma:uniformity-acceptance-concentration} 
for the formal statement.
Towards this goal, consider the joint distribution 
of two random sample sets $S, S'$ generated as follows: 
pick a random distribution $\p \sim \mathcal M_\xi$ 
and then sample $S, S'$ independently from $\mathcal M_\xi$.
The distribution of $S'$ conditioned on $S$ 
then naturally defines a random walk $\RW_\xi$ 
on the space of all possible sample sets.
For convenience, denote by $\RW_\xi^k(\p)$ the distribution 
over sample sets $\bar S$ obtained by first sampling 
$T \sim \p$ and then performing $k$ steps of the random walk.
Lying at the heart of our proof is the following 
two structural claims:
(1) for most $\p \sim \mathcal M_\xi$,
the acceptance probability of the tester 
under $\bar S \sim \RW_\xi^{ \polylog(n) }(\p)$ 
is roughly the same as that under $S \sim \p$. 
(2) the random walk $\RW_\xi$ has mixing time at most $\polylog(n)$.
Combining these two claims gives that the acceptance probability 
under most $\p \sim \mathcal M_\xi$ must be roughly the same, 
as the acceptance probability under the stationary distribution 
of the random walk (which by construction is exactly equal to the expected 
acceptance probability under $\mathcal M_\xi$).

It thus remains for us to establish these two claims.
The proof of (1) mainly follows from the definition 
of the random walk, and the assumption that the tester 
is moderately replicable. See \Cref{lem:random-walk-indistinguishability} and its proof for details.
The canonical way for showing (2) is to bound from below 
the eigenvalue gaps of the transition matrix of the random walk.
To analyze this, we note that after a careful use 
of Poissonization (see \Cref{def:poi_samp} 
for the definition of Poisson sampling), 
we can make $\RW_\xi$ a product of $n$ independent random walks.
Formally, since the number of samples is Poissonized, 
the sample frequency of each bucket is independent, 
even conditioned on the choice of distribution $\p \sim \calM_{\xi}$.
It then suffices for us to bound the mixing time 
of much simpler random walks on the sample frequencies of each individual domain element.
Fortunately, the eigenvalue gap of this random walk can be analyzed 
conveniently using elementary properties of the Poisson distributions. 
See \Cref{thm:unif-rw-mixing-delta} and its proof for details.
The formal proofs of \Cref{thm:unif-test-lower-bound} 
and the relevant lemmas are in \Cref{sec:uniformity-lb}.

The same framework also applies to the task of closeness testing, leading to the proof of 
\Cref{thm:closeness}. The main change needed is to 
replace the meta-distribution $\mathcal M_\xi$ 
to be the standard hard instance for closeness testing. See \cite{chan2014optimal, diakonikolas2016new} 
for the construction of the hard instance. 
The formal proof can be found in \Cref{sec:clossness_lb}.

\paragraph{Replicable Testing Upper Bounds}
We begin with the observation that many testers 
from the literature share the following 
nice form: compute a test statistic $Z$ 
and compare it to a threshold $R$. 
Usually, the analysis (without replicability requirements) 
involves showing that $\E{Z} = 0$ in the completeness case, 
$\E{Z} \gg R$ in the soundness case,
and $\Var[Z]$ is at most a small constant multiple of $R^2$.
For testers of this form, 
we can employ the same strategy 
as the one used in \cite{liu2024replicable} 
to transform them into replicable testers:
we can compute the same test statistic $Z$, 
and then compare it to a {\em randomly chosen} threshold $r$ 
between $0$ and $R/2$.
In particular, if we further have that 
$\Var[Z] \ll  R^2 \rho^2 $ (at the cost of taking more samples), 
the variance bound on $Z$ implies that $Z$ computed 
with different sample sets drawn 
from the same underlying distribution 
are likely to be close to each other.
Consequently, the values of $Z$ in two runs are unlikely 
to be separated by a randomly chosen $r$, ensuring replicability.

For closeness testing,  the high probability tester from~\cite{diakonikolas2021optimal}
satisfies exactly the conditions needed, 
and in turn yields our replicable tester 
after combining it with the random thresholding strategy.
The formal proof of \Cref{thm:closeness} 
is given in \Cref{app:closeness-ub}.

Designing good replicable testers for independence 
turns out to be significantly more involved for the following reason: 
known high probability independence testers \cite{diakonikolas2021optimal} 
do not satisfy the required variance bounds 
within our sample complexity budget.
In its essence, the bottleneck lies in an extra 
randomized ``flattening'' procedure employed 
by the tester, which significantly increases 
the overall variance of the final statistic computed.
Specifically, the procedure utilizes a random subset 
of the input samples to ``split'' domain elements 
with large mass into sub-elements. This step aims to 
ensure that there will be no extremely heavy elements 
after the procedure (otherwise, the tester may fail 
to satisfy even the basic correctness requirements).
To show correctness of their tester,~\cite{diakonikolas2021optimal}  
demonstrated that (1) the flattening procedure preserves 
the product/non-product structure of the original distribution, 
and (2) the variance of the final test statistic
\emph{conditioned on} the flattening samples 
(and some other technical conditions) is small.
Notably, a bound on the total variance of the test statistic $Z$ 
is not needed in their context, 
as the above two properties suffice for them to 
show upper/lower bounds on $Z$ in the completeness/soundness cases.
Yet, to additionally guarantee replicability, 
we do need to show that $Z$ concentrates around a small interval.
As a result, the lack of a good bound on the total variance 
(as compared to just conditional variance) 
of the final test statistic turns out to be
a major technical obstacle 
in converting the high probability tester into a replicable one.
In fact, the randomness in using different samples 
for flattening purposes can easily cause 
the total variance of $Z$ to be much larger 
than the conditional variance.

To overcome this difficulty, 
{we leverage the following idea from 
\cite{aliakbarpour2019private} (in the context of differentially private testing): 
to make a test statistic computed with internal randomness more stable, 
we can replace it 
with the \emph{averaged} version of it.
In particular, we apply this idea 
to the statistic $Z$ computed by the high probability independence statistic, and obtain a new averaged statistic $Z_a$---essentially, 
the expected value of $Z$ averaged over all possible partitions 
of the input samples into flattening samples 
and testing samples (see \Cref{def:average-Z} for the formal definition).}
As our main technical lemma, we show that 
the total variance of this averaged test statistic $Z_a$ 
can be bounded from above by the expected value 
of $N$---the number of \emph{non-singleton} samples, 
i.e., the testing samples which still collide with another testing sample after the flattening procedure (see \Cref{lem:IS-var-bound}).
At a high level, our argument uses an 
Efron-Stein style inequality that bounds 
the variance by the sum of the expected 
square differences of the test statistic $Z_a$ 
caused by removal of each individual sample.
Suppose that there are in total $m$ samples 
and the probability of selecting a sample for flattening is $p$.
We then proceed by a case analysis.
If the sample removed is used for computing 
the final test statistic, we show that the (non-averaged) 
test statistic $Z$ will only be different 
if the sample also happens to be a singleton sample 
after flattening, which happens with probability roughly $O(N/m)$.
If the sample is selected for flattening, 
we show that removing it can change the test statistic $Z$ by 
at most $N$ divided by the number of flattening samples, 
which is roughly $O(p m)$.
Consequently, the contribution to the variance of $Z_a$ 
in this case is at most 
$\lp( p \; N / (mp) \rp)^2 \leq N^2 / m^2$, 
which is also $O(N / m)$.

It remains for us to control the non-singleton sample count $N$.
Fortunately, \cite{diakonikolas2021optimal} 
already established sharp bounds
on the expected value of $N$, 
when $\p$ is known to be a product distribution.
This then motivates us to run a pre-test 
to check whether $\Ep[N]$ is within a constant factor 
of the desired bound, 
before computing the averaged independence statistic $Z_a$.
In particular, we consider the statistic $N_a$, 
defined similarly to $Z_a$ as the expected 
non-singleton sample count $N$ averaged over the random choice 
of the flattening sample set, and use an almost identical argument
to show that $\Var[N_a]$ can also be bounded by 
$O(\E{N})$ (see \Cref{lem:N-variance-bound}). 
Equipped with the variance bound, it is not hard to show 
that comparing $N_a$ with an appropriately chosen 
random threshold yields a tester that replicably 
determines whether the magnitude of $\Ep[N]$ 
is within a constant factor 
of the bound it should satisfy when $\p$ is a product distribution.
If we pass this test, we can then proceed to apply 
the main test, which compares $Z_a$ to a randomized threshold. 
This concludes our proof sketch.
The relevant lemma statements and proofs can be found 
in \Cref{sec:independence-ub-body}.

\section{Preliminaries}
\label{sec:prelims}
\subsection{Notation}
Let $[n] = \set{1, \dotsc, n}$.
We use $n$ to denote the domain size and $m$ to denote the sample complexity of our testers.
We use bold letters (e.g., $\p, \q$) to denote distributions or measures and $\p(i)$ to denote the mass of $i$ under $\p$.
Let $\Poi(\lambda)$ denote a Poisson distribution with parameter $\lambda$ and $\PoiS(m, \p)$ denote $m' \sim \Poi(m)$ \iid samples from $\p$.
Let $\B(\alpha)$ denote a Bernoulli distribution with parameter $\alpha$. Let $\mathcal U(S)$ denote the uniform distribution over set $S$, where $S$ can be either a discrete set of points or an interval. Let Multinom$(m,(p_1,\cdots,p_k))$ denote the multinomial distribution with parameter $m$ being the number of trials and $p_1,\cdots,p_k$ be the event probabilities where $\sum_{i=1}^k p_i=1.$
For any distribution $\p$, let $\p^{\otimes m}$ denote $m$ \iid samples from $\p$. We use the terms ``algorithm'' and ``tester'' interchangeably. {For a multiset $S$ of samples, we denote the set of all elements appearing in $S$ by $supp(S)$.}
\subsection{Probability and Information Theory}
In this subsection, we present basic lemmas from probability and 
information theory that will be useful in our technical analysis.

\begin{lemma}[Poisson Concentration (see, e.g., \cite{CCPoisson}]
    \label{lemma:poisson-concentration}
    Let $X \sim \PoiD{\lambda}$.
    Then for any $x > 0$,
    \begin{equation*}
        \max(\Pr(X \geq \lambda + x), \Pr(X \leq \lambda - x)) < e^{-\frac{x^2}{2(\lambda + x)}} \text{.}
    \end{equation*}
\end{lemma}

    \begin{claim}[Asymptotic Upper Bound of Mutual Information]\label{MI_asymp}
    Let $X$ be an unbiased uniform random bit, and $M$ be a discrete random variable such that $\Pr[M=a|X=0] = \Theta(1)\Pr[M=a|X=1].$ 
    Then the mutual information between them satisfies 
    $$I(X:M) = O(1) \; \sum_{a}  \frac{(\Pr[M=a|X=0] - \Pr[M=a|X=1])^2}{\Pr[M=a|X=0] + \Pr[M=a|X=1]}.$$ 
    \end{claim}
    See \Cref{app:MI_asymp} for the proof of \Cref{MI_asymp}.

\subsection{Random Walks}

Let $\RW$ denote a random walk with transition matrix $P$, 
where $P(x, y)$ denotes the probability of transitioning 
to state $y$ from state $x$.
Let $P^t(x, y)$ denote the probability that starting from state $x$, the random walk $\RW$ is at state $y$ in $t$ steps.
We require the following basic definitions. 

\begin{definition}
    \label{def:irreducible-rw}
    A random walk $\RW$ is irreducible if for all $x, y$, there exists $t > 0$ such that $P^t(x, y) > 0$.
\end{definition}

\begin{definition}
    \label{def:periodic-rw}
    For a given state $x$, let $\calT(x) := \set{t > 0 \given P^t(x, x) > 0}$.
    The period of $x$ is the greatest common divisor of $\calT(x)$.
    For an irreducible random walk,
    the period of the random walk is the period 
    of every state (note this is in fact well-defined).
    A random walk is aperiodic if its period is $1$.
\end{definition}

A random walk that is both irreducible and aperiodic is called ergodic.

\begin{definition}[Detailed Balance Criteria]
    \label{def:reversible}
    A random walk is reversible if and only if for all states $x, y$,
    \begin{equation*}
        \pi(x) P(x, y) = \pi(y) P(y, x) \text{.}
    \end{equation*}
\end{definition}

A distribution $\pi$ is stationary if $\pi = \pi P$.
We will use the following fact.

\begin{theorem}
    \label{thm:stationary-distribution}
    Any irreducible and aperiodic random walk has a unique stationary distribution.
\end{theorem}

While the random walk is guaranteed to converge to its stationary 
distribution, we are interested in establishing a quantitative bound.
We next define the notion of mixing time and give the relevant 
preliminaries; see, e.g., \cite{LevinPeres, Guruswami16} 
for a detailed treatment of the subject.

\begin{definition}
    \label{def:mixing-time}
    The mixing time $\tau(\delta)$ of a random walk with stationary distribution $\pi$ is defined by
    \begin{equation*}
        \tau(\delta) = \max_{i \in \Omega} \min \bigset{t \given (\forall t' \geq t) \sum_{j \in \Omega} \left| P^t_{ij} - \pi_{j} \right| < \delta } \text{.}
    \end{equation*}
\end{definition}

Intuitively, from any initial state $i$ and at any time $t' \geq \tau(\delta)$, the distribution of states is similar to the stationary distribution $\pi$.
We require the following facts regarding mixing time.
First, the mixing time of a product is at most the mixing time of any individual coordinate (up to polynomial factors in the dimension).

\begin{lemma}
    \label{lemma:product-random-walk-mixing-time}
    Suppose $\RW = (\RW_{1}, \dotsc, \RW_{n})$ where $\RW_{i}$ has mixing time $\tau_i(\delta)$.
    Then $\tau(\delta) \leq \max_{i} \tau_i(\delta/n)$.
\end{lemma}

\begin{proof}
    Let $P^{(i)}$ denote the transition matrix for each coordinate $i$ and $\pi^{(i)}$ denote the stationary distribution.
    Then,
    \begin{align*}
        \norm{\prod_{\ell = 1}^{n} P^{(\ell)^{t}} - \prod_{\ell = 1}^{n} \pi^{(\ell)}}{1} \leq \sum_{\ell = 1}^{n} \norm{P^{(\ell)^{t}} - \pi^{(\ell)}}{1} \text{.}
    \end{align*}
\end{proof}

For any ergodic random walk $\RW$ with transition matrix $P$ and stationary distribution $\pi$, we let $\lambda$ denote the eigenvalues of $P$.
The following lemma relates the mixing time to 
a quantity known as the relaxation time, 
or alternatively the absolute spectral gap $\lambda_{*}$.

\begin{definition}
    \label{def:spectral-gap}
    The absolute spectral gap of a random walk
    is $\gamma_{*} = 1 - \lambda_{*}$ where $\lambda_{*} = \max{|\lambda| \given \lambda \neq 1}$.
\end{definition}

\begin{definition}
    \label{def:relaxation-time}
    The relaxation time of a random walk 
    is $t_{\rel} = \frac{1}{\gamma_{*}}$.
\end{definition}

\begin{theorem}[Theorem 12.5 of \cite{LevinPeres}]
    \label{thm:mixing-time-relaxation-time}
    For an ergodic and reversible random walk,
    its mixing time satisfies that
    \begin{equation*}
        \tau(\delta) \geq (t_{\rel} - 1) \log(1/2 \delta) \, ,
    \end{equation*}
    where $t_{\rel}$ is the relaxation time of the random walk defined as in \Cref{def:relaxation-time}.
\end{theorem}

\section{Replicable Closeness Testing Algorithm}
\label{app:closeness-ub}
In this section, we present a replicable closeness tester with nearly optimal sample complexity.
 \begin{algorithm}[H]
\caption{\textsc{RepClosenessTester}  
    }\label{closeness_alg}
    \hspace*{\algorithmicindent} \textbf{Input: }sample access to distributions $\p$ and $\q$ supported on $[n].$ \\
    \hspace*{\algorithmicindent} \textbf{Parameters: } tolerance $\epsilon\in(0,1/4)$ , 
    replicability parameter $\rho\in (0,1/4)$ , 
    support size $n$. \\
    \hspace*{\algorithmicindent} \textbf{Output: }\textsc{accept} if $\p=\q$, \textsc{reject} if $\dtv(\p,\q)\ge \epsilon.$
    \begin{algorithmic}[1]
    \State $m \gets \Theta\left(\frac{n^{2/3}}{\rho^{2/3}\epsilon^{4/3}}+\frac{\sqrt{n}}{\epsilon^2\rho^1} +  \frac{1}{\rho^2\epsilon^2} \right)$,\newline $(m_\p,m_\p',m_\q,m_\q')\gets$ Multinom$(4m,(1/4,1/4,1/4,1/4))$.
    \State Draw two multisets $D_1,D_2$ of i.i.d. samples from $\p$ 
    of sizes $m_{\p},m_{\p}'$ respectively; and two multisets $D_3,D_4$ of iid samples from $\q$ of sizes $m_{\q},m'_{\q}$ respectively. $\forall i\in[n]$ let $X_i,X_i',Y_i,Y_i'$ be the occurrence of $i$ in $D_1,D_2,D_3,D_4$, respectively.
    \State Compute the statistic $\forall i\in [n],Z_i \gets |X_i-Y_i| +|X_i'-Y_i'|-|X_i-X_i'| -|Y_i-Y_i'|$ and $Z \gets \sum_{i=1}^n Z_i$.
    \State Set threshold $r\gets C_1\sqrt{m} +r_0 \left(R-C_1\sqrt{m}\right)$, where 
    $R,C_1$ are the same as in Lemma \ref{exp_gap_closeness}, and $r_0\gets $Unif$\left(\frac{1}{4},\frac{3}{4}\right)$.\\
    \Return \textsc{accept} if $Z \le r$. \textsc{reject} otherwise.
    \end{algorithmic}
    \end{algorithm}
    
    In what follows, we present the analysis of \Cref{closeness_alg}, which serves as the proof of the upper bound statement of \Cref{thm:closeness}.
    \snew{To guarantee correctness 
    of the algorithm},
    the threshold  picked should fall between an upper bound on the test statistic in the completeness case and a lower bound on the test statistic in the soundness case. Secondly, to guarantee replicability we need to further make sure that the randomly picked threshold $r$ falls in the high confidence interval of the statistic \snew{(in which $Z$ typically falls)} with probability at most $\rho$, 
    so that upon multiple runs the algorithm 
    will give identical answers with high probability. 
    
    The main ingredients of the proof are the following lemmas: 
    a bound on the variance of $Z$ and 
    the expectation gap of $Z$ between the case when $\p=\q$ and the case when $\dtv(\p,\q)\ge \epsilon$. 
    Fortunately, the latter was shown in \cite{diakonikolas2021optimal}.
    \begin{lemma}(Expectation Gap, Lemma 3.3 in \cite{diakonikolas2021optimal})
        Given $m, \epsilon, \rho, n, Z$ as specified in \Cref{closeness_alg}, there exists universal constants $C_1,C_2>0$ such that
        \begin{enumerate}[leftmargin=*]
            \item (Completeness) If $\p=\q$, $\E Z \le C_1\sqrt{m}$;
            \item (Soundness) If $\dtv(\p,\q)\ge \epsilon$, $\E Z\ge R:=C_2\min\left( \epsilon m, \frac{m^2\epsilon^2}{n},\frac{m^{3/2}\epsilon^2}{n^{1/2}}\right).$ In particular, $R \ge C_2 \sqrt{m}$ for $m,n,\eps,\rho$ chosen as in \Cref{closeness_alg}.
        \end{enumerate}
        \label{exp_gap_closeness}
    \end{lemma}

    We next bound the variance of $Z$.

    \begin{lemma}
        \label{lemma:var-z-bound}
        Let $Z$ be the test statistic computed by \Cref{closeness_alg}. Then 
            $\Var(Z) \leq 4 m$.
    \end{lemma}

The proof of \Cref{lemma:var-z-bound} 
requires the following standard fact. 
    
    \begin{lemma}[Efron-Stein Inequality]
        \label{lemma:efron-stein-inequality}
        Let $X = (X_1, \dotsc, X_m)$ be independent random variables and $Z = f(X_1, \dotsc, X_m)$. 
        For all $i \in [m]$, define $X^{(i)} = (X_1, \dotsc, X_{i - 1}, X_i', X_{i + 1}, \dotsc, X_{m})$ where $X_i'$ is independently sampled from the same distribution as $X_i$ and $Z_i = f(X^{(i)})$.
        Then
            $\Var(Z) \leq \sum_{i = 1}^{m}  \E{(Z - Z_i)^{2}} $. 
    \end{lemma}

    \begin{proof}[Proof of \Cref{lemma:var-z-bound}]
        Note that if we change a single sample, the test statistic $Z$ changes by at most $2$, since any bucket frequency appears in at most two terms.
        Thus, the variance bound follows from the Efron-Stein inequality.
    \end{proof}

   \begin{proof}[Proof of \Cref{thm:closeness}]
        We first establish the correctness of \Cref{closeness_alg}.
        Suppose $\p = \q$.
        From \Cref{exp_gap_closeness} and \Cref{lemma:var-z-bound}, Chebyshev's inequality implies that
        \begin{equation*}
            \Pr(Z > C_1 \sqrt{m}) < \Pr(|Z - \E{Z}| > C_1 \sqrt{m}) < \bigO{\frac{1}{C_1^2}} \leq \frac{1}{6} 
        \end{equation*}
        for sufficiently large $C_1$.
        On the other hand, if $\dtv(\p, \q) \geq \eps$, Chebyshev's inequality yields that
        \begin{equation*}
            \Pr\left(Z < \frac{3}{4}R + \frac{1}{4} C_1 \sqrt{m} \right) < \Pr\left( |Z - \E{Z}| > \frac{1}{4} R - \frac{1}{4} C_1 \sqrt{m} \right) < \Pr\left( |Z - \E{Z}| > \frac{1}{8} R \right) = \bigO{\frac{1}{C_2^2}} < \frac{1}{6}
        \end{equation*}
        by choosing $C_2$ sufficiently large so that $\frac{1}{8} R \geq C_1 \sqrt{m}$.
    
        Now we establish replicability.
        Recall that we randomly sample a threshold in an interval of width $R - C_{1} \sqrt{m} \geq \frac{R}{2}$ for sufficiently large $C_2$.
        Consider two runs of the algorithm that produce test statistics $Z_1, Z_2$.
        As long as the random threshold $r$ does not lie in $\lp[\min(Z_1, Z_2),\max(Z_1, Z_2)\rp]$ replicability holds.
        This occurs with probability at most $\frac{2 |Z_1 - Z_2|}{R}$.
        Thus, consider the random variable $|Z_1 - Z_2|$.
        The probability $r \in [Z_1, Z_2]$ is exactly
        \begin{align*}
            \Pr_{r, Z_1, Z_2} \left( r \in [Z_1, Z_2] \right) &= \Ep_{Z_1, Z_2}\left[ \Pr_{r} \left( r \in [Z_1, Z_2] \right) \right] \\
            &\leq \Ep_{Z_1, Z_2}\left[ \frac{2|Z_1 - Z_2|}{R} \right] \\
            &\leq \frac{2}{R} \sqrt{\Var(Z)} \\
            &\leq \frac{4 \sqrt{m}}{R} \, ,
        \end{align*}
        where in the first equality we use the law of total probability, the second line follows from our above discussion, the third inequality follows from linearity of expectation, Jensen's inequality, and the definition of variance, and the last inequality follows from \Cref{lemma:var-z-bound}.
        We conclude the proof by considering separate cases based on the functional form of $R$.
        \begin{enumerate}
            \item If $R = C_2 \eps m$, since $m \geq \frac{1600 C_2}{\rho^2 \eps^2}$ we have $\frac{4 \sqrt{m}}{R} = \frac{4 \sqrt{m}}{C_2 \eps m} = \frac{4}{C_2 \eps \sqrt{m}} \leq \frac{\rho}{10}$.
        
            \item If $R = C_2 \frac{m^2 \eps^2}{n}$, since $m \geq \frac{12 C_2 n^{2/3}}{\rho^{2/3} \eps^{4/3}}$ we have $\frac{4 \sqrt{m}}{R} = \frac{4n\sqrt{m}}{C_2 \eps^2 m^2} = \frac{4n}{C_2 \eps^{2} m^{3/2}} \leq \frac{\rho}{10}$.
    
            \item If $R = C_2 \frac{m^{3/2} \eps^2}{n^{1/2}}$, since $m \geq \frac{40 C_2 n^{1/2}}{\rho \eps^{2}}$ we have $\frac{4 \sqrt{m}}{R} = \frac{4 \sqrt{m n}}{C_2 \eps^{2} m^{3/2}} = \frac{4 \sqrt{n}}{C_2 \eps^{2} m} \leq \frac{\rho}{10}$.
        \end{enumerate}
    
        This concludes the proof of \Cref{thm:closeness}.
    \end{proof}

\section{Replicable Independence Testing Algorithm}
\label{sec:independence-ub-body}
In this section, we give our replicable independence tester.
At a high level, we compute the same statistic used by the high probability independence tester from \cite{diakonikolas2021optimal}, but average over the internal randomness of the tester to enhance replicability.

Our starting point is the (randomized) flattening technique developed in \cite{diakonikolas2016new} and enhanced in~\cite{diakonikolas2021optimal} that helps decrease the $\ell_2$ norm of input distributions while maintaining the properties to be tested in total variation distance.
The original description is as follows.
First, one draws a set of samples $X$, and randomly partitions $X$ into a flattening sample set, and a testing sample set.
Next, one uses the flattening samples to determine the number of sub-bins for each original domain element, and then randomly assigns original testing samples to the sub-bins.
For our analysis, it is more convenient to consider an equally effective random process, where we randomly sort all samples, partition them into flattening and testing samples, and make two testing samples be in the same sub-bin if and only if they are originally from the same bin and there are no flattening samples from the same bin between them. 
The formal description is as follows.
\begin{definition}[Randomized Flattening]
Let $X=\{X_1, \cdots, X_m\}$ be a multiset of samples over $[n]$, and $F \in \{0, 1\}^m$ be a binary vector. 
Then the randomized flattening procedure 
$X^f:=\{ X_\ell^f \}_{\ell : F_{\ell} = 0  } \gets \text{Flatten}( \{ X_\ell \}_{\ell=1}^m;  F )$
is as follows.
(1) Assign a random order $\sigma$ to the samples.
(2) For each sample $X_\ell$, count the number of samples $X_{\ell'}$ before it according to $\sigma$ such that
$X_{\ell'} = X_{\ell}$ and $F_{\ell'} = 1$. Denote the number as $f_\ell$.
(3) For each $ \ell $ such that $F_{\ell} = 0$,
set $X_{\ell}^f \gets ( X_{\ell}, f_{\ell} )$.
Moreover, given a parameter $\alpha \in (0, 1)$,
we denote by $\text{Flatten}( X;\alpha)$ the randomized sample set obtained from $X^f \gets \text{Flatten}( X ; F)$, where $F \sim \B(\alpha)^{\otimes m}$. 
\end{definition}

A useful byproduct of the flattening procedure is that it ensures that there will be no ``heavy'' bins after the operation with high probability.
\begin{lemma}[Flattening splits heavy bins]
\label{lem:no-heavy-element}
Let $S$ be a set of samples over $[n]$
with $|S| = \poly{n}$,  
and $S^f = \text{Flatten}( S; \alpha )$.
Denote by $T^f$ the sample count vector of the flattened samples.
For any constant $C$, it holds that $T_i^f \leq O\lp( \alpha^{-1} \log n \rp) $ for all $i \in \text{supp}\lp( S^f \rp)$ with probability at least $1 - n^{-C}$,
where the randomness is over the internal randomness of $\text{Flatten}(\cdot)$.
\end{lemma}
\begin{proof}
After sorting the samples in $S$, we note that the position of the first flattening sample follows exactly a geometric distribution with mean $1 / \alpha$.
Denote by $Y$ its position.
We have that $ \Pr[ Y\geq t] = (1 - \alpha)^{t-1} 
= \exp \lp(  (t-1)  \log(1 - \alpha) \rp)
\leq \exp \lp(  -\alpha (t-1)  \rp)$.
In particular, this implies that 
$\Pr[ Y \geq  a \alpha^{-1} \log(n) ] \leq n^{-a} $ for any number $a > 0$.
This shows that with probability at least $n^{-a}$ it holds that the number of samples falling in the first bin is at most $a \alpha^{-1} \log(1/n)$.
If we choose $a$ to be a sufficiently large constant, we then have that $Y \leq  O\lp( \alpha^{-1} \log(n) \rp)$ with probability at least $1 - 1 / \text{poly}(n)$.
Since $Y$ is also the number of samples within the first sub-bin, \Cref{lem:no-heavy-element} follows from applying this argument 
to all subsequent samples and the union bound.
\end{proof}

In independence testing, we need to perform the flattening operation on the  marginals of a multi-dimensional distribution independently.
For clarity, we formalize this operation below.
\begin{definition}[Multi-dimensional Flattening]
\label{def:multi-flatten}
Let $\alpha, \beta \in (0,1)$, and $P=\{ P_\ell = (X_\ell, Y_\ell) \}_{\ell=1}^m$ be a multiset of samples over $[n_1] \times [n_2]$.
Then the multi-dimensional flattening operation $\{P_\ell^f\} \gets
\text{Flatten}( \{P_\ell\}_{\ell=1}^m ; \alpha, \beta  )
$
is as follows.
(1) Choose $F^x \sim \B(\alpha)^{\otimes m}$ and $F^y \sim \B(\beta)^{\otimes m}$.
(2) $ \{ X_{\ell}^f \}_{\ell:F^x_\ell = 0}\gets
\text{Flatten}( \{ X_{\ell} \}_{\ell=1}^m; F^x)
$, $\{ Y_{\ell}^f \}_{\ell:F^y_\ell = 0} \gets
\text{Flatten}( \{ Y_{\ell} \}_{\ell=1}^m; F^y )
$.
(3) Map $P_\ell$ to $P_\ell^f \gets
( X_\ell^f, Y_\ell^f )$ if $F^x_\ell = F^y_\ell = 0$.
When flattening two bags of samples $A$ and $B$ together, we denote by $A^f\cup B^f\gets
\text{Flatten}( A\cup B ; \alpha, \beta  )$, where $A^f$($B^f$, resp.) contains all elements mapped from $A$(B, resp.).
\end{definition}

Another key idea behind the tester from \cite{diakonikolas2021optimal} is to use the samples from $\p$ to simulate samples from another product distribution $\q$ that equals to the product of the marginals of $\p$. In particular, a sample from $\q$ can be simulated by taking two samples from $\p$, and combining the first coordinate of the first sample to the second coordinate of the second sample.
Hence, we can readily assume that we have sample access to both $\p$ and such $\q$.
\begin{definition}[Product of Marginals]
Given a distribution $\p$ over $[n_1] \times [n_2]$, we say that 
$\q$ is the product of marginals of $\p$ if the marginals of $\q$ agree with those of $\p$ and $\q$ is a product distribution.
\end{definition}
Given  samples from the original distribution, and the ones from the product of the marginals, the final step of the tester from \cite{diakonikolas2021optimal} is to compute the closeness test statistic, which we reiterate below. 
\begin{definition}[Closeness Statistic]
Given two bags of samples $S_\p, S_\q$ over some finite discrete domain, the closeness statistic
$Z_C(S_\p, S_\q)$ is defined as follows.
(1) For each sample in $S_\p \cup S_\q$, mark it independently with probability $1/2$.
(2) For $i \in  supp\lp(S_\p \cup S_\q \rp)$, 
let $T_{i}^{\p_0}$, $T_{i}^{\q_0}$ be the number of times the element $i$ appears marked in $S_\p, S_\q$, and $T_{i}^{\p_1}$, $T_{i}^{\q_1}$ be the corresponding counts of the unmarked samples.
(3) Compute $Z_C(S_\p, S_\q) \gets \lp| T_{i}^{\p_0} -  T_{i}^{\q_0}\rp|
+ \lp| T_{i}^{\p_1} -  T_{i}^{\q_1}\rp|
- \lp| T_{i}^{\p_0} -  T_{i}^{\p_1}\rp|
- \lp| T_{i}^{\q_0} -  T_{i}^{\q_1}\rp|$.
\end{definition}
A useful fact of this test statistic is that any singleton sample does not contribute to its value.
\begin{fact}
\label{clm:ignore-singleton}
Consider two sets of samples $S_\p, S_\q$ over some finite discrete domain.
Assume that $P$ is a singleton sample among $S_\p \cup S_\q$.
It holds that 
$
\Ep [ Z_C(S_\p, S_\q) ]  = \Ep [ Z_C(S_\p \backslash \{ P \}, S_\q \backslash \{ P \}) ],
$ where the randomness is over the internal randomness of the test statistic $Z_C$.
\end{fact}

We are now ready to state the tester from \cite{diakonikolas2021optimal}, which forms the building block of our replicable independence tester.
\begin{algorithm}[H]
\caption{\textsc{IndependenceStats}}\label{independence_alg}
    \hspace*{\algorithmicindent} \textbf{Input: } a sample set $S_{\p}$ from the unknown distribution $\p$ over $[n_1] \times [n_2]$, where $n_1 \geq 
 n_2$, and another sample set $S_{\q}$ from $\q$, the product of marginals of $\p$. 
    \\
    \hspace*{\algorithmicindent} \textbf{Parameters: }
    domain sizes $n_1 \geq n_2$ , tolerance $\epsilon\in(0,1/4)$ , replicability $\rho\in (0,1/4)$.  \\
    \hspace*{\algorithmicindent} \textbf{Output: } A test statistic related to whether these samples came from an independent distribution.
    \begin{algorithmic}[1]
    \State Set $m \gets \tilde \Theta \lp(  \indcomplexity \rp)$.
    \label{line:m-def}
    \State Set $\alpha \gets \min( n_1/(100 m ), 1/100)$, $\beta = n_2 / (100 m )$.
    \label{line:alpha-beta-def}
    \State Compute the flattened samples
    $S_{\p}^f\cup S_{\q}^f \gets
    \text{Flatten}\lp(  S_{\p} \cup S_{\q} ; \alpha, \beta\rp)$.
    \State Abort and return $0$ if $|S_{\p}| -  |S_{\p}^f| > 10 n_1$ or
    $|S_{\q}| - |S_{\q}^f| > 10 n_2$.
    
    \State Sample $\ell, \ell' \sim \Poi(m)$. Abort and return $0$ if $\ell > |S_{\p}^f|$ or $\ell' > |S_{\q}^f|$.
    \State Keep only the first $\ell$ samples of $S_{\p}^f$ and only the first $\ell'$ samples of $S_{\q}^f$.
    \label{line:flatten}
    \State Compute and return the closeness test statistic
    $Z_C(S_{\p}^f, S_{\q}^f)$.
    \label{line:awd}
    \end{algorithmic}
\end{algorithm}

A basic property we need is that the final test statistic computed has a wide expectation gap.
\begin{lemma}[Expectation Gap of Independence Statistics; Section 3.2 and Claim 4.14 of \cite{diakonikolas2021optimal}]
\label{lem:independence-expectation-gap}
Let $\p$ be some unknown distribution over $[n_1] \times [n_2]$,  $\q$ be the product of marginals of $\p$, and $m$ be defined as in Line~\ref{line:m-def} of \textsc{IndependenceStats}.
Let $S_\p,S_\q$ be samples from $\p,\q$ respectively with size  $ 
 |S_{\p}| = |S_{\q}| = 100 m$, and $Z \gets \textsc{IndependenceStats}( S_{\p}, S_{\q})$.
 Define $G := \zexpgap$. Then the following holds:
 \begin{itemize}[leftmargin=*]
\item If $\p$ is a product distribution, then $\Ep[Z] \le C_{I_1} G$. 
\item 
If $\p$ is $\eps$-far from any product distribution in TV distance, then 
$\Ep[Z] >  C_{I_2} G$ for some  constants $C_{I_1} < C_{I_2}$.
\end{itemize}
\end{lemma}

To make the final test statistic computed by $\textsc{IndependenceStats}$ more replicable, we consider a new test statistic $Z_a$ computed by averaging over the internal randomness of  $\textsc{IndependenceStats}$.
\begin{definition}[Averaged Independence Statistic]
\label{def:average-Z}
Let $S_{\p}, S_{\q}$ be samples over $[n_1] \times [n_2]$.
We define $Z_a( S_{\p}, S_{\q})$ to be the expected value of $\textsc{IndependenceStats}\lp(S_{\p}, S_{\q}\rp)$ averaged over the internal randomness of 
$\textsc{IndependenceStats}$.
\end{definition}
An important quantity in analyzing the concentration of the above new statistic is the concept of non-singleton sample counts.
\begin{definition}[Non-singleton Sample Count]
Let $S$ be a set of samples over some finite discrete domain.
We define the non-singleton sample count $N(S)$ as
the total number of samples within $S$ that collide with another sample.
\end{definition}
Specifically, we focus on the non-singleton sample count of the flattened sample sets $S_{\p}^f, S_{\q}^f$ constructed by $\textsc{IndependenceStats}$.
\begin{definition}[Non-singleton Sample Count of Flattened Samples]
\label{def:N_a}
Let $S_{\p}, S_{\q}$ be two arbitrary sets of samples over $[n_1] \times [n_2]$.
Consider the two random flattened sample sets $S_{\p}^f, S_{\q}^f$ constructed by $\textsc{IndependenceStats}(S_{\p}, S_{\q})$ on Line~\ref{line:flatten} \footnote{We think of the two sets as being empty if the algorithm aborts before reaching Line~\ref{line:flatten}.}.
We define $N_a(S_{\p}, S_{\q})
:= \Ep \lp[ N( S_{\p}^f \cup S_{\q}^f ) \rp]$, where the expectation is over the internal randomness of \textsc{IndependenceStats}.
\end{definition}
Our main insight is that we can bound the variance of the new statistic $Z_a\lp( S_{\p}, S_{\q}\rp)$ by the expected value of  $N_a(  S_{\p} , S_{\q })$.
\snew{In fact, we will prove something that is slightly more general. In particular, we show that the variance of the averaged version of any sufficiently ``stable'' statistic computed based on the random flattened sample sets $S_\p^f, S_\q^f$ can be controlled by the expected value of $N_a(S_\p, S_\q)$.
The formal statement is as follows.}
\begin{lemma}[Bound on Variance of Averaged  Statistic in Expected Non-Singleton Sample Count]
\label{lem:AS-var-bound}
Let $\p$ be a distribution over $[n_1] \times [n_2]$, and $\q$ be the product of marginals of $\p$.
Consider the two random flattened sample sets $S_{\p}^f, S_{\q}^f$ constructed by $\textsc{IndependenceStats}(S_{\p}, S_{\q})$ on Line~\ref{line:flatten} \footnote{Similarly, we think of the two sets as being empty if the algorithm aborts earlier.}. 
\snew{Let $L(S_{\p}^f, S_{\q}^f)$ be a test statistic satisfying that (1) $L(S_{\p}^f, S_{\q}^f)$ stays unchanged if any singleton sample from $S_{\p}^f$ or $S_{\q}^f$ is removed, and (2)$L(S_{\p}^f, S_{\q}^f)$ changes by at most $O(1)$ if a non-singleton sample is removed.
Consider the averaged non-singleton sample count $N_a(S_\p, S_\q)$ defined as in \Cref{def:N_a}, and
the averaged test statistic 
$L_a(S_{\p}, S_{\q}) := \Ep \lp[ 
L(S_{\p}^f, S_{\q}^f)
\rp]$, where the expectation is over the randomness in constructing the flattened sample sets $S_{\p}^f, S_{\q}^f$.}
Then it holds that
$ \Var[ L_a( S_{\p}, S_{\q }) ] \leq O( \log^3 (n_1n_2) ) \Ep[ N_a(  S_{\p }, S_{\q }) ]$, where the randomness is over the samples $S_{\p}, S_{\q}$.
\end{lemma}

\begin{proof}
We will bound the variance of $L_a(S_{\p}, S_{\q})$ by the expected sum over samples of the squared difference in the final test statistic by removing each sample. 

In particular, suppose that $S_{\p}, S_{\q}$ contains the samples $\{ P_{\ell} \}_{\ell=1}^{k}$. 
For convenience, we denote by $S_{\p, -\ell}, S_{\q, -\ell}$ the corresponding set after removing the sample $P_{\ell}$.\footnote{If $ P_\ell \not \in S_{\p}$, then $S_{\p, -\ell} = S_{\p}$, and the same for $S_{\q, -\ell}$.}
It then follows from the Efron–Stein inequality that
\begin{align}
\label{eq:efron}
\Var[ L_a(S_\p, S_\q) ]
\leq O(1) \; 
\Ep\lp[ 
\sum_{\ell=1}^k
\lp( 
L_a\lp( S_{\p, -\ell}, S_{\q, -\ell}
\rp)
- L_a\lp( S_\p , S_\q \rp) \rp)^2
\rp] \, ,
\end{align}
where the randomness is over the samples.
Fix some sample sets $S_\p, S_\q$.
For notational convenience, 
consider the random variables $Z = L_a\lp(  S_\p, S_\q \rp)$ and 
$Z_{-\ell} = L_a\lp(S_{\p, -\ell}, S_{\q, -\ell}\rp)$.
We claim that it suffices for us to show that
\begin{align}
\label{eq:internal-randomness-bound}
\sum_{\ell=1}^k
\lp( 
\Ep \lp[ 
Z 
\rp]
-  \Ep \lp[  Z_{-\ell}  \rp] \rp)^2 
\leq O\lp( \log^2(n_1) \rp) N_a(S_\p, S_\q) \, ,
\end{align}
where the expectation is over the internal randomness of $\textsc{IndependenceStats}$.
After that, taking expectation over the randomness of the samples on both sides of \Cref{eq:internal-randomness-bound}
and combining it with \Cref{eq:efron} then concludes the proof.

It then remains for us to show \Cref{eq:internal-randomness-bound}.
Recall that the tester first partitions the samples into flattening samples and testing samples randomly.
We denote by $S_\p^f, S_\q^f$ the flattened testing samples constructed in Line~\ref{line:flatten} from the original sample set $S_\p, S_\q$, and
$S_{\p, -\ell}^f, S_{\q, -\ell}^f$ the ones constructed from the leave-one-out sample sets $S_{\p, -\ell}, S_{\q, -\ell}$.

Denote by $F_{\ell}^x, F_{\ell}^y \in \{0, 1\}$ the indicator variables of whether $P_\ell$ is selected for row  or column flattening purpose respectively (see \Cref{def:multi-flatten}) while constructing $S_\p^f, S_\q^f$.
We then break into cases based on the values of $F_{\ell}^x$, $F_{\ell}^y$.

In the first case, we have that $F_{\ell}^x = F_{\ell}^y = 0$. 
This suggests that the $\ell$-th sample is not selected as a flattening sample.
Hence, there exists a flattened version of $P_\ell$, which we denote by
$P_{\ell}^f$, within $S_\p^f \cup S_\q^f$.
In this case, $S_{\p, -\ell}^f, S_{\q, -\ell}^f$ are obtained by deleting exactly $P_\ell^f$ from $S_{\p}^f, S_{\q}^f$.
There are then again two sub-cases. 
Either $P_{\ell}^f$ is a singleton sample among $S_{\p, -\ell}^f \cup S_{\q, -\ell}^f$.
In that case, 
we must have $ L(S_\p^f, S_\q^f) =  
L( S_{\p,-\ell}^f,  S_{\q,-\ell}^f)
$ by property (1) of the test statistic $L(\cdot)$.
Otherwise, by property (2) of $L(\cdot)$, we have that
$ \lp| L(S_\p^f, S_\q^f) - L( S_{\p,-\ell}^f, S_{\q,-\ell}^f) \rp| \leq O(1) $.
As a result, it follows that
\begin{align}
\label{eq:case-1}
\sum_{\ell=1}^k
\lp( 
\Ep[ Z  \; \mathds 1 \{ F_{\ell}^x = F_{\ell}^y = 0 \} ]
- 
\Ep[ Z_{-\ell} ] \rp)^2
\leq \Ep[
N( S_\p^f \cup S_\q^f )].
\end{align}

Next, consider the case that $F_{\ell}^x = 1$ and 
$F_{\ell}^y = 0$, which happens with probability at most 
$\p_x := \min \lp(  n_1/(100m), 1/100 \rp)$.
This suggests that the $P_{\ell}$ is selected as a row flattening sample. 
Denote by $K_\ell$ the number of samples lying in the same row as $P_\ell$.
Consider the following coupling between $(S_\p^f, S_\q^f)$ (conditioned on $F_\ell^x = 1$ and $F_\ell^y = 0$) and $(S_{\p, -\ell}^f, S_{\q, -\ell}^f)$:
(1)Pick a random sub-row $a$ (in the flattened domain) weighted by the total sample count of sub-row $a$ within $  S_{\p,-\ell}^f \cup  S_{\q,-\ell}^f $ divided by  $K_\ell$, (2) subdivide the sub-row into two sub-rows $a_1, a_2$, and (3) randomly assign the samples from $a$ to $a_1, a_2$.
Denote by $T_i$ the total number of samples among $S_{\p, -\ell}^f \cup S_{\q, -\ell}^f$
within the sub-row $i$, and $N_i$ the corresponding non-singleton sample count.
Note that if a flattened sample is a singleton sample among $S_{\p,-\ell}^f \cup  S_{\q,-\ell}^f$, then it remains a singleton sample after the subdivision, and hence has no impact on the final closeness statistic.
Therefore, such subdivision can change the final statistic 
by at most $O(N_i)$.
On the other hand, the probability of the sub-row $i$ being selected and the sample $\ell$ being selected for flattening purpose is at most
$  \alpha \; T_i / K_\ell  $.
Hence, the averaged statistic changes by at most 
$$ 
\lp( 
\Ep[ Z  \; \mathds 1 \{ F_{\ell}^x =1, F_{\ell}^y = 0 \} ]
- 
\Ep[ Z_{-\ell} ] \rp)^2
\leq 
O(1) \; \lp( \sum_{i: \text{sub-rows of the row of} P_\ell }  
\Ep\lp[  \frac{N_i  \alpha T_i}{K_\ell} \rp] \rp)^2.
$$
Summing over all $\ell'$ such that $P_\ell$ and $P_\ell'$ lie in the same row
then gives that
$$ 
\sum_{\ell': P_\ell \text{ lies in the same row as } P_{\ell'}}
\lp( 
\Ep[ Z  \; \mathds 1 \{ F_{\ell}^x =1, F_{\ell}^y = 0 \} ]
- 
\Ep[ Z_{-\ell} ] \rp)^2
\leq
O(1) \; 
{
\Ep^2\lp[ 
\sum_{i: \text{sub-rows of the row of} P_\ell }   N_i  \alpha T_i
\rp]} / {K_\ell}.
$$
By \Cref{lem:no-heavy-element}, we have that $T_i$ is at most $ \log(n_1) \alpha^{-1}$ with probability at least $1 - 1 / \poly{n_1}$.
Moreover, 
$\sum_{i: \text{sub-rows of the row of} P_\ell }   N_i$ is always at most $K_\ell$.
It then follows that
\begin{align*}
&\sum_{\ell': P_\ell \text{ lies in the same row as } P_{\ell'}}
\lp( 
\Ep[ Z  \; \mathds 1 \{ F_{\ell}^x =1, F_{\ell}^y = 0 \} ]
- 
\Ep[ Z_{-\ell} ] \rp)^2 \\
&\leq
O( \log^2 n ) \; 
{
\Ep^2\lp[ 
\sum_{i: \text{sub-rows of the row of} P_\ell }   N_i
\rp]} / {K_\ell} \\
&\leq 
O( \log^2 n ) \; 
{
\Ep\lp[ 
\sum_{i: \text{sub-rows of the row of} P_\ell }   N_i
\rp]}.
\end{align*}
Note that the non-single sample count can only increase conditioned on that $P_\ell$ is not selected for flattening, which happens with at least constant probability.
As a result, 
the expected number of non-singleton samples among $S_{\p, -\ell}^f \cup S_{\q, -\ell}^f$ is always at most a constant factor of 
the expected number of non-singleton samples among 
$S_{\p}^f \cup S_{\q}^f$.
Summing over all $\ell$ then gives that 
\begin{align}
\label{eq:case-2}
\sum_{\ell=1}^k
\lp( 
\Ep[ Z  \; \mathds 1 \{ F_{\ell}^x =1, F_{\ell}^y = 0 \} ]
- 
\Ep[ Z_{-\ell} ] \rp)^2 
\leq 
O( \log^2 n_1 ) \; 
\Ep\lp[ 
N( S_\p^f, S_\q^f )
\rp].
\end{align}
This then concludes the analysis of the second case.

In the third case, we assume that $F_{\ell}^x =0, F_{\ell}^y = 1$. 
Using an argument that is almost identical to the second case, one can show that 
\begin{align}
\label{eq:case-3}
\sum_{\ell=1}^k
\lp( 
\Ep[ Z  \; \mathds 1 \{ F_{\ell}^x =0, F_{\ell}^y = 1 \} ]
- 
\Ep[ Z_{-\ell} ] \rp)^2 
\leq 
O( \log^2 n_2 ) \; 
\Ep\lp[ 
N( S_\p^f, S_\q^f )
\rp] \;,
\end{align}
as this corresponds to the case when the $k$-th sample is chosen for column flattening.

Finally, for $F_k^x = F_k^y = 1$, we can use an argument that is almost identical to the second case to show that 
\begin{align*}
\sum_{\ell=1}^k
\lp( 
\Ep[ Z  \; \mathds 1 \{ F_{\ell}^x =1, F_{\ell}^y = 1 \} ]
- 
\Ep[ Z  \; \mathds 1 \{ F_{\ell}^x =0, F_{\ell}^y = 1 \} ]
\rp)^2 
\leq 
O( \log^2 n_1 ) \; 
\Ep\lp[ 
N( S_\p^f, S_\q^f )
\rp].
\end{align*}
It then follows from the triangle inequality that 
\begin{align}
&\sum_{\ell=1}^k
\lp( 
\Ep[ Z  \; \mathds 1 \{ F_{\ell}^x =0, F_{\ell}^y = 1 \} ]
- 
\Ep[ Z_{-\ell} ] \rp)^2  \nonumber \\
&\leq 
O(1) \; 
\bigg( 
\sum_{\ell=1}^k
\lp( 
\Ep[ Z  \; \mathds 1 \{ F_{\ell}^x =1, F_{\ell}^y = 1 \} ]
- 
\Ep[ Z  \; \mathds 1 \{ F_{\ell}^x =0, F_{\ell}^y = 1 \} ]
\rp)^2  \nonumber \\
&+  
\sum_{\ell=1}^k
\lp( 
\Ep[ Z  \; \mathds 1 \{ F_{\ell}^x =0, F_{\ell}^y = 1 \} ]
- 
\Ep[ Z_{-\ell} ]
\rp)^2 
\bigg)
\leq O( \log^2 n_1 )
\Ep\lp[ 
N( S_\p^f, S_\q^f )
\rp].
\label{eq:case-4}
\end{align}
Combining the case analysis (\eqref{eq:case-1}, \eqref{eq:case-2}, \eqref{eq:case-3}, \eqref{eq:case-4})
then yields that
$$
\sum_{\ell=1}^k \lp(  \Ep[Z] - \Ep[ Z_{-\ell} ] \rp)^2
\leq O( \log^2 n_1 )
\Ep \lp[  N( S_\p^f \cup S_\q^f ) \rp] =
O( \log^2 n_1 ) N_a(S_\p, S_\q) \;.
$$
This concludes the proof of \eqref{eq:internal-randomness-bound} as well as \Cref{lem:AS-var-bound}.
\end{proof}

By \Cref{clm:ignore-singleton}, the closeness statistic is 
indeed insensitive to removal of any singleton sample. 
Moreover, it is not hard to check from its definition that 
it may change by at most $O(1)$ after removal of any non-
singleton sample.
Hence, as an immediate corollary of \Cref{lem:AS-var-bound}, we have the desired bound on the variance of the averaged independence statistic $Z_a(S_\p, S_\q)$.
\begin{corollary}[Bound on Variance of Averaged Independence Statistic in Expected Non-Singleton Sample Count]
\label{lem:IS-var-bound}
Let $\p$ be a distribution over $[n_1] \times [n_2]$, and $\q$ be the product of marginals of $\p$.
Let $S_{\p}, S_{\q}$ be samples from $\p, \q$ respectively.
Consider the averaged independence statistics $Z_a( S_{\p}, S_{\q })$, and the averaged non-singleton sample count
$N_a(  S_{\p }, S_{\q })$.
Then it holds that
$ \Var[ Z_a( S_{\p}, S_{\q }) ] \leq O( \log^3 (n_1n_2) ) \Ep[ N_a(  S_{\p }, S_{\q }) ]$, where the randomness is over the samples $S_{\p}, S_{\q}$.
\end{corollary}

It then remains for us to control $ N_a( S_\p, S_\q )$.
Fortunately, the expected value of the non-singleton count has already been shown to be small by \cite{diakonikolas2021optimal} when the underlying distribution $\p$ is known to be a product distribution.
\begin{lemma}[Expected Non-singleton Sample Count under Product Distribution, Lemma 4.9 of \cite{diakonikolas2021optimal}]
\label{lem:N-expectation-bound}
Let $m \in \mathds{Z}_+, \alpha, \beta \in (0, 1)$ be defined as in Line~\ref{line:m-def} and Line~\ref{line:alpha-beta-def} from \textsc{IndependenceStats} respectively. 
Let $S$ be samples from a product distribution $\q$ over $[n_1] \times [n_2]$ with $|S| = 100 m$.
Consider the random variable 
$N(  S^f )$, where $S^f \gets \text{Flatten}(S;\alpha, \beta)$.
Then there exists a universal constant $C_N$ such that
$  \Ep[ N(  S^f ) ]
\leq 
C_N  
\max \lp( 
m^2/(n_1 n_2),
m/ n_2
\rp)
$, where the randomness is over the internal randomness of $\text{Flatten}(\cdot)$ as well as the samples.
\end{lemma}

This motivates a two-stage testing strategy: we can first test that the expected value of 
$N_a( S_\p \cup S_\q )$ is sufficiently small, and then compute the averaged statistic $Z_a( S_\p, S_\q )$.
To ensure replicability of the first testing stage, we also need to control the variance of the averaged non-singleton sample count $N_a( S_\p \cup S_\q )$.
Fortunately, the non-singleton sample count is itself a statistic that (1) stays invariant after removal of any singleton sample, and (2) changes by at most $2$ after removal of any non-singleton sample.
Hence, as another application of \Cref{lem:AS-var-bound}, we obtain an upper bound on the variance of $N_a(S_\p, S_\q)$ also in terms of its expectation.
\begin{corollary}[Bound on Variance of Averaged Non-Singleton Sample Count]
\label{lem:N-variance-bound}
Let $\p$ be a distribution over $[n_1] \times [n_2]$, and $\q$ be the product of marginals of $\p$.
Let $S_\p, S_\q$ be samples from $\p, \q$ respectively.
Consider the random variable 
$N_a(S_\p, S_\q)$ defined as in \Cref{def:N_a}.
Then it holds that
$ \Var[ N_a(S_\p, S_\q) ] \leq O( \log^3 (n_1 n_2) ) \Ep[ N_a(S_\p, S_\q) ]$, where the randomness is over $S_\p, S_\q$.
\end{corollary}

Using \Cref{lem:N-variance-bound}, we show that we can replicably test whether $\Ep[ N_a(S_\p, S_\q) ]$ is on the order of $\max \lp( m^2/(n_1 n_2),
m/ \min(n_1, n_2) \rp)$ by simply drawing random sample sets $S_\p, S_\q$, (approximately) computing $N_a(S_\p, S_\q)$, and then comparing it with an appropriately chosen random threshold.

We are now ready to present our full independence tester. 
\begin{algorithm}[H]
\caption{\textsc{RepIndependenceStats}}\label{rep_independence_alg}
    \hspace*{\algorithmicindent} \textbf{Input: } sample access to an unknown distribution $\p$ over $[n_1] \times [n_2]$ \\
    \hspace*{\algorithmicindent} \textbf{Parameter: 
    }tolerance $\epsilon\in(0,1/4)$ , replicability parameter $\rho\in (0,1/4)$ . \\
    \hspace*{\algorithmicindent} \textbf{Output: } Whether $\p$ is a product distribution.
    \begin{algorithmic}[1]
    \State Let $m$ be defined as in \Cref{line:m-def} of \textsc{IndependenceStats}.
    \State $\bar S_\p \gets 100m$ samples from $\p$, and  $\bar S_\q \gets 100m$ samples from $\q$, the product of marginals of $\p$.
    \State Estimate $N_a( \bar S_\p, \bar S_\q )$ (see \Cref{def:N_a})
    up to error $o(1)$ by running $ \textsc{IndependenceStats}( \bar S_\p, \bar S_\q )$ with fresh randomness for sufficiently many times. 
    \State Draw $r \sim \mathcal U([2C_N, 100 C_N])$, where $C_N$ is the constant from \Cref{lem:N-expectation-bound}.
    \State Reject if (estimated) $N_a( \bar S_\p, \bar S_\q ) > r \max \lp( m^2 / (n_1 n_2), m/n_2\rp)$.
    \State $ S_\p \gets 100m$ samples from $\p$, and  $S_\q \gets 100m$ samples from $\q$, the product of marginals of $\p$.
    \State Estimate $Z_a(S_\p, S_\q)$ (see \Cref{def:average-Z}) up to error $o(1)$
    by running $ \textsc{IndependenceStats}( S ) $ with fresh randomness for sufficiently many times.
    \State Draw  $r \sim \mathcal U([C_{I_1}, C_{I_2}])$, where $C_{I_1}, C_{I_2}$ are constants from \Cref{lem:independence-expectation-gap}.
    \State Reject if (estimated) $Z_a(S_\p, S_\q) > r \; \zexpgap $. Otherwise, accept.
    \end{algorithmic}
\end{algorithm}

The full analysis of our replicable independence tester is as follows.

\begin{proof}[Proof of \Cref{thm:independence}]
\label{pf-thm:independence}
Recall that the algorithm has two steps.
In the first step, it verifies that the size of the expected value of the non-singleton sample count is not large by comparing $N_a(S_\p, S_\q)$ with a random threshold. 
In the second step, it computes the averaged independence statistics $Z_a(S_\p, S_\q)$ with fresh samples, and compare it with another appropriately chosen random threshold.

We first analyze the correctness and replicability of the first step. 
Let $m$ be defined as in \Cref{line:m-def} of \textsc{IndependenceStats}, and $S_\p, S_\q$ be sample sets with size $100m$.
By \Cref{lem:N-expectation-bound}, the expected number of non-singleton sample count 
$\Ep \lp[ N_a(S_\p, S_\q) \rp]$
is at most $C_N \max \lp( m^2 / (n_1 n_2), m/n_2 \rp)$ for some constant $C_N$
if the underlying distribution $\p$ is indeed a product distribution. 
By \Cref{lem:N-variance-bound}, we have that
$\Var\lp[ N_a(S_\p, S_\q) \rp] \leq \Ep \lp[ N_a(S_\p, S_\q) \rp]$.
We first show the validity of the following bound:
\begin{align}
\label{eq:na-var-key}    
\log^2(n_1) \max\lp(  m^2 / (n_1 n_2), m/n_2 \rp)
\ll 
 \rho^2 \lp( \max\lp(  m^2 / (n_1 n_2), m/n_2 \rp) \rp)^2.
\end{align}
In particular, we will see that for this step, it is sufficient if
$m \gg n_1^{2/3} n_2^{1/3} \rho^{-2/3}
+ \sqrt{n_1 n_2} \rho^{-1}$.
We begin with a case analysis.
In the first case, we have that
$m^2 / (n_1 n_2)$ is the dominating term in \eqref{eq:na-var-key}. 
It is not hard to verify that 
$$
m^2 / (n_1 n_2) \leq \rho^2 m^4 / (n_1 n_2)^2
$$
as long as $m \gg \sqrt{n_1 n_2} \rho^{-1}$. So \Cref{eq:na-var-key} easily holds in this case.
In the second case, we have that 
$m^2 / (n_1 n_2) \leq  m/n_2$ and so $m/n_2$ is the dominating term.
In this case, we need to show that 
$m/n_2 \leq \rho^2 (m/n_2)^2$, which is true as long as $m \geq n_2 \rho^{-2}$.
In particular, the case assumption indicates that 
$n_1^{2/3} n_2^{1/3} \rho^{-2/3}\ll m < n_1$. This implies that
$n_1 \gg n_2 \rho^{-2}$, which further implies that
$m \gg \sqrt{n_1 n_2} \rho^{-1} > n_2 \rho^{-2}$. This hence concludes the proof of \eqref{eq:na-var-key}.

To argue the correctness of the tester, we analyze the completeness and the soundness cases separately.
Denote by $G:= C_N \max \lp( m^2 / (n_1 n_2), m/n_2 \rp)$.
In the completeness case, the expectation is at most $G$, and the variance is at most $ O(1) \; \log^2(n_1)  G $.
By Chebyshev's inequality and \Cref{eq:na-var-key}, the statistic $N_a(S_\p, S_\q)$ will be at most  $2G$ with high constant probability.
In the soundness case, suppose $\Ep\lp[  N_a(S_\p, S_\q) \rp] \geq 101 G$.
It is not hard to verify that $G \gg \log^2(n_1)$ as our choice of $m$ ensures that $m \gg \log^2(n_1) \sqrt{n_1 n_2} \geq  \log^2(n_1) n_2$.
In particular, this implies that 
$\sqrt{ \log(n_1) G } \ll G$.
In this case, by Chebyshev's inequality, $N_a(S_\p, S_\q)$ is at least $101 G - \sqrt{ 
\log^2(n_1) G} \geq 100 G$.
The above ensures that the tester will be correct with high constant probability. 
Combining this with the standard median trick then ensures correctness with probability at least $1 - \rho$ at the cost of increasing the sample complexity by an extra $\log(1 / \rho)$ factor.

To argue the replicability of the tester when we are in neither the completeness nor the soundness case, we note that the variance is at most $ \log(n_1) G $.
By Chebyshev's inequality and \eqref{eq:na-var-key}, we have that the test statistic will concentration around an interval of size $ \sqrt{ \log(n_1) G } \ll \rho G $ with high constant probability.
Again, combining this with the median trick ensures that $N_a(S_\p, S_\q)$ will lie in an interval (around its expected value) of size $\rho G$ with probability at least $1 - \rho$ (at the cost of increasing the sample complexity by an extra factor of $\log(1/\rho)$).
Conditioned on that, we have that the tester will be replicable as long as the random threshold lies outside this interval of size $\rho G$, which happens with probability at least $1 - \rho$.
We can therefore conclude that the tester is replicable with probability at least $1 - 2\rho$ by the union bound.

Conditioned on that the first-stage testing passes, we hence must have that
$$
\Var[ Z_a(S_\p, S_\q))] \leq \log^2(n_1) \; \Ep[ N_a(S_\p, S_\q) ]
\leq   O\lp( \log^2(n_1) \rp) \; \lp( m^2 / (n_1 n_2) +  m/n_2  \rp).
$$
Besides, since $Z_a(S_\p, S_\q)$ is the average over some statistic that is Lipchitz in the input samples, we also have the trivial variance bound 
$\Var[Z_a(S_\p, S_\q)] \leq O(m)$.
Again, we begin with a quantitative bound that will be useful for both the replicability and correctness analysis:
\begin{align}
\label{eq:z-chebyshev}
\sqrt{
\log^2(n_1)
\min(
m, m^2 / (n_1 n_2) +  m/n_2 )
}
\ll \rho \; \zexpgap.
\end{align}
Again, we proceed by a case analysis.
Suppose the right hand side evaluates to $\eps m$.
We note that $\log(n_1) \sqrt{m} \ll  \rho \eps m$ as long as $ m \gg 
\log^2(n_1) \rho^{-2} \eps^{-2}$.
So \eqref{eq:z-chebyshev} clearly holds in this case.
Suppose that the right hand side evaluates to $m^2 \eps^2 / (n_1 n_2)$.
The case assumption implies that
$m^{1/2} \leq \sqrt{n_1 n_2}$, which further implies that $m \leq n_1 n_2$.
Since we always have $m \gg n_2$, this suggests that $m/n_2$ will be the dominating term on the left hand side.
However, we always have
$\log(n_1) \sqrt{m/ n_2} \ll \rho m^2 \eps^2 / (n_1 n_2)$  as long as 
$m \gg \log^{2/3}(n_1) n_1^{2/3} n_2^{1/3} \rho^{-2/3} \eps^{-4/3}$. This verifies the validity of \eqref{eq:z-chebyshev} in this case.
The last case is when the right hand side evaluates to $m^{3/2} \eps^2 / \sqrt{n_1}{n_2}$. 
In this case, it suffices to show that 
$ \log(n_1) \sqrt{m} \ll \rho m^{3/2} \eps^{2} / \sqrt{n_1 n_2}$, which is true as long as $m \gg \log(n_1) \sqrt{n_1n_2} \eps^{-2} \rho^{-1}$.
This concludes the proof of \eqref{eq:z-chebyshev}.

To argue the correctness of the second stage, we again break into the completeness and the soundness cases.
For convenience, 
denote by $H_E := \zexpgap$ and 
$H_V := \log^2(n_1) \min(
m, m^2 / (n_1 n_2) +  m/n_2 )$.
In the completeness case, by \Cref{lem:independence-expectation-gap}, we have that 
$\Ep[ Z_a(S_\p, S_\q) ] \leq C_{I_1} \; H_E$.
By \eqref{eq:z-chebyshev} and Chebyshev's inequality, we have that
$Z_a(S_\p, S_\q) \leq C_{I_1} \; H_E + O\lp( \sqrt{H_V} \rp) \leq  \lp( C_{I_1} + o(1) \rp) \; H_E $ with high constant probability.
In the soundness case, by \Cref{lem:independence-expectation-gap}, we have that
$\Ep[ Z_a(S_\p, S_\q) ] \geq C_{I_2} \; H_E$.
By \eqref{eq:z-chebyshev} and Chebyshev's inequality, we have that
$Z_a(S_\p, S_\q) \geq C_{I_2} \; H_E - O \lp( \sqrt{H_V} \rp) \leq  \lp( C_{I_2} - o(1) \rp) \; H_E $ with high constant probability.
This shows that the test on $Z_a(S_\p, S_\q)$ is correct with high constant probability.
Combining this with the median trick ensures correctness with probability at least $1 - \rho$ at the cost of increasing the sample complexity by an extra factor of $\log(1/\rho)$.

To argue the replicability of the second stage, we note that $Z_a(S_\p, S_\q)$ must lie in an interval around its expected value with size at most $ \sqrt{H_V} \ll 
\rho H_E $ with high constant probability.
Again, combining this with the median trick ensures that $Z_a(S_\p, S_\q)$ must lie in an interval $L$ of size $\rho H_E$ with probability at least $1 - \rho$ (at the cost of increasing the sample complexity by an extra factor of $\log(1/\rho)$).
Thus, the tester will be replicable as long as the random threshold chosen uniformly random from $[C_{I_1} H_E, C_{I_2} H_E]$ falls outside of this interval $L$.
This happens with probability at least $1 -  \frac{ \rho H_E }{  (C_{I_2} - C_{I_1}) H_E } \geq 1 - O(\rho) $.
We can then conclude that the tester is replicable with probability at least $1 - O(\rho)$ by the union bound.

Lastly, it is clear from the description of $\textsc{RepIndependenceStats}$ that the tester draws $\Theta(m)$ many samples, and $m$ (\Cref{line:m-def} of $\textsc{IndependenceStats}$) is within the sample budget of \Cref{thm:independence}.
This concludes the proof of \Cref{thm:independence}.
\end{proof}

\section{Lower Bounds for Replicable Testing}
\label{sec:lower-bounds}

In this section, we prove our sample complexity lower bounds for replicable uniformity and closeness testing.

\subsection{Poissonization and Internal Randomness Elimination}
\label{sec:poissonization-randomness}
Let $\innerAlg$ be a replicable tester that satisfies the correctness requirement of the corresponding testing problem.
To show a sample complexity lower bound against $\innerAlg$, we  often construct a meta-distribution $\calB_{\xi}$ parametrized by a positive number $\xi \in [0, \eps]$
over potential testing instances.
In particular, $\calB_\xi$ will be constructed such that
$\calB_0$ represents a collection of instances satisfying the property to be tested while $\calB_{\eps}$ represents ones that are ``far'' from satisfying the property.

Our end goal is to show that $\innerAlg$ cannot be $\rho$-replicable under a random choice of $\p \sim \calB_\xi$, where $\xi \sim \mathcal U([0, \eps])$, with non-trivial probability.

There are two common techniques towards the goal. 
Firstly, the tester is usually assumed to take a fixed number of samples from a probability distribution.
Nonetheless, a common practice in distribution testing is to first show lower bounds in the so-called Poisson sampling model, which allows for the more general sampling process for pseudo-distributions, i.e., non-negative measures over the discrete domain, and is often more amenable to analyze.
After that, one can use a reduction-based argument to translate the lower bound back to the standard sampling model.
\begin{definition}[Poisson Sampling]
Given a non-negative measure $\distribution$ over $[n]$ and an integer $m$, the Poisson sampling model samples a number $m' \sim 
\Poi \lp( m \| \distribution \|_1 \rp)$, and draws $m'$ samples from $\distribution / \| \distribution \|_1$. 
Define $T \in \R^n$ to be the random vector where $T_i$ counts the number of element $i$ seen.
We write $\PoiS( m , \distribution )$ to denote the distribution of the random vector $T$.
We say $\innerAlg$ is a Poissonized tester with sample complexity $m$ if it takes as input a sample count vector $T \sim \PoiS( m , \distribution )$.
\label{def:poi_samp}
\end{definition}
Secondly, the tester $\innerAlg$ is in general allowed to use internal randomness. Yet, since we have already fixed the hard instance meta-distribution over the testing instances, a common approach in showing replicability lower bounds is to use a minimax style argument that 
allows us to fix a ``good'' random string $r$ such that the induced deterministic algorithm $\innerAlg(;r)$ enjoys about the same correctness and replicability guarantees under the meta-distribution as the original randomized algorithm. This then allows us to focus on analyzing the replicability of deterministic algorithms under $\distribution \sim \calB_\xi$.
To facilitate the discussion of the minimax argument, we introduce the notion of distributional correctness and replicability.
\begin{definition}[Distributional Correctness/Replicability]
\label{def:distributional-correctness-poisson}
Let $\calB_0, \calB_\eps, \mathcal H$ be meta-distributions over non-negative measures over $[n]$.
Let $\innerAlg$ be a Poissonized tester with sample complexity $m$, \jnew{and $r$ denotes a binary string representing the internal randomness used by $\innerAlg$. }
\begin{itemize}[leftmargin=*]
\item We say  $\innerAlg$ is $\delta$-correct with respect to $\calB_0$ and $ \calB_\eps$ if  $\Pr_{ r, \p \sim \calB_0, T \sim \PoiS(m, \p) }
\lp[  \innerAlg(T) = \text{Accept} \rp] \geq 1- \delta
$ and $\Pr_{ r, \p \sim\calB_{\eps}, T \sim \PoiS(m, \p) }
\lp[  \innerAlg(T) = \text{Accept} \rp] \leq \delta
$. 
\item We say $\innerAlg$ is $\rho$-replicable with respect to $\mathcal H$ if it holds that
$
\Pr_{ r, \p \sim \mathcal H, T,T' \sim \PoiS(m, \p) }
\lp[  \innerAlg(T) \neq \innerAlg(T') \rp] \leq \rho.
$
\end{itemize}

The notions of distributional correctness/replicability for a non-Poissonized tester taking $m$ samples are defined similarly
with the sampling process $S \sim ( \p / \| \p \|_1 )^{\otimes m}$
instead of $T \sim \PoiS(m, \p)$.
\end{definition}

To make our lower bound arguments more modular,
we prove the following meta-lemma that allows us to focus on lower bounds against deterministic algorithm within the Poisson sampling model.
\begin{lemma}
\label{lem:poisson-good-string}
Let $\calB_\xi$ be a meta-distribution parametrized by a number $\xi \in (0, \eps)$, $\mathcal{H}$ be a meta-distribution. Both $\calB_{\xi}$ and $\mathcal{H}$ are over non-negative measures
$\p$ over a finite universe $\mathcal X$ satisfying $\| \p \|_1 \in (0.5, 2)$.
Let $\delta, \rho \in (0, 1/3)$, and $m$ be a positive integer satisfying $m \geq \log(10/\delta) + \log(10/\rho)$. Consider the following two statements:
\begin{itemize}[leftmargin=*]
\item For any deterministic Poissonized tester $\innerAlg$ 
with sample complexity $m$,
if $\innerAlg$ is $\delta$-correct with respect to $\calB_0$ and $\calB_\eps$, then $\innerAlg$ cannot be $\rho$-replicable with respect to $\mathcal{H}$.
\item 
For any randomized tester $\innerAlg$ that consumes  $m' \leq m/10$ samples over $\mathcal X$,
if $\innerAlg$ is $\delta/10$-correct with respect to $\calB_0$ and $\calB_\xi$, then $\innerAlg$ cannot be $\rho/10$-replicable with respect to $\mathcal{H}$.
\end{itemize}
The first statement implies the second statement.
\end{lemma}

\begin{proof}
Let $\innerAlg$ be a randomized tester that consumes $m'$ samples.
Consider the negation of the second statement.
In particular, assume that $\innerAlg$ is $\delta/10$-correct with respect to $\calB_0$ and $\calB_\eps$ as well as $\rho/10$ replicable with respect to $\mathcal{H}$.
We show that this will contradict the first statement.

By Markov's inequality, with probability at least $2/3$ over the choice of the random string $r$, we have that the induced deterministic tester $\innerAlg(;r)$ is $0.3 \delta$-correct with respect to $\calB_0$ and $\calB_\eps$.
Similarly, with probability at least $2/3$ over the choice of $r$, $\innerAlg(;r)$ is $0.3\rho$-replicable with respect to $\mathcal{H}$.
By the union bound, with probability at least $1/3$ over the choice of $r$, 
$\innerAlg(;r)$ is at the same time $0.3\rho$-replicable with respect to $\mathcal{H}$.
and 
$0.3\delta$-correct with respect to $\calB_0$ and $ \calB_\eps$.

We will now convert the tester into a Poissonized one.
In particular, consider the Poissonized tester $\bar \innerAlg$ obtained as follows. 
We first take $k \sim \Poi(m)$ samples from the underlying distribution $\p / \| \p \|_1$.
If $k \geq m'$, we take the first $k$ samples, and feed it to $\innerAlg(;r)$.
If $k < m'$, we simply return reject.
Since we assume $\| \p \|_1 \in (0.5, 2)$ and $m \geq \log(10/\delta) + \log(10/\rho)$, 
it then follows from standard Poisson concentration that 
$k \geq m$ with probability at least $ 1 - \min(\delta, \rho)$.
In particular, this implies that $\bar \innerAlg$ is a Poissonized tester with sample complexity $m$ that is at the same time $0.4\rho$-replicable w.r.t. $\mathcal{H}$ and $0.4\delta$-correct with respect to $\calB_0$ and $\calB_{\eps}$.
This therefore contradicts the first statement of the lemma.
\end{proof}

\subsection{Lower Bound for Replicable Uniformity Testing}
\label{sec:uniformity-lb}

In this subsection, we show the sample complexity lower bound $\tilde{\Omega}\left(\epsilon^{-2}\rho^{-2}+ \sqrt{n}\epsilon^{-2}\rho^{-1}\right)$ for $(n,\epsilon,\rho)$-replicable uniformity testing.
The $\tilde{\Omega}(\epsilon^{-2}\rho^{-2})$ part follows from the lower bound in Lemma 7.2 of \cite{impagliazzo2022reproducibility} for the naive case when $n=2$, 
i.e., distinguishing a fair from a biased coin. 
So we focus on establishing the lower bound of 
$\tilde{\Omega}(\sqrt{n}\epsilon^{-2}\rho^{-1})$. 
{As such, we assume that $\tilde{o}\lp(\sqrt{n} \eps^{-2} \rho^{-1}\rp) = \eps^{-2} \rho^{-2}$, which implies the 
bound $\sqrt{n} \eps^{-2} \rho^{-1} =\tilde{o}\lp( n \eps^{-2}\rp)$, throughout this section.}

We start by describing the hard instance, i.e., the meta-distribution $\mathcal H_U$ over non-negative measures over $[n]$.
\begin{definition}[Uniformity Hard Instance]
\label{def:uniform-hard-instance}
For $\xi \in [0, \eps]$,
we define $\mathcal M_{\xi}$ 
to be the distribution over non-negative measures $\distribution_{\xi}$ defined as follows:
\begin{equation}
    \distribution_{\xi}(i)=
\begin{cases}
\frac{1+\xi}{n} &w.p. \frac{1}{2}\\
\frac{1-\xi}{n} &o.w.
\end{cases}
\end{equation}
The hard instance $\calH_{U}$ for replicable uniformity testing is given by the following sampling process: picking $\xi \sim \mathcal{U}([0, \eps])$, then sample $\distribution_{\xi} \sim \calM_{\xi}$.
\end{definition}

While $\p \sim \calH_{U}$ may not be a distribution (since the total mass may not be $1$), it is with high probability a non-negative measure with mass close to $1$ (say $\norm{\p}{1} \in (0.5, 2)$).
To counter this issue, we will assume the algorithm is Poissonized (i.e., draws $\Poi(m \norm{\p}{1})$ samples).
Due to the fact that $\norm{\p}{1} \leq 2$ and the concentration 
of the Poisson distribution, this does not significantly increase the sample complexity of our algorithm.
Equivalently, we can assume without loss of generality that the algorithm (independently) draws $\Poi(m \p(i))$ samples for each bucket $i$. 

Formally, using \Cref{lem:poisson-good-string} it suffices for us to show lower bounds against any deterministic uniformity tester $\innerAlg$ within the Poisson sampling model.
The main result of this subsection is as follows.
\begin{proposition}
\label{prop:unif-test-poisson-lower-bound}
Let $\mathcal M_\xi$ and $\mathcal{H}_U$ be the meta-distributions defined as in \Cref{def:uniform-hard-instance},  $\innerAlg$ be a deterministic Poissonized tester with sample complexity $m = \tilde o( \sqrt{n} \eps^{-2} \rho^{-1} )$. 
If $\innerAlg$ is $0.1$-correct with respect to $\mathcal M_0$ and $\mathcal M_\eps$, 
then $\innerAlg$ cannot be $\rho \log^{-2} n$-replicable with 
respect to $\mathcal{H}_U$.
\end{proposition}
\Cref{thm:unif-test-lower-bound} then follows immediately from \Cref{prop:unif-test-poisson-lower-bound} and \Cref{lem:poisson-good-string}.
Thus, we focus on the proof of \Cref{prop:unif-test-poisson-lower-bound} in this subsection.

Since $\mathcal A$ is assumed to be a deterministic tester, 
we note that $\Pr_{T, T' \sim \PoiS( m, \distribution ) }\lp[\innerAlg(T) \neq  \innerAlg(T') \rp] \geq 0.1$ holds as long as the acceptance probability of $\innerAlg(T)$ lies in the interval $[1/3, 2/3]$.
For convenience, we define the function 
$\text{Acc}_m( \distribution, \innerAlg):= 
\Pr_{ T \sim \PoiS( m, \distribution ) } \lp[ \innerAlg(T) = \text{Accept}
\rp].
$
It then suffices for us to show 
\begin{align}
\label{eq:acc-close-1/2}
\Pr_{ \distribution  \sim \mathcal H_U }
\lp[ \text{Acc}_m( \distribution, \innerAlg) \in (\log^{-2} n, 1 - \log^{-2} n)
\text{ and }
\| \distribution \|_1 \in (0.5, 2)
\rp] \geq \rho.
\end{align}
Recall that $\mathcal H_U$ is defined to first select $\xi$ randomly from $[0, \eps]$, and then sample from the distribution family $\mathcal M_{\xi}$.
Hence, towards showing \Cref{eq:acc-close-1/2}, 
we first show the intermediate result that the \emph{average} acceptance probability $\Ep_{ \distribution \sim \mathcal M_\xi }
\lp[ \text{Acc}_m( \distribution, \innerAlg) \rp]$ is close to $1/2$ with probability at least $1 - \rho$ over the random choice of $\xi$.
At a high-level, we draws tools from information theory 
to show that the expected acceptance probability function must evaluate to exactly $1/2$ for some $\xi \in (0, \eps)$ and is in general $0.1(\eps \rho)^{-1}$-Lipchitz with respect to the parameter $\xi$.
The argument is similar to the one employed in \cite{liu2024replicable}, and so we defer it to \Cref{app:pf-lem:avg-acceptance-prob-unif}.
The formal statement is given below.
\begin{lemma}
\label{lem:avg-acceptance-prob-unif}Let $\innerAlg$ be a deterministic Poissonized tester that's $0.1$-correct w.r.t. $\mathcal{M}_{0}$ and $\mathcal M_\eps$ and $m = \tilde{o}(\sqrt{n} \eps^{-2} \rho^{-1})$, 
then
$
\Pr_{ \xi \sim \mathcal U([0, \eps]) }
\lp[
\Ep_{ \distribution \sim \mathcal M_{\xi} }
\lp[ 
\text{Acc}_m( \distribution, \innerAlg ) 
\rp] \in (1/3, 2/3) \rp] \geq \rho $.
\end{lemma}
To conclude the proof of \eqref{eq:acc-close-1/2}, 
we would like to then relate the acceptance probability $\text{Acc}_m( \distribution ) $ for a random $\distribution \sim \mathcal M_{\xi}$ to its expected value.
To achieve the goal, \cite{liu2024replicable} exploit the assumption that the underlying tester $\innerAlg$ is symmetric. Namely, the output of the tester is invariant upon relabeling of the domain elements. 

One of our main technical contributions is that we manage to remove this symmetry assumption. 
In particular, we show that even if the underlying tester is not symmetric, 
the acceptance probability $\text{Acc}_m( \distribution )$ will nonetheless satisfy strong 
concentration properties as long as the tester is moderately replicable \jnew{w.r.t. $\mathcal{H}_U.$}
\begin{lemma}[Concentration of Acceptance Probabilities]
    \label{lemma:uniformity-acceptance-concentration}
    Let $m = \tilde{o}(\sqrt{n} \eps^{-2} \rho^{-1})$, $\xi \in (0, \eps)$  
    and $\innerAlg$ 
    be a deterministic tester that is $\log^{-2} n$-replicable with respect to $\mathcal{H}_U$ \footnote{
    \snew{
    We note that by analyzing the eigenvalues of this random walk directly, we can remove the requirement that 
    $\innerAlg$ is at least $\log^{-2} n$-replicable. We will include this argument in a future version of this paper.}}.
    Then it holds
    \begin{equation*}
        \Pr_{\distribution  \sim \mathcal M_{\xi}} \left( \left| 
        \text{Acc}_m( \distribution, \innerAlg )
        -
        \Ep_{ \distribution' \sim \mathcal M_{\xi} }
        \lp[ \text{Acc}_m( \distribution', \innerAlg ) \rp]
         \right| > \frac{1}{4} \right) \leq \frac{1}{2} \text{.}
    \end{equation*}
\end{lemma}
To show \Cref{lemma:uniformity-acceptance-concentration}, we construct a random walk on the sample space whose stationary distribution is the same as $T \sim \p$, where $\p \sim \mathcal M_\xi$.

\begin{definition}[Sample Random Walk]
    \label{def:sample-rw}
    Let $\mathcal M$ be a meta-distribution over non-negative measures over $[n]$, and $m \in \mathds Z_+$.
    The Sample Random Walk $\RW_{m, \mathcal M }$ is defined on the graph whose vertex set is $\N^{n}$ (where each vertex corresponds to a sample count vector $T$) and transitions $(T_1, T_2)$ are defined by the conditional distribution of $T_2$ given $T_1$ induced by the joint distribution given by the following process:
    \begin{enumerate}
    \item Choose $\distribution \sim \mathcal M$.
    \item $T_1, T_2$ are sampled independently from $\PoiS(m, \distribution)$.
    \end{enumerate}
    Moreover, 
    for a sample count vector $T$,
    we denote by $\RW_{m,\mathcal M}^k(T)$ the random variable representing the outcome after $k$ steps of the random walk $\RW_{m, \calM}$ from $T$.
    For a non-negative measure $\p$ over $[n]$,
    we denote by $\RW_{m, \mathcal M}^k(\distribution)$ the distribution of $\RW_{m,\mathcal M}^k(T)$, where $T \sim \PoiS(m, \distribution)$.
\end{definition}

For simplicity, we define $\RW_{m, \xi} := \RW_{m, \calM_{\xi}}$ where $\calM_{\xi}$ is the meta-distribution given in \Cref{def:uniform-hard-instance}.
The random walk turns out to mix very rapidly within only polylogarithmic in $n$ many steps. 
This allows us to approximate the stationary distribution by $\RW_{m, \xi}^k( \distribution )$ for some $k = \polylog(n)$.
As a result, 
we can write
\begin{align*}
&
\lp| \Ep_{ T \sim \PoiS(m, \distribution) }
\lp[  \innerAlg( T ) \rp]
-
\Ep_{ \distribution' \sim \mathcal M_{\xi} }
\lp[ 
\Ep_{ T \sim \PoiS(m, \distribution') }
\lp[  \innerAlg( T ) \rp]
\rp] \rp|\\
&\approx
\lp|
\Ep_{ T \sim \PoiS(m,\distribution) }
\lp[  
\innerAlg( T ) 
\rp] 
- 
\Ep_{ T' \sim \RW_{m, \xi}^k( \distribution ) }\lp[ \innerAlg(T') \rp] \rp|.
\end{align*}
Since $\innerAlg$ is replicable with respect to $\mathcal{H}_U$, we can use the triangle inequality and some simple algebraic manipulation to further bound the above by the sum of the terms
\begin{align}
\label{eq:rw-diff}
\Pr_{ T \sim \RW_{m, \xi}^{i-1}( \distribution )  \, , \,
T' \sim \RW_{m, \xi}^i(\p)
}
\lp[  
\innerAlg( T )
\neq 
\innerAlg( T' )
\rp] 
\, ,
\end{align}
where $i \in [k]$.
While it is challenging to establish a uniform bound on the terms for an arbitrary non-negative measure $\distribution$, it turns out this is not so hard for an ``average'' $\distribution \sim \mathcal M_\xi$. 
At a high level, if we consider the expected value of the disagreement probability in \Cref{eq:rw-diff} over $\distribution \sim \mathcal M_{\xi}$, the term simplifies to
$
\Pr_{ T \sim \pi_\xi, 
T' \sim \RW_{m, \xi}(T)}
\lp[  
\innerAlg( T )
\neq 
\innerAlg( T' )
\rp] 
= 
\Pr_{ 
\distribution \sim \mathcal M_{\xi},
T, T' \sim \PoiS(m,\distribution)}
\lp[  
\innerAlg( T )
\neq 
\innerAlg( T' )
\rp] \, , 
$
where the equality follows from the definition of the stationary distribution $\pi_{\xi}$ of the random walk.
Therefore, the expected value of the disagreement probability cannot be too large as long as the tester $\innerAlg$ is moderately replicable w.r.t. $\mathcal{H}_U$.
The formal statement is given below.
\begin{lemma}[Indistinguishability of Random Walk Step]
\label{lem:random-walk-indistinguishability}
Let $\innerAlg$ be a deterministic uniformity tester and $\distribution \sim \mathcal M_\xi$.
Define $\kappa := 
\Pr_{ 
\distribution \sim \mathcal M_{\xi},
T, T' \sim \PoiS(m,\distribution)}
\lp[  
\innerAlg( T )
\neq 
\innerAlg( T' )
\rp]$.
With probability at least $1/2$, 
it holds that
$
    \sum_{i=1}^k
\Pr_{ T \sim \RW_{m, \xi}^{i-1}( \distribution )  \, , \,
T' \sim \RW_{m, \xi}(T)
}
\lp[  
\innerAlg( T )
\neq 
\innerAlg( T' )
\rp]  
< 2 k \kappa \, ,
$
where the randomness is over choice of $\distribution \sim \mathcal M_\xi$.
\end{lemma}
\begin{proof}
We first show that 
\begin{equation}
\label{eq:before-markov}
    \sum_{i=1}^k
\Ep_{ \p \sim \mathcal M_\xi }
\lp[
\Pr_{ T \sim \RW_{m, \xi}^{i-1}( \distribution )  \, , \,
T' \sim \RW_{m, \xi}(T)
}
\lp[  
\innerAlg( T )
\neq 
\innerAlg( T' )
\rp]
\rp]
< k \kappa \, ,
\end{equation}
After that, the lemma will follow from Markov's inequality.

Note that if we sample $\p \sim \mathcal M_\xi$, and then $T \sim \PoiS(m, \p)$, we obtain exactly the stationary distribution 
of the random walk.
Thus, the distribution of $T \sim \RW_{m, \xi}^{i-1}( \distribution )  \, , \,
T' \sim \RW^i_{m, \xi}(\p)$ is equivalent as
$\p \sim \mathcal M_\xi, T \sim  \PoiS(m, \p) \, , \,
T' \sim \RW_{m, \xi}(T)$.
If we focus on just the joint distribution of $T, T'$, by the definition of $\RW_{m, \xi}$, this is the same as
$ T, T' \sim \PoiS(m, \p) $, where $\p \sim \mathcal M_\xi$.
This therefore gives rise to the identity
$\Ep_{ \p \sim \mathcal M_\xi }
\lp[
\Pr_{ T \sim \RW_{m, \xi}^{i-1}( \distribution )  \, , \,
T' \sim \RW_{m, \xi}(T)
}
\lp[  
\innerAlg( T )
\neq 
\innerAlg( T' )
\rp]
\rp]
= \Pr_{ \p \sim \mathcal M_\xi, 
T, T' \sim \PoiS( m, \p )
} \lp[  \innerAlg( T )
\neq 
\innerAlg( T' ) \rp] = \kappa.
$
Summing over all $i$ then concludes the proof of \Cref{eq:before-markov} as well as \Cref{lem:random-walk-indistinguishability}.
\end{proof}
The proof of \Cref{lemma:uniformity-acceptance-concentration} then largely follows from
\Cref{lem:random-walk-indistinguishability} and the fact that the random walk is fast mixing.

\begin{restatable}{theorem}{UnifRWMixingTime}
    \label{thm:unif-rw-mixing-delta}
    The random walk $\RW_{m, \xi}$ has mixing time
$\tau(\delta) = O(\log(n/\delta))$.
\end{restatable}

For simplicity, we identify the $\Accept$ outcome with $1$ so that $\Acc_m(\distribution, \innerAlg) = \Pr_{T \sim \PoiS(m, \distribution)} [\innerAlg(T) = \Accept] = \Ep_{T \sim \PoiS(m, \distribution)} [\innerAlg(T)]$. To show \Cref{thm:unif-rw-mixing-delta}, we analyze the transition probability of the random walk $\RW_{m, \xi}$.
We first note that
drawing $\PoiD{m}$ samples from $\distribution \sim \calM_{\xi}$ is equivalent as drawing $\PoiD{m \distribution_{i}}$ samples from each bucket independently where $\p_i \in \set{\frac{1 + \xi}{n}, \frac{1 - \xi}{n}}$ is the mass of bucket $i$ under the measure $\distribution$.
Since the observed count of each bucket $i \in [n]$ is independent, we may decompose the random walk $\RW_{m, \xi}$ as a product of $n$ independent random walks.

\begin{definition}
    \label{def:unif-coord-rw}
    The Coordinate Sample Random Walk $\RW_{m, \xi, i}$ is defined on the graph whose vertex set is $\N$  and transitions $(T_1[i], T_2[i])$ are defined by the conditional distribution of $T_2[i]$ given $T_1[i]$ induced by the joint distribution given by the following process:
    \begin{enumerate}
        \item Choose $\p_i \in \mathcal{U}\lp(\lp\{\frac{1 + \xi}{n}, \frac{1 - \xi}{n}\rp\}\rp)$.
        \item $T_1[i], T_2[i]$ are sampled independently from $\PoiS(m, \p_{i})$.
    \end{enumerate}
    Given a sample count $T[i]$, we denote $\RW_{m, \xi, i}^{k}(T[i])$ the random variable representing the outcome after $k$ steps of random walk from $T$.
    For $\p_{i} \geq 0$, we denote by $\RW_{m, \xi, i}^{k}(\p_{i})$ the distribution of $\RW_{m, \xi, i}^{k}(T[i])$ where $T[i] \sim \PoiS(m, \p_{i})$.
\end{definition}

By independence, we have that 
\begin{equation*}
    \RW_{m, \xi} = \prod_{i = 1}^{n} \RW_{m, \xi, i} \text{.}
\end{equation*}

For a sample $T$, let $T[i]$ denote the empirical frequency of the $i$-th bucket in $T$.
Let $T_{1} \sim \RW_{m, \xi, i}(T_0)$.
We can write the joint distribution of $T_{0}, T_{1}$ as

\begin{align*}
    \Pr(T_0[i] = a, T_1[i] = b) &= \frac{\left(e^{-2m(1 + \xi)/n} \frac{((1 + \xi)m/n)^{a + b}}{a!b!} + e^{-2m(1 - \xi)/n} \frac{((1 - \xi)m/n)^{a + b}}{a!b!}\right)}{2} \\
    &= \frac{1}{2 a! b!} e^{-2m/n} \left( \frac{m}{n} \right)^{a + b} \left(e^{-2\xi m/n} (1 + \xi)^{a + b} + e^{2 \xi m/n} (1 - \xi)^{a + b}\right).
\end{align*}

Furthermore, we have that

\begin{align*}
    \Pr(T_0[i] = a) &= 
    \sum_{b = 0}^{\infty} \frac{\left(e^{-2(1 + \xi)m/n} \frac{((1 + \xi)m/n)^{a + b}}{a!b!} + e^{-2(1 - \xi)m/n} \frac{((1 - \xi)m/n)^{a + b}}{a!b!}\right)}{2} \\
    &= 
    \frac{e^{-(1 + \xi)m/n} ((1 + \xi)m/n)^{a} + e^{-(1 - \xi)m/n} ((1 - \xi)m/n)^{a}}{2a!} \\
    &= \frac{1}{2a!} e^{-m/n} \left( \frac{m}{n} \right)^{a} \left( e^{-\xi m / n} (1 + \xi)^{a} + e^{\xi m / n} (1 - \xi)^{a} \right) \;.
\end{align*}
Combining the two gives that the probability of the transition $P(a, b) := \Pr(T_1[i] = b | T_0[i] = a)$ is

\begin{align*}
    P(a, b) &= \frac{\Pr(T_1[i] = b, T_0[i] = a)}{\Pr(T_0[i] = a)} \\
    &= \frac{1}{b!} e^{-m/n} \left( \frac{m}{n} \right)^{b} \left( \frac{e^{-2 \xi m/n} (1 + \xi)^{a + b} + e^{2 \xi m/n} (1 - \xi)^{a + b}}{e^{-\xi m/n} (1 + \xi)^{a} + e^{\xi m/n} (1 - \xi)^{a}} \right) \text{.}
\end{align*}
This defines the random walk $\RW_{m,\xi, i}$ for each $i \in [n]$ with transition probabilities given above.
Given $\RW_{m, \xi, i}$ for all $i$, we can write the transition probability of $\RW_{m, \xi}$ from $a = (a_1, \dotsc, a_{n})$ to $b = (b_1, \dotsc, b_n)$ as

\begin{align*}
    \Pr(\RW_{m, \xi}(a) = b) &= \prod_{i = 1}^{n} \Pr\left( T_1[i] = b_i | T_0[i] = a_i \right) \\
    &= e^{-m} \prod_{i = 1}^{n} \frac{1}{b_i!}  \left( \frac{m}{n} \right)^{b_i} \left( \frac{e^{-2 \xi m/n} (1 + \xi)^{a_i + b_i} + e^{2 \xi m/n} (1 - \xi)^{a_i + b_i}}{e^{-\xi m/n} (1 + \xi)^{a_i} + e^{\xi m/n} (1 - \xi)^{a_i}} \right) \text{.}
\end{align*}
In particular, the stationary distribution of $\RW_{m, \xi}$ is the 
vector $\pi \in [m]^{n}$ given by
\begin{align*}
    \pi((a_1, \dotsc, a_n)) &= \prod_{i = 1}^{n} \Pr(T_0[i] = a_i) \\
    &= e^{-m} \prod_{i = 1}^{n} \frac{1}{2a_i!} \left( \frac{m}{n} \right)^{a_i} \left( e^{-\xi m / n} (1 + \xi)^{a_i} + e^{\xi m / n} (1 - \xi)^{a_i} \right) \text{.}
\end{align*}

It is not hard to see that our random walk is ergodic and reversible.
\begin{lemma}
    \label{lemma:markov-chain-reversible}
    The random walk $\RW_{m, \xi}$ is ergodic and reversible.
\end{lemma}

\begin{proof}
    The random walk $\RW_{m, \xi}$ is ergodic since every transition is possible (including self-loops). 
    Furthermore, $\RW_{m, \xi}$ is reversible since $\pi(i) P(i, j)$ is a joint distribution that is the same as $\pi(j) P(j, i)$.
\end{proof}

We proceed to show that $\RW_{m, \xi, i}$ mixes rapidly.

\begin{lemma}
    \label{lemma:coord-rw-mix}
    Suppose $m = o(n/\eps^2)$.
    The random walk $\RW_{m, \xi, i}$ has mixing time $\tau(0.04) \leq 2$.
\end{lemma}

\begin{proof}
    Note that the random walk $\RW_{m,\xi, i}$ has transition probabilities from $Y_0 \geq 0$ to $Y_1 \geq 0$ given by the conditional distribution induced by the following joint distribution.
    \begin{enumerate}
        \item Let $X \sim \mathcal{U}(\set{0, 1})$ be a uniformly random bit.
        \item Independently sample $Y_0, Y_1 \sim \PoiD{m(1 - \xi)/n}$ if $X = 0$ and otherwise sample $Y_0, Y_1 \sim \PoiD{m(1 + \xi)/n}$ if $X = 1$.
    \end{enumerate}

    A useful fact is that the total variation distance between $\PoiD{m(1 - \xi)/n}$ and $\PoiD{m(1 + \xi)/n}$ is small.
    \begin{claim}
        \label{clm:poisson-tvd}
        Let $\lambda_{1} > \lambda_{2} > 0$.
        Let $X \sim \PoiD{\lambda_{1}}$ and $Y \sim \PoiD{\lambda_{2}}$.
        Then
        \begin{equation*}
            \tvd{X, Y} \leq \sqrt{\frac{(\lambda_{1} - \lambda_{2})^{2}}{2 \lambda_{2}}} \text{.}
        \end{equation*}
    \end{claim}

    \begin{proof}
        We begin by bounding the KL-divergence as
        \begin{equation*}
            D_{\KL} = \lambda_{1} \log \frac{\lambda_{1}}{\lambda_{2}} + \lambda_{2} - \lambda_{1} \leq \lambda_{1} \left( \frac{\lambda_{1} - \lambda_{2}}{\lambda_{2}} \right) + \lambda_{2} - \lambda_{1} \leq \frac{(\lambda_{1} - \lambda_{2})^{2}}{\lambda_{2}}
        \end{equation*}
        where we have used $\log x \leq x - 1$ for $x > 0$.
        Now, using Pinsker's inequality, we can bound
        \begin{equation*}
            \tvd{X, Y} \leq \sqrt{\frac{(\lambda_{1} - \lambda_{2})^{2}}{2 \lambda_{2}}} \text{.}
        \end{equation*}
        This concludes the proof of \Cref{clm:poisson-tvd}.
    \end{proof}

    We handle the sublinear and superlinear cases separately.
    
    \paragraph{Sublinear Case: $m \leq n$.}
    Note that \Cref{clm:poisson-tvd} implies that the total variation distance between $Z_0 \sim \PoiD{m(1-\xi)/n}$ and $Z_1 \sim \PoiD{m(1+\xi)/n}$ is at most
    \begin{equation*}
        \sqrt{\frac{(2m\xi/n)^{2}}{2m(1-\xi)/n}} \leq \sqrt{\frac{2m\xi^2/n}{1-\xi}} \leq 2 \xi \;,
    \end{equation*}
    where in the final inequality we have used $m/n \leq 1$ and $1 - \xi > 0.5$.
    
    Now consider a step of the random walk from initial state $Y_0 = \ell$.
    The distribution of $Y_1$ is given by the mixture of two Poisson distributions
    \begin{equation*}
        \Pr(Y_1 = k | Y_0 = \ell) = \Pr(X = 0 | Y_0 = \ell) \Pr(Z_0 = k) + \Pr(X = 1 | Y_0 = \ell) \Pr(Z_1 = k) \text{.}
    \end{equation*}
    The total variation distance between this distribution and the stationary distribution $\pi$ is at most
    \begin{align*}
        \frac{1}{2} \sum_{k} \left| \Pr(Y_1 = k | Y_0 = \ell) - \pi(k) \right| &\leq 2 \left| \Pr(X = 0 | Y_{0} = \ell) - \frac{1}{2} \right| \xi \leq 2 \xi \text{.}
    \end{align*}
    Now consider a step from $Y_{1} = k$.
    By the total variation distance bound, we can conclude that an algorithm cannot distinguish $X = 0$ with advantage better than $2 \xi < 0.2$.
    Therefore, we can bound $0.3 \leq \Pr(X = 0|Y_{1} = k) \leq 0.7$, otherwise the algorithm that returns $X$ with this conditional probability is a distinguisher.
    From our previous calculation we conclude that after two steps, the random walk mixes to within $0.4 \xi < 0.04$ of the stationary distribution.

    \paragraph{Superlinear Case: $n \leq m = \litO{\frac{n}{\eps^2}}$}

    Following similar arguments as in the sublinear case, 
    the total variation distance between $Z_0, Z_1$ 
    is at most $\litO{1} < 0.1$.
    As in the sublinear case, no algorithm can distinguish 
    $X = 0$ with advantage better than the total variation 
    distance $0.1$.
    Since $\Pr(X = 0|Y_1 = k) \leq 0.6$, 
    we can conclude that the random walk mixes within $0.01$ 
    of the stationary distribution in two steps.
    
 \smallskip   
 
    This concludes the proof of \Cref{lemma:coord-rw-mix}.
\end{proof}

Given \Cref{lemma:coord-rw-mix}, we can bound the relaxation time of each coordinate random walk via \Cref{thm:mixing-time-relaxation-time}.
In particular, we have that for $\RW_{m,\xi, i}$, $t_{\rel} \leq \frac{\tau(\delta)}{\log(1/2 \delta)} + 1$.
Combining this with \Cref{lemma:coord-rw-mix} gives that
\begin{equation*}
    t_{\rel} \leq \frac{\tau(0.04)}{\log(1/0.08)} + 1 \leq \frac{2}{\log(1/0.08)} + 1 = O(1) \text{.}
\end{equation*}

We are now ready to bound the mixing time of the product random walk $\RW_{m, \xi}$.

\begin{lemma}
    \label{lemma:coord-rw-mixing-delta}
    Let $m = \litO{\frac{n}{\eps^2}}$.
    Let $\gamma(x) \sim \PoiD{(1 + \xi)m/n}$ or $\gamma(x) \sim \PoiD{(1 - \xi)m/n}$ denote the initial distribution.
    Then, under either initial distribution $\gamma$,  $\RW_{m, \xi, i}$ has mixing time 
$\tau_{i}(\delta) = O(\log(1/\delta))$.
\end{lemma}

\begin{proof}
    Consider a coordinate random walk $\RW_{m,\xi, i}$.
    Let $P$ denote the transition matrix and $P^{t}$ denote the transition matrix after $t$ steps.
    Let $\pi$ denote the stationary distribution.
    Recall that we have shown that $\RW_{m,\xi, i}$ has constant relaxation time and therefore constant absolute spectral gap $\lambda_{*}$.
    Given either initial distribution $\gamma(x)$, our goal is to bound the quantity
    \begin{equation*}
        \sum_{y} \left| \left( \sum_{x} \gamma(x) P^{t}(x, y) \right) - \pi(y) \right| \text{.}
    \end{equation*}
    We begin with the following inequality that follows from the proof of Theorem 12.5 of \cite{LevinPeres}.
    For any two states $x, y$, 
    \begin{equation*}
        \left| \frac{P^{t}(x, y)}{\pi(y)} - 1 \right| \leq \frac{\lambda_{*}^{t}}{\sqrt{\pi(x) \pi(y)}}\text{.}
    \end{equation*}
    Multiplying both sides by $\pi(y) \gamma(x)$ we obtain the inequality
    \begin{equation*}
        \left| \gamma(x) P^{t}(x, y) - \gamma(x) \pi(y) \right| \leq \frac{\lambda_{*}^{t} \gamma(x) \sqrt{\pi(y)}}{\sqrt{\pi(x)}}\text{.}
    \end{equation*}

    The next claim bounds the ratio between $\gamma(x)$ and $\pi(x)$.
    \begin{claim}
        \label{clm:gamma-pi-diff}
        For any $x \geq 0$,
        \begin{equation*}
            \frac{\gamma(x)}{\pi(x)} \leq 2 \text{.}
        \end{equation*}
    \end{claim}

    \begin{proof}
        Let $Z_0 \sim \PoiD{(1 - \xi)m/n}$ and $Z_1 \sim \PoiD{(1 + \xi)m/n}$.
        Let $\lambda_{0}, \lambda_{1}$ denote the means of $Z_0, Z_1$ respectively.
        First, we show that $\gamma(x)/\pi(x)$ is bounded when $\gamma \sim Z_0$.
        \begin{align*}
            \frac{\Pr(Z_0 = x)}{(\Pr(Z_0 = x) + \Pr(Z_1 = x))/2} &= \frac{2 \Pr(Z_0 = x)}{\Pr(Z_0 = x) + \Pr(Z_1 = x)} \\
            &\leq \frac{2 \Pr(Z_0 = x)}{\Pr(Z_0 = x)} = 2.
        \end{align*}
        A similar argument holds for $\gamma \sim Z_1$.
        This concludes the proof of \Cref{clm:gamma-pi-diff}.
    \end{proof}
    
    Continuing from our previous calculation, we obtain
    \begin{equation*}
        \left| \gamma(x) P^{t}(x, y) - \gamma(x) \pi(y) \right| \leq \lambda_{*}^{t} \sqrt{2 \gamma(x) \pi(y)}\text{.}
    \end{equation*}
    Summing over $x$, applying the triangle inequality and noting that $\sum_{x} \gamma(x) = 1$, we now have
    \begin{equation*}
        \left| \left( \sum_{x} \gamma(x) P^{t}(x, y) \right) - \pi(y) \right| \leq \lambda_{*}^{t} \sqrt{2 \pi(y)} \sum_{x} \sqrt{\gamma(x)}\text{.}
    \end{equation*}

    For the remainder of the proof, we will consider the sub-linear and super-linear cases separately.
    \paragraph{Sublinear Case: $m \leq n$.}
    Since $m \leq n$, in both cases the Poisson distribution has parameter $\lambda \leq (1 + \xi) \leq 1.1$. 
    By standard Poisson concentration, for both $i \in \set{0, 1}$ 
    \begin{equation*}
        \Pr(Z_{i} > \lambda + t) < e^{- t^2/2(\lambda + t)} = e^{-\Omega(t)} \text{.}
    \end{equation*}
    In particular, $\gamma(x + 2) < e^{-\Omega(x)}$ for all $x$.
    Therefore,
    \begin{align*}
        \sum_{x = 0}^{\infty} \sqrt{\gamma(x)} \leq \sum_{x = 0}^{C} \sqrt{\gamma(x)} + \sum_{x = C}^{\infty} \sqrt{\gamma(x)} = C + \sum_{x = C}^{\infty} e^{-\Omega(x/2 - 1)} = O(1)
    \end{align*}
    for some large enough absolute constant $C$.
    Here, we observe that for $x \geq C$, $e^{-\Omega(x/2 - 1)} < e^{-x/C'}$ for some constant $C'$ so that the second term is an infinite geometric series with ratio $e^{-1/C'} < 1$.
    Similarly, we can bound $\sum_{y = 0}^{\infty} \pi(y) = O(1)$.
    Thus, to conclude we sum over $y$ and note that
    \begin{align*}
        \sum_{y} \left| \left( \sum_{x} \gamma(x) P^{t}(x, y) \right) - \pi(y) \right| &\leq \sqrt{2} \lambda_{*}^{t} \sum_{y} \sqrt{\pi(y)} \sum_{x} \sqrt{\gamma(x)} = \bigO{\lambda_{*}^{t}} \text{.}
    \end{align*}
    In particular, from the initial distribution $\gamma$, the random walk $\RW_{\xi}$ mixes to $\delta$ in time $O(\log(1/\delta))$.

    \paragraph{Superlinear Case: $n < m \leq \litO{\frac{n}{\eps^2}}$.}
    Recall that $x, y \sim \PoiD{\lambda}$ for $\lambda \in \set{(1 + \xi)m/n, (1 - \xi)m/n}$.
    Using standard Poisson concentration (e.g., \Cref{lemma:poisson-concentration}) and noting that $\lambda > 1 - \xi$, we observe that for any $x$, we have $\gamma(x) < e^{-\Omega(|x - \lambda|)}$.
    As in the sublinear case, we begin by bounding $\sum \sqrt{\gamma(x)}$ for initial distribution $\gamma$.
    For sufficiently large constant $C$, we can bound
    \begin{align*}
        \sum_{x = 0}^{\infty} \sqrt{\gamma(x)} &\leq \sum_{|x - \lambda| \leq C} \sqrt{\gamma(x)} + \sum_{|x - \lambda| > C} \sqrt{\gamma(x)} = O(1) \text{.}
    \end{align*}
    As above, we observe that for large enough $C$, $\sum_{|x - \lambda|>C} \sqrt{\gamma(x)}$ can be decomposed into two geometric series with ratio strictly less than $1$.
    Since $\pi$ is a mixture of both $\gamma$, we have that $\sum_{y} \sqrt{\pi(y)} = O(1)$ as well.
    The conclusion then follows as in the sublinear case.

    This concludes the proof of \Cref{lemma:coord-rw-mixing-delta}.
\end{proof}

\begin{proof}[Proof of \Cref{thm:unif-rw-mixing-delta}]
The proof follows immediately from     
\Cref{lemma:coord-rw-mixing-delta} and \Cref{lemma:product-random-walk-mixing-time}.
\end{proof}

We are now ready to show that the acceptance probability of the algorithm on samples drawn from $\PoiS(m, \distribution)$ for a random $\distribution \sim \calM_{\xi}$ is well concentrated assuming that the algorithm is sufficiently replicable in terms of the mixing time of the random walk.

\begin{lemma}
    \label{lemma:S-concentration-for-good-xi}
    Let $K = \tau(0.01)$ and $\xi \in [0, \eps]$.
    Suppose $\innerAlg$ is $\frac{1}{10K}$-replicable with respect to $\mathcal{H}_U$.
    Then,
    \begin{equation*}
        \Pr_{\distribution \sim \calM_{\xi}} \left( \left| \Ep_{T \sim \PoiS(m, \distribution)} \left[ \innerAlg(T) \right] - \Ep_{\distribution' \sim \calM_{\xi}, T' \sim \PoiS(m, \distribution')} \left[ \innerAlg(T') \right] \right| > \frac{1}{4} \right) \leq \frac{1}{2} \text{.}
    \end{equation*}
\end{lemma}

\begin{proof}
    Consider the following sampling process:
    \begin{enumerate}
        \item Sample $\distribution \sim \calM_{\xi}$.
        \item Sample $T_0 \sim \PoiS(m, \distribution)$.
        \item For $1 \leq k \leq K := \tau(0.01) = O(\log n)$, sample $T_{k} \sim \RW_{m, \xi}(T_{k - 1})$.
    \end{enumerate}
    From \Cref{lem:random-walk-indistinguishability} we know that on average over $\distribution \sim \calM_{\xi}$, $\innerAlg(T_0) = \innerAlg(T_1) = \dotsc = \innerAlg(T_{K})$ with probability at least $0.9$.
    By Markov's inequality, for $\frac{1}{2}$-fraction of $S$, we have that $\innerAlg(T_0) = \innerAlg(T_1) = \dotsc = \innerAlg(T_k)$ with probability at least $0.8$.

    We now argue that the distribution of $T_{k}$ is $0.01$-close to the distribution of $T$ drawn from the stationary distribution $\pi$ in total variation distance.
    Since $\RW_{m, \xi}$ is a product random walk, \Cref{lemma:product-random-walk-mixing-time} implies that it suffices to argue that each coordinate random walk $\RW_{m,\xi, i}$ mixes to within $\frac{0.01}{n}$ of the stationary distribution on that coordinate in $\tau_{i}(0.01/n)$ steps.
    Fix a coordinate $i$.
    Note that $T_0[i] \sim \PoiD{(1 + \xi)m/n}$ or $\PoiD{(1 - \xi)m/n}$.
    In either case, the initial distribution satisfies the assumptions of \Cref{lemma:coord-rw-mixing-delta}, so that after $O(\log n)$ steps, the random walk $\RW_{m,\xi, i}$ mixes to within $\frac{0.01}{n}$ of the stationary distribution.
    Given that $T_{k}$ is close to the stationary distribution, the data processing inequality says that the probability of acceptance under either distribution cannot differ by more than $0.01$.
    
    Here, recall that $T' \sim \pi$ is given by $T' \sim \distribution$ where $\distribution \sim \calM_{\xi}$.
    In particular, since the stationary distribution is exactly the probability of a sample drawn from random $\distribution \sim \calM_{\xi}$, we have that for $\frac{1}{2}$-fraction of $S$,
    \begin{align*}
        \left| \Ep_{\substack{\distribution \sim \calM_{\xi}, \\ T_{0} \sim \distribution}} (\innerAlg(T_{0})) - \Ep_{T' \sim \pi}(\innerAlg(T')) \right| &\leq \left| \Ep_{\substack{\distribution \sim \calM_{\xi}, \\ T_{0} \sim \distribution}} (\innerAlg(T_{0})) - \Ep_{T_{K} \sim \RW_{m, \xi}^{K}(T_0)} (\innerAlg(T_{K})) \right| \\
        &+ \left| \Ep_{T_{K} \sim \RW_{m, \xi}^{K}(T_0)} (\innerAlg(T_{K})) - \Ep_{T' \sim \pi}(\innerAlg(T')) \right| \\
        &\leq 0.2 + 0.01 \\
        &\leq \frac{1}{4} \text{.}
    \end{align*}
\end{proof}

We are now ready to prove \Cref{lemma:uniformity-acceptance-concentration}.

\begin{proof}[Proof of \Cref{lemma:uniformity-acceptance-concentration}]
    From \Cref{thm:unif-rw-mixing-delta} we note that the mixing time $K = \tau(0.01) = O(\log n)$.
    Then, since we assume that $\innerAlg$ is $(\log n)^{-2}$-replicable with respect to $\mathcal{H}_U$, we have $\innerAlg$ is $\frac{1}{10 K}$-replicable w.r.t. $\mathcal{H}_U$.
    Applying \Cref{lemma:S-concentration-for-good-xi}, we obtain the desired result noting that $\Acc_m(\distribution, \innerAlg) = \Ep_{T \sim \PoiS(m, \distribution)} [\innerAlg(T)]$.
\end{proof}

We are then ready to conclude the proof of \Cref{prop:unif-test-poisson-lower-bound}.
\begin{proof}[Proof of \Cref{prop:unif-test-poisson-lower-bound}]
By \Cref{lem:avg-acceptance-prob-unif}, 
with probability at least $1 - \Omega(\rho)$ over the choice of $\rho$, we have that
\begin{align}
\label{eq:good-xi}
\Ep_{ \p \sim \mathcal M_{\xi} }
\lp[ \Acc_m( \p, \innerAlg) \rp] \in (1/3, 2/3).    
\end{align}
Conditioned on some $\xi$ satisfying the above, 
we claim that we must have 
\begin{align}
\label{eq:disagreement-condition-xi}
\Ep_{ \p \sim \mathcal M_\xi }
\lp[ 
\Pr_{ T, T' \sim\PoiS(m,\p) }
\lp[  \innerAlg(T) \neq \innerAlg(T') \rp]
\rp] \geq \log^{-2} n.
\end{align}
We will proceed by a proof of contradiction.
Assume that the opposite of \Cref{eq:disagreement-condition-xi} holds.
In that case,
\Cref{lemma:uniformity-acceptance-concentration} 
becomes applicable, which gives that
\begin{align}
\label{eq:good-p}
\Pr_{ \p \sim \mathcal M_{\xi}  }
\lp[ 
\lp| 
\Acc_m( \distribution, \innerAlg )
- \Ep_{ \p \sim \mathcal M_{\xi} }
\lp[ \Acc_m( \p, \innerAlg) \rp]
\rp| > 1/4
\rp] \leq 1/2.
\end{align}
Combining \eqref{eq:good-xi} and \eqref{eq:good-p} then gives that
$
\Acc_m( \distribution, \innerAlg ) \in (1/3 - 1/4, 2/3 + 1/4) 
$
with probability at least $1/2$ when we choose $\p \sim \mathcal M_\xi$.
Conditioned on such a $\p$,
we immediately have that
$  
\Pr_{ T, T' \PoiS(m,\p) }[ 
\innerAlg(T) \neq \innerAlg(T')
] \geq \Omega(1).
$
This therefore implies that
$
\Ep_{ \p \sim \mathcal M_\xi }
\lp[ 
\Pr_{ T, T' \sim \PoiS(m,\p) }[ 
\innerAlg(T) \neq \innerAlg(T')
]
\rp] \geq \Omega(1) \, ,
$
which contradicts the assumption 
$\Ep_{ \p \sim \mathcal M_\xi }
\lp[ 
\Pr_{ T, T' \sim \PoiS(m,\p) }[ 
\innerAlg(T) \neq \innerAlg(T')
]
\rp] \leq \log^{-2}(n)$.
This concludes the proof of \eqref{eq:disagreement-condition-xi}.

Recall that the meta-distribution $\mathcal H_U$ is precisely the distribution of $\p$ if one first chooses $\xi$ from $[0, \eps]$ uniformly at random, and then chooses $\p \sim \mathcal M_{\xi}$.
Thus, combining \eqref{eq:good-xi} and \eqref{eq:disagreement-condition-xi} gives that
\begin{align*}
\Pr_{ \p \sim \mathcal H_U  }
\lp[ 
\Pr_{ T, T' \sim \PoiS(m,\p) }
\lp[  \innerAlg(T) \neq \innerAlg(T') \rp]
\rp] \geq \Omega( \rho \log^{-2} n ).
\end{align*}
Moreover, $ \| \p \|_1 \in ( 1 - \eps, 1 +\eps ) \subseteq (0.5, 2)  $ almost surely. 
This concludes the proof of \Cref{prop:unif-test-poisson-lower-bound}.
\end{proof}
Our lower bound for replicable uniformity testing 
easily follows from \Cref{prop:unif-test-poisson-lower-bound} 
and \Cref{lem:poisson-good-string}.
\begin{proof}[Proof of \Cref{thm:unif-test-lower-bound}]
    Assume without loss of generality that $m \geq \log(10/\delta) + \log(10 \log^{2} n/\rho)$.
    From \Cref{prop:unif-test-poisson-lower-bound}, any deterministic Poissonized tester with sample complexity $m = \tilde{o}(\sqrt{n} \eps^{-2} \rho^{-1})$ that is $0.1$-correct with respect to $\calM_{0}$ and $\calM_{\eps}$ cannot be $\rho \log^{-2} n$-replicable with respect to $\calH_{U}$.
    Furthermore, any $\distribution \sim \calH_{U}$ satisfies $\norm{\distribution}{1} \in (0.5, 2)$ with high probability.
    Thus, even conditioned on $\distribution \sim \calH_{U}$ satisfying the norm condition, the deterministic Poissonized tester cannot be both $0.1$-correct \jnew{with respect to $\calM_{0}$ and $\calM_{\eps}$} and $\ll \rho \log^{-2} n$-replicable \jnew{with respect to $\calH_{U}$}.
    The conclusion therefore follows from \Cref{lem:poisson-good-string} (i.e. any randomized tester that is $0.01$-correct \jnew{with respect to $\calM_{0}$ and $\calM_{\eps}$} cannot be $\rho/\polylog(n)$-replicable \jnew{with respect to $\calH_{U}$}.)
\end{proof}

\subsection{Lower Bound for Replicable Closeness Testing}
\label{sec:clossness_lb}
In this section, we give a sample complexity lower bound of $\tilde \Omega(n^{2/3} \eps^{-4/3} \rho^{-2/3} + \sqrt{n} \eps^{-2} \rho^{-1} + \eps^{-2} \rho^{-2} )$ for replicable closeness testing.

Note that closeness testing is at least as hard as uniformity testing (even when replicability is of concern).
Hence, it remains for us to show a lower bound of $\tilde \Omega \lp( n^{2/3} \eps^{-4/3} \rho^{-2/3} \rp)$.
Note that this term dominates exactly in the sublinear regime, 
so it suffices to prove a lower bound in the regime $m \ll n$.

We start by describing the hard instance for replicable closeness testing in this regime. In particular, we construct meta-distributions over pairs of non-negative measures that will be used as inputs to the closeness testing problem.
\begin{definition}[Closeness Test Hard Instance]
\label{def:closeness-hard-instance}
For $\xi\in[0,\epsilon]$, we define $\mathcal N_{\xi}$ to be the distribution over pairs of non-negative measures 
$\p_{\xi},\q_{\xi}$ generated as follows: 
$\forall i \in [n]$
\begin{equation}
    (\p_{\xi}(i),\q_{\xi}(i)) = 
    \begin{cases}
\left(\frac{1-\epsilon}{m},\frac{1-\epsilon}{m}\right) &w.p. \quad\frac{m}{n}\\
\left(\frac{2\epsilon +\xi}{2(n-m)},\frac{2\epsilon - \xi}{2(n-m)}\right) &w.p.\quad \frac{n-m}{2n}\\
\left(\frac{2\epsilon - \xi}{2(n-m)},\frac{2\epsilon + \xi}{2(n-m)}\right) &w.p.\quad \frac{n-m}{2n}.
\end{cases}
\end{equation}
The meta-distribution $\mathcal H_{ C}$ is the distribution over random pairs of non-negative measures $(\p, \q)$ generated as follows: choose $\xi$ uniformly at random from $[0,\epsilon]$, and return $(\p, \q) \sim \mathcal N_{\xi}$.
\end{definition}
Again, thanks to \Cref{lem:poisson-good-string}, after fixing the hard instance to be $\mathcal H_{ C}$,
it then suffices for us to show sample complexity lower bounds 
against any deterministic closeness tester $\innerAlg$ 
within the Poisson sampling model.

\begin{proposition}
\label{prop:closeness-test-poisson-lower-bound}
Let $\mathcal H_C$ be the meta-distribution defined as in \Cref{def:closeness-hard-instance},  $\innerAlg$ be a deterministic tester that takes as input a sample count vector $T \in \N^{2n}$,\footnote{Note that a closeness tester $\innerAlg$ should in principle receive two sets of samples (or two sample count vectors in the Poisson sampling model)---one from $\p$ and the other from $\q$.
However, it is not hard to see that we do not lose any information if we simply concatenate the two sample count vectors together. 
For notational convenience, we denote by $(\p \oplus \q)$ the non-negative measures over $[2n]$, where the first $n$ entries agree with $\p$ and the last $n$ entries agree with $\q$. Then the distribution over the concatenated sample count vectors is simply
given by $ \Poi(m, \p \oplus \q) $.
} and $m = \tilde o( n^{2/3} \eps^{-4/3} \rho^{-2/3} )$. Then it holds 
\begin{align*}
\Pr_{ (\p, \q)  \sim \mathcal H_C}
\lp[ 
\Pr_{T,T' \sim \PoiS\lp( m, \p \oplus \q \rp)}
\lp[
\innerAlg(T) \neq \innerAlg(T')
\rp] 
\geq \log^{-2} n
\text{ and }
\| \lp(\p \oplus \q\rp)/2 \|_1 \in (0.5, 2)
\rp] \geq \rho.
\end{align*}
\end{proposition}

In the rest of the section, we focus on showing \Cref{prop:closeness-test-poisson-lower-bound}.

Define 
$ 
\text{Acc}_m( \p , \q, \innerAlg)
:= \Pr_{ T \sim \PoiS(m, \p \oplus \q) } \lp[ \innerAlg(T) =\text{Accept}\rp].
$
Similar to the corresponding argument in the 
replicable uniformity testing lower bound, 
we begin by showing the intermediate result 
that the average acceptance probability 
$\Ep_{ (\p, \q) \sim \mathcal N_{\xi} }
\lp[ \text{Acc}_m( \p , \q, \innerAlg) \rp]$ is close to $1/2$ 
with probability at least $\rho$ if $\xi$ is chosen randomly 
from $[0, \eps]$. Formally,
\begin{lemma}
\label{lem:avg-acceptance-prob-closeness}
Let $\innerAlg$ be a deterministic Poissonized tester that is 
$0.1$-correct w.r.t. $\mathcal{N}_{0}$ and $\mathcal{N}_{\eps}$. 
Then
$
\Pr_{ \xi \sim \mathcal U([0, \eps]) }
\lp[
\Ep_{ (\p, \q) \sim \mathcal N_{\xi} }
\lp[ \text{Acc}_m( \p , \q, \innerAlg) \rp]
\in (1/3, 2/3)
\rp] \geq \rho$ as long as $m =  o( n^{2/3} \eps^{-4/3} \rho^{-2/3} )$.
\end{lemma}
The argument again uses information theory and is similar to the proof of \Cref{lem:avg-acceptance-prob-unif}. 
Hence, we defer it to \Cref{app:pf-lem:avg-acceptance-prob-closeness}.
Conditioned on some $\xi$ satisfying the probabilistic condition in \Cref{lem:avg-acceptance-prob-closeness},
we then proceed to show that the acceptance probability $\text{Acc}_m( \p , \q, \innerAlg)$ concentrates around the expected acceptance probability $\Ep_{ (\p, \q) \sim \mathcal N_{\xi} }
\lp[ \text{Acc}_m( \p , \q, \innerAlg) \rp]$.

\begin{lemma}[Concentration of Acceptance Probabilities]
    \label{lemma:closeness-acceptance-concentration}
    Let $\xi \in (0, \eps)$  
    and $\innerAlg$ 
    be a deterministic tester satisfying that is $\log^{-2} n$-replicable with respect to $\mathcal{H}_C$.
    Then it holds
    \begin{equation*}
        \Pr_{(\p,\q) \sim \mathcal N_{\xi}} \left( \left| 
        \text{Acc}_m( \p , \q, \innerAlg)
        -
        \Ep_{ (\p',\q') \sim \mathcal N_{\xi} }
        \lp[ \text{Acc}_m( \p' , \q', \innerAlg) \rp]
         \right| > \frac{1}{4} \right) \leq \frac{1}{2} \text{.}
    \end{equation*}
\end{lemma}
This is achieved by analyzing the sample random walk 
$\RW_{m, \mathcal N_\xi}$ analogous to the one considered in the uniformity testing case. 

We begin by defining a random walk on samples drawn from distributions in $\calN_{\xi}$.

\begin{definition}
    \label{def:closeness-sample-rw}
    The Sample Random Walk $\RW_{m, \xi}$ is defined on the graph with vertex set $\N^{2n}$ (where each vertex corresponds to sample count vector $T$ drawn from $\PoiS(m,\p\oplus\q)$,) and transitions $(T_1, T_2)$ are defined by the conditional distribution of $T_2$ given $T_1$ induced by the joint distribution given by the following process:
    \begin{enumerate}
        \item $\pq{} \sim \calN_{\xi}$
        \item $T_1, T_2$ are independently sampled from $\PoiS(m, \p \oplus \q)$.
    \end{enumerate}
    For any sample count vector $T$, let $\RW_{m,\xi}(T)$ be the random variable representing the next step of the random walk from $T$.
\end{definition}

Given a sample count vector $T$, we denote by 
$T_{\p}[i]$ (resp. $T_{\q}[i]$) the frequency of bucket $i$ 
from $\p$ (resp. $\q$).
Let us analyze the random walk $\RW_{m, \xi}$.
As before, we have $T_{\p}[i] \sim \PoiD{m \p_i}$ and $T_{\q}[i] \sim \PoiD{m \q_i}$ independently for all $i$.
Thus, $\RW_{m, \xi}$ is the product of $n$ independent random walks $\RW_{m, \xi, i}$ on vertex set $[m] \times [m]$.
We describe $\RW_{m, \xi}$ by describing each random walk $\RW_{m,\xi, i}$.

If $S, T$ are drawn from the joint distribution defining $\RW_{m,\xi, i}$, 
\begin{align*}
    &\Pr(S_{\p}[i] = a, T_{\p}[i] = b, S_{\q}[i] = c, T_{\q}[i] = d) \\
    &~~~~= \frac{m}{n} \frac{e^{4(1 - \eps)} \left( 1 - \eps \right)^{a + b + c + d}}{a!b!c!d!} \\
    &~~~~~~~~+ \frac{n - m}{2n} \Bigg( \frac{e^{(2 \eps + \xi)m/(n-m)} \left( \frac{(2 \eps + \xi)m}{2(n - m)} \right)^{a+b}}{a!b!} \frac{e^{(2 \eps - \xi)m/(n-m)} \left( \frac{(2 \eps - \xi)m}{(n - m)} \right)^{c + d}}{c!d!} \\
    &~~~~~~~~+ \frac{e^{(2 \eps - \xi)m/(n-m)} \left( \frac{(2 \eps - \xi)m}{(n - m)} \right)^{a+b}}{a!b!}\frac{e^{(2 \eps + \xi)m/(n-m)} \left( \frac{(2 \eps + \xi)m}{2(n - m)} \right)^{c+d}}{c!d!}  \Bigg) \\
    &~~~~= \frac{m e^{4(1 - \eps)} \left( 1 - \eps \right)^{a+b+c+d}}{n \cdot a!b!c!d!} + \frac{n - m}{2n} \left( \frac{m}{2(n - m)} \right)^{a + b + c + d} \frac{e^{4 \eps m/(n - m)}}{a!b!c!d!} f_{a, b, c, d}(\xi) \;,
\end{align*}
where
\begin{equation*}
    f_{a + b, c + d}(\xi) = \left( 2 \eps + \xi \right)^{a+b} \left( 2 \eps - \xi \right)^{c+d} + \left( 2 \eps - \xi \right)^{a+b} \left( 2 \eps + \xi \right)^{c+d} \text{.}
\end{equation*}

Similarly, we compute the marginal distribution as
\begin{align*}
    \Pr(S_{\p}[i] = a, S_{\q}[i] = c) &= \frac{m e^{2(1 - \eps)} \left( 1 - \eps \right)^{a+c}}{n \cdot a!c!} + \frac{n - m}{2n} \left( \frac{m}{2(n - m)} \right)^{a + c} \frac{e^{2\eps m/(n - m)}}{a!c!} f_{a, c}(\xi) \text{.}
\end{align*}

To describe the random walk transition probabilities, 
we compute the conditional distribution
\begin{align*}
    P((a, c), (b, d)) &= \Pr(T_{\p}[i] = b, T_{\q}[i] = d \mid S_{\p}[i] = a, S_{\q}[i] = c) \\
    &= \frac{\Pr(T_{\p}[i] = b, T_{\q}[i] = d, S_{\p}[i] = a, S_{\q}[i] = c)}{\Pr(S_{\p}[i] = a, S_{\q}[i] = c)}
\end{align*}
and note that, as before, the stationary distribution 
is given by the probability vector 
$\pi(a, c) = \Pr(S_{\p}[i] = a, S_{\q}[i] = c)$.

Following identical arguments as \Cref{lem:random-walk-indistinguishability}, we show that over few steps of the random walk, the outcome of the algorithm does not change significantly.

\begin{lemma}
    \label{lemma:closeness-random-walk-agreement}
    Let $\innerAlg$ be $1/(10 K)$-replicable with respect to $\mathcal{H}_C$.
    Let $\pq{} \sim \calN_{\xi}$ and $T_0 \sim \PoiS(m, \p \oplus \q)$.
    For $1 \leq k \leq K$, let $T_k \sim \RW_{m, \xi}(T_{k - 1})$.
    Then, 
    \begin{equation*}
        \Pr_{T_{0}, \dotsc, T_{k}}\left( \bigcup_{k = 1}^{K} \set{ \innerAlg(T_{k - 1}) \neq \innerAlg(T_{k}) } \right) < \frac{1}{10} \text{.}
    \end{equation*}

\end{lemma}

We now bound the mixing time of the random walk $\RW_{m, \xi}$.
As in the argument for uniformity, we begin by bounding the (constant) mixing time of a single coordinate.

\begin{lemma}
    \label{lemma:closeness-rw-coord-mix}
    Suppose $m \leq n/10$.
    The random walk $\RW_{m, \xi, i}$ has mixing time $\tau(0.11) = O(1)$.
\end{lemma}

\begin{proof}
    Let $Y_{0} = (\ell_{1}, \ell_{2})$ denote the current step.
    Let $X \sim \set{A, B, C}$ be drawn randomly with probabilities $\frac{m}{n}$, $\frac{n - m}{2n}$, $\frac{n - m}{2n}$ respectively.
    Let $Z_{1}^{A} \sim \PoiD{1-\eps}, Z_{1}^{B} \sim \PoiD{\frac{(2\eps + \xi)m}{2(n - m)}}, Z_{1}^{C} \sim \PoiD{\frac{(2\eps - \xi)m}{2(n - m)}}$ and $Z_{2}^{A} \sim \PoiD{1-\eps}, Z_{2}^{B} \sim \PoiD{\frac{(2\eps - \xi)m}{2(n - m)}}, Z_{2}^{C} \sim \PoiD{\frac{(2\eps + \xi)m}{2(n - m)}}$.
    The next step of the random walk $Y_{1} \sim \RW_{m,\xi, i}(Y_{0})$ is taken according to the distribution
    \begin{align*}
        P((\ell_1, \ell_2), (k_1, k_2)) &= \Pr(X = A \mid Y_{0} = (\ell_{1}, \ell_{2})) \Pr(Z_{1}^{A} = k_1, Z_{2}^{A} = k_{2}) \\
        &~~~~~~+ \Pr(X = B \mid Y_{0} = (\ell_{1}, \ell_{2})) \Pr(Z_{1}^{B} = k_1, Z_{2}^{B} = k_{2}) \\
        &~~~~~~+ \Pr(X = C \mid Y_{0} = (\ell_{1}, \ell_{2})) \Pr(Z_{1}^{C} = k_1, Z_{2}^{C} = k_{2}) \text{.}
    \end{align*}
    We show that regardless of the current state $(\ell_1, \ell_2)$, the random walk reaches state $(0, 0)$ with reasonable probability.
    Since $\eps < 0.1$, we have 
    \begin{align*}
        \Pr(Z_{1}^{A} = 0, Z_{2}^{A} = 0) &= \left( e^{-(1 - \eps)} \frac{(1 - \eps)^{0}}{0!} \right)^{2} = e^{-2(1 - \eps)} \geq 0.13 \text{,} \\
        \Pr(Z_{1}^{B} = 0, Z_{2}^{B} = 0) &= e^{-\frac{(2\eps+\xi)m}{2(n-m)}} e^{-\frac{(2\eps-\xi)m}{2(n-m)}} = e^{-\frac{2 \eps m}{n - m}} \geq e^{-2\eps} > 0.8 \text{,} \\
        \Pr(Z_{1}^{C} = 0, Z_{2}^{C} = 0) &> 0.8 \;,
    \end{align*}
    where we have used $m < n/10$ in the second bound. 
    In particular, note that $\frac{m}{n - m} \leq \frac{1}{9}$ (here we only use that $1/9 < 1$).
    The last follows identically.
    Then, regardless of $(\ell_1, \ell_2)$, we have $P((\ell_1, \ell_2), (0, 0)) \geq 0.13$.
    In particular, after $O(1)$ steps, we can guarantee that with probability $0.99$ we reach the state $(0, 0)$.
    We can then assume without loss of generality that $\ell_1 = \ell_2 = 0$.

    We now examine the distribution of $X$ conditioned on $Y_{0} = (0, 0)$.
    First, note that
    \begin{align*}
        \Pr(Y_{0} = (0, 0)) &= \frac{m}{n} e^{-2(1-\eps)} + \frac{n - m}{2n} \left( e^{-2\eps m/(n - m)} + e^{-2\eps m/(n - m)} \right) \\
        &= \frac{m}{n} e^{-2(1-\eps)} + \frac{n - m}{n} e^{-2\eps m/(n - m)} \text{.}
    \end{align*}
    Then we argue that the distribution of $X$ conditioned on $Y_{0} = (0, 0)$ is reasonably random.
    \begin{align*}
        \Pr(X = B \mid Y_{0} = (0, 0)) &= \frac{\Pr(X = B, Y_{0} = (0, 0))}{\Pr(Y_{0} = (0, 0)} \\
        &= \frac{\frac{n - m}{2n} e^{-2\eps/(n - m)}}{\frac{m}{n} e^{-2(1-\eps)} + \frac{n - m}{n} e^{-2\eps/(n - m)}} \\
        &= \frac{1/2}{\frac{m}{n - m} \exp \left( \frac{2 \eps m}{n - m} - 2 (1 - \eps) \right) + 1} \text{.}
    \end{align*}
    Observe that $\frac{m}{10n}
    \leq \frac{m}{n} \exp\left( - 2\right) 
    \leq \frac{m}{n - m} \exp \left( \frac{2 \eps m}{n - m} - 2 (1 - \eps) \right) \leq 
    \frac{2m}{n} \exp \left( \frac{\eps}{9} - 1.8 \right) \leq 
    \frac{m}{n}$. 
    Applying the valid inequalities 
    $0.5 - x \leq \frac{0.5}{1 + x} \leq 0.5 - x/3$, 
    for small $x > 0$, we obtain 
    \begin{equation*}
        0.5 - \frac{m}{n} \leq \Pr(X = B \mid Y_{0} = (0, 0)) \leq 0.5 - \frac{m}{30n} \text{.}
    \end{equation*}
    As a result, we can conclude
    \begin{equation*}
        \frac{m}{15 n} \leq \Pr(X = A \mid Y_{0} = (0, 0)) \leq \frac{2m}{n} \text{.}
    \end{equation*}
    In the stationary distribution, we have
    \begin{equation*}
        \Pr(X = A) = \frac{m}{n}~~,~~\Pr(X = B) = \Pr(X = C) = \frac{n - m}{2n} \text{.}
    \end{equation*}
    In particular, the total variation distance between $X$ in the stationary distribution and $X$ conditioned on $Y_{0} = (0, 0)$ is at most $\frac{m}{n} \leq \frac{1}{10}$ using our assumption $m \leq n/10$.
    Thus, from the initial state $(0, 0)$, the random walk mixes to within $0.1$ total variation distance to the stationary distribution.
    We union bound with the $0.01$ probability of not reaching the stationary distribution in $O(1)$ steps to conclude the argument.
\end{proof}

Thus, we can bound the relaxation time of $\RW_{m,\xi, i}$ as
\begin{equation*}
    t_{\rel} \leq \frac{\tau(0.11)}{\log(1/0.22)} + 1 = O(1)\text{.}
\end{equation*}

We now bound the mixing time from the initial distribution.
\begin{lemma}
    \label{lemma:closeness-coord-rw-mixing-delta}
    Let $m \leq n/10$.
    Let $\gamma(x) \sim \PoiD{1 - \eps} \otimes \PoiD{1 - \eps}$, $\gamma(x) \sim \PoiD{\frac{(2 \eps + \xi)m}{2(n - m)}} \otimes \PoiD{\frac{(2 \eps - \xi)m}{2(n - m)}}$, or $\gamma(x) \sim \PoiD{\frac{(2 \eps - \xi)m}{2(n - m)}} \otimes \PoiD{\frac{(2 \eps + \xi)m}{2(n - m)}}$ denote the initial distribution.
    The random walk $\RW_{m,\xi, i}$ has mixing time $\tau(\delta) = O(\log(n/\delta))$.
\end{lemma}

\begin{proof}
    As in \Cref{lemma:coord-rw-mixing-delta}, we begin with following the inequality for any pair of states $x = (x_1, x_2)$ and $y = (y_1, y_2)$.
    \begin{equation*}
        \left| \gamma(x) P^{t}(x, y) - \gamma(x) \pi(y) \right| \leq \frac{\lambda_{*}^{t} \gamma(x) \sqrt{\pi(y)}}{\sqrt{\pi(x)}} \text{.}
    \end{equation*}

    We bound the ratio $\gamma(x)/\pi(x)$.
    \begin{claim}
        \label{clm:closeness-gamma-pi-ratio}
        For all states $x = (x_1, x_2)$,
        \begin{equation*}
            \frac{\gamma(x)}{\pi(x)} \leq \frac{n}{m} \text{.}
        \end{equation*}
    \end{claim}

    \begin{proof}
        We split into the cases $\gamma \sim Z^{A}, \gamma \sim Z^{B}, \gamma \sim Z^{C}$ as defined in \Cref{lemma:closeness-rw-coord-mix}.
        First, we write
        \begin{align*}
            \pi(x) &= \frac{m}{n} \Pr(Z^{A} = x) + \frac{n - m}{2n} \Pr(Z^{B} = x) + \frac{n - m}{2n} \Pr(Z^{C} = x) \;. 
        \end{align*}
        Then
        \begin{align*}
            \frac{\Pr(Z^{A} = x)}{\pi(x)} &\leq \frac{\Pr(Z^{A} = x)}{\frac{m}{n} \Pr(Z^{A} = x)} \leq \frac{n}{m} \\
            \frac{\Pr(Z^{B} = x)}{\pi(x)} &\leq \frac{\Pr(Z^{B} = x)}{\frac{n - m}{2 n} \Pr(Z^{B} = x)} \leq \frac{2 n}{n - m} \leq 4 \\
            \frac{\Pr(Z^{C} = x)}{\pi(x)} &\leq 4 \;,
        \end{align*}
        where in the second and third cases we used $n - m \geq n/2$.
        Finally, we conclude by observing $\frac{n}{m} \geq 4$.
    \end{proof}

    Then, we sum over $x$ to obtain 
    \begin{equation*}
        \left| \left( \sum_{x} \gamma(x) P^{t}(x, y) \right) - \pi(y) \right| \leq \lambda_{*}^{t} \sqrt{\frac{n}{m} \pi(y)} \sum_{x} \sqrt{\gamma(x)} \;.
    \end{equation*}
    We now bound $\sum_{x} \sqrt{\gamma(x)}$.
    In all three cases $Z^{A}, Z^{B}, Z^{C}$, we have that the Poisson distribution (in both distributions) has parameter $\lambda \leq 1$. 
    By standard Poisson concentration, for any $i \in \set{1, 2}$ and $D \in \set{A, B, C}$ we have
    \begin{equation*}
        \Pr(Z_{i}^{D} > x) = \Pr(Z_{i}^{D} > \lambda + (x - 1)) = e^{-\Omega(x - 1)} \text{.}
    \end{equation*}
    Then, we bound for any $X \in \set{A, B, C}$
    \begin{align*}
        \sum_{x} \sqrt{\gamma(x)} &= \sum_{x_1 = 0}^{\infty} \sum_{x_2 = 0}^{\infty} \sqrt{ \Pr \left( Z_{1}^{X} = x_{1} \right) \Pr \left( Z_{2}^{X} = x_{2} \right) } \\
        &= \sum_{x_1 = 0}^{\infty} \sqrt{\Pr \left( Z_{1}^{X} = x_{1} \right)} \sum_{x_2 = 0}^{\infty} \sqrt{\Pr \left( Z_{2}^{X} = x_{2} \right)} \\
        &= O(1) \;,
    \end{align*}
    where the first equality follows by definition of $\gamma$ and independence of $\p, \q$, and the second and third equalities follow as $\sum_{x} \sqrt{\Pr(Z_{i}^{X} = x)}$ converges absolutely, which we showed in \Cref{lemma:coord-rw-mixing-delta}.
    Thus, we arrive at the inequality
    \begin{equation*}
        \left| \left( \sum_{x} \gamma(x) P^{t}(x, y) \right) - \pi(y) \right| = \bigO{ \lambda_{*}^{t} \sqrt{\frac{n}{m} \pi(y)} }  \text{.}
    \end{equation*}
    Summing over $y$ and applying a similar argument (see \Cref{lemma:coord-rw-mixing-delta} for details), we obtain
    \begin{equation*}
        \sum_{y} \left| \left( \sum_{x} \gamma(x) P^{t}(x, y) \right) - \pi(y) \right| = \bigO{ \lambda_{*}^{t} \sqrt{\frac{n}{m}} }  \text{.}
    \end{equation*}
    Thus, since $\lambda_{*} < 1$ is an absolute constant less than $1$, we conclude that from any of the three initial distributions, the random walk $\RW_{m,\xi, i}$ mixes to within $\delta$ of the stationary distribution in time $\tau(\delta) = O(\log((n/m)/\delta)) = O(\log(n/\delta))$.

\end{proof}

Now, using identical arguments as in \Cref{lemma:S-concentration-for-good-xi}, we can conclude with the following lemma.

\begin{lemma}
    \label{lemma:A-B-C-concentration-for-good-xi}
    Let $K = \tau(0.01)$ and $\xi \in [0, \eps]$.
    Suppose $\innerAlg$ is $\frac{1}{10 K}$-replicable with respect to $\mathcal{H}_C$.
    Then,
    \begin{equation*}
        \Pr_{\pq{} \sim \calN_{\xi}} \left( \left| \Ep_{T \sim \pq{}} \left[ \innerAlg(T) \right] - \Ep_{\pq{2} \sim \calN_{\xi}, T' \sim \pq{2}} \left[ \innerAlg(T') \right] \right| > \frac{1}{4} \right) \leq \frac{1}{2} \text{.}
    \end{equation*}
\end{lemma}

We now prove \Cref{lemma:closeness-acceptance-concentration}.

\begin{proof}
    From \Cref{lemma:closeness-coord-rw-mixing-delta}, we have that $\tau(0.01/n) = O(\log n))$.
    Since $\innerAlg$ is $\log^{-2} n$-replicable w.r.t. $\mathcal{H}_C$, it is also $1/(10 K)$-replicable w.r.t. $\mathcal{H}_C$.
    The conclusion follows.
\end{proof}

Combining \Cref{lem:avg-acceptance-prob-closeness} and \Cref{lemma:closeness-acceptance-concentration} then yields the proof of \Cref{prop:closeness-test-poisson-lower-bound}.

\begin{proof}[Proof of \Cref{prop:closeness-test-poisson-lower-bound}]
    The proof follows using analogous arguments as \Cref{prop:unif-test-poisson-lower-bound}, applying \Cref{lem:avg-acceptance-prob-closeness} and \Cref{lemma:closeness-acceptance-concentration} where appropriate.
\end{proof}

We are ready to prove \Cref{thm:closensess-lb}. 
The theorem follows from \Cref{prop:closeness-test-poisson-lower-bound} and \Cref{lem:poisson-good-string}.

\begin{proof}[Proof of \Cref{thm:closensess-lb}]
    Note that a lower bound of $\tilde{\Omega}(\sqrt{n} \eps^{-2} \rho^{-1} + \eps^{-2} \rho^{-2})$ follows immediately from lower bounds for uniformity testing and bias estimation respectively.
    It suffices to show a lower bound of $\tilde{\Omega}(n^{2/3} \eps^{-4/3} \rho^{-2/3})$.

    \Cref{prop:closeness-test-poisson-lower-bound} says that any deterministic tester that is $0.01$-correct takes Poissonized samples with sample complexity $m = \tilde{o}(n^{2/3} \eps^{-4/3} \rho^{-2/3})$ is not $\rho/(\log n)^{2}$-replicable with respect to the hard instance $\calH_{C}$.
    Then, from \Cref{lem:poisson-good-string} we may conclude that any randomized tester with fixed sample complexity $m = \tilde{o}(n^{2/3} \eps^{-4/3} \rho^{-2/3})$ is not $\rho/\polylog(n)$-replicable with respect to $\calH_{C}$, concluding the proof.
\end{proof}

\bibliographystyle{alpha}
\bibliography{references}

\newpage

\appendix

\section*{Appendix}

\section{Proof of \Cref{MI_asymp}}
\label{app:MI_asymp}
        Denote $\alpha:=\Pr[M=a,X=1],\beta:= \Pr[M=a,X=0]$ for simplicity. Since $\Pr[X=1]=\Pr[X=0]=1/2\implies \beta=\Theta(\alpha)$, then $\frac{(\beta-\alpha)^2}{\beta+\alpha} =\Theta\left(\frac{(\beta-\alpha)^2}{\alpha}\right)=  \Theta\left(\frac{(\beta-\alpha)^2}{\beta}\right)$. 
        By definition, 
        \begin{align*}
            I(X:M) &= \sum_a \sum_{i=0,1}\Pr[X=i,M=a] \log \left(\frac{\Pr[X=i,M=a]}{\Pr[X=i] \Pr[M=a]}\right)\\
            &= \frac{1}{2}\sum_a \left(\beta \log \left(\frac{2\beta}{\beta+\alpha}\right)+ \alpha \log \left(\frac{2\alpha}{\beta+\alpha}\right) \right)\\
            &\text{(rearranging)}\\
            & = \Theta(1)\sum_a \left(\beta\log \left(\frac{1}{1+\frac{\alpha-\beta}{2\beta}}\right)+ \alpha \log \left(\frac{1}{1-\frac{\alpha-\beta}{2\alpha}}\right)\right).
        \end{align*}
Denote $A:=\frac{\alpha-\beta}{2\beta}$ and $B:=\frac{\alpha-\beta}{2\alpha}$. Then by Taylor expansion of $\log\left(\frac{1}{1+A}\right) $ and $\log\left(\frac{1}{1-B}\right) $, we have that 
        \begin{align*}
            I(X:M) &= \Theta(1) \sum_a \left(\beta\left(\sum_{n=1}^{\infty} (-1)^n\frac{A^n}{n}\right)+\alpha \left(\sum_{n=1}^{\infty} \frac{B^n}{n}\right)\right)\nonumber\\
            &= \Theta(1)\sum_a \left(\sum_{\substack{n=3\\ n\text{ odd}}} \frac{1}{n}(\alpha B^n-\beta A^n) + \sum_{\substack{n=2\\ n\text{ even}}} \frac{1}{n}(\beta A^n+\alpha B^n)\right)\label{IXMub}\\
            &\le O(1) \sum_a  2\sum_{n=2}(\alpha B^n+\beta A^n)\le O(1) \alpha B^2 = \sum_aO\left(\frac{(\beta-\alpha)^2}{\alpha}\right)\nonumber \;,
        \end{align*}
        as desired.

\section{Proof of \texorpdfstring{\Cref{lem:avg-acceptance-prob-unif}}{}}
\label{app:pf-lem:avg-acceptance-prob-unif}
At a high level, we appeal to the same argument as in \cite{liu2024replicable} 
{to analyze
the expected acceptance probability of the tester.}
{The framework proceeds as follows. Fix any $\epsilon_0 < \epsilon_1$ in $[0, \epsilon]$ such that $\epsilon_1 - \epsilon_0 < \epsilon \rho$. Let $X$ be an unbiased random bit,  $\tilde{\distribution} \sim \mathcal{M}_{\epsilon_X}$ be defined as in Definition~\ref{def:uniform-hard-instance}, and $T \sim \PoiS(m, \tilde{\distribution})$ be defined as in Definition~\ref{def:poi_samp}. Then, the mutual information between $X$ and $T$ is bounded from above by a function of the parameters $m$, $n$, $\epsilon$, and $\rho$, as stated formally in Lemma~\ref{MI_unif}.} 
Secondly, given the mutual information bound,
we know that with limited amount of samples, for any pair of $\epsilon_0,\epsilon_1\in [0,\epsilon]$ that are $\rho\epsilon$ close to each other, $\Ep_{ \distribution \sim \mathcal M_{\epsilon_0} }[ \text{Acc}_m(\distribution,\mathcal{A})]$ and $\Ep_{ \distribution \sim \mathcal M_{\epsilon_1} }[ \text{Acc}_m(\distribution,\mathcal{A})]$ 
\snew{must be}
close to each other. 
\snew{See \Cref{lem:lipchi_apf} for the formal statement.}
Lastly, given $\mathcal{A}$ as above and is $0.1$-correct w.r.t. $\mathcal{M}_{0}$ and $\mathcal{M}_{\eps}$, then the acceptance probability function should satisfy that $\Ep_{ \distribution \sim \mathcal M_{0} }[ \text{Acc}_m(\distribution,\mathcal{A})] \ge 0.9$ and $\Ep_{ \distribution \sim \mathcal M_{\epsilon} }[ \text{Acc}_m(\distribution,\mathcal{A})] <0.1$. Thus, by the mean value theorem there exists $\xi^*\in(0,\epsilon)$ such that $\Ep_{ \distribution \sim \mathcal M_{\xi^*} }[ \text{Acc}_m(\distribution,\mathcal{A})] = \frac{1}{2}$. Furthermore, from the above Lipschitzness of $\Ep_{ \distribution \sim \mathcal M_{\xi} }[ \text{Acc}_m(\distribution,\mathcal{A})] $ in $\xi$ we know that for at least $\rho$ fraction of $\xi\in[0,\epsilon],$ $\Ep_{ \distribution \sim \mathcal M_{\xi} }[ \text{Acc}_m(\distribution,\mathcal{A})]\in (1/3,2/3)$, which concludes the proof of Lemma \ref{lem:avg-acceptance-prob-unif}.

We \snew{begin by} by showing the mutual information bound.

\begin{lemma}[Mutual Information Bound for Uniformity Testing Hard Instance]
    Let $m = o(n\eps^{-2}\log^{-2} n)$, $\epsilon_0<\epsilon_1\in[0,\epsilon] $ be such that $\epsilon_1-\epsilon_0 < \epsilon \rho$, $X$ be an unbiased random bit, $\mathcal{M}_{\epsilon_X}$ be the distribution over measures defined as in \Cref{def:uniform-hard-instance}, $\tilde{\distribution}\sim\mathcal{M}_{\epsilon_X}$, $T\sim \PoiS(m,\tilde{\distribution})$.
     Then the mutual information $I(X:T_1,\cdots,T_n)$ 
    satisfies:
    \[
    I(X:T_1,\cdots,T_n)=O\left(\frac{m^2}{n}\epsilon^4\rho^2\log^4n\right)+o(1).
    \]
    \label{MI_unif}
\end{lemma}

\begin{proof}[Proof of Lemma \ref{MI_unif}]
Let $\delta := \epsilon_1-\epsilon_0 = O(\epsilon \rho)$. 
Since $T_i's$ are conditionally independent conditioned on $X$, we have that 
\[
I(X:T_1,\cdots,T_n) \le \sum_{i=1}^nI(X:T_i)=nI(X:T_1).
\] 
Therefore, it suffices show that $I(X:T_1)=O\left(\frac{m^2}{n^2}\epsilon^4\rho^2\log^4n\right)+o\left(\frac{1}{n}\right).$ 

We start by expanding the conditional probabilities of $T_1$ conditioned on value of $X$.
        \begin{align*}
            T_1|(X=0) \sim \frac{1}{2}\PoiD{\frac{m}{n}(1+\epsilon_0)} + \frac{1}{2}\PoiD{\frac{m}{n}(1-\epsilon_0)},
        \end{align*}
        and similarly,
        \begin{align*}
            T_1|(X=1) \sim  \frac{1}{2}\PoiD{\frac{m}{n}(1+\epsilon_1)} + \frac{1}{2}\PoiD{\frac{m}{n}(1-\epsilon_1)},
        \end{align*}
        then we can expand $\Pr[T_1=a|X=0]$ and $\Pr[T_1=a|X=1]$ accordingly. Indeed, 
    \begin{align*}
        \Pr[T_1=a|X=0]& =\frac{1}{2a!}\left(\frac{m}{n}\right)^a(1+\epsilon_0)^a \exp\left(-\frac{m}{n}(1+\epsilon_0)\right) + \frac{1}{2}\frac{1}{a!}\left(\frac{m}{n}\right)^a(1-\epsilon_0)^a \exp\left(-\frac{m}{n}(1-\epsilon_0)\right)\\
        &= \frac{1}{2a!}\left(\frac{m}{n}\right)^a\exp\left(-\frac{m}{n}\right)\left( \exp\left(-\frac{m\epsilon_0}{n}\right)(1+\epsilon_0)^a + \exp\left(\frac{m\epsilon_0}{n}\right)(1-\epsilon_0)^a\right),\\
        \Pr[T_1=a|X=1]&= \frac{1}{2a!}\left(\frac{m}{n}\right)^a\exp\left(-\frac{m}{n}\right)\left(\exp\left(-\frac{m(\epsilon_0+\delta)}{n}\right)(1+\epsilon_0+\delta)^a + \exp\left(\frac{m(\epsilon_0+\delta)}{n}\right)(1-\epsilon_0-\delta)^a\right).
    \end{align*} 
    Since $\delta = o(\epsilon),$ $ \Pr[T_1=a|X=0]= \Theta\left( \Pr[T_1=a|X=1]\right).$ By \Cref{MI_asymp}, it suffices to show that $\bar{I}:=\sum_{a} \frac{(\Pr[T_1=a|X=0] - \Pr[T_1=a|X=1])^2}{\Pr[T_1=a|X=0] + \Pr[T_1=a|X=1]} =O\left(\frac{m^2}{n^2}\epsilon^4\rho^2\log^4n\right)+o\left(\frac{1}{n}\right).$
    
    Let 
    \begin{equation}
        f_a(y):= \exp\left(-\frac{my}{n}\right)(1+y)^a + \exp\left(\frac{my}{n}\right)(1-y)^a\label{fexpform}.
    \end{equation} 
    Then it holds 
    \begin{align*}
        \bar{I} & = O(1)\sum_{a=0}^{\infty}\frac{1}{a!}\left(\frac{m}{n}\right)^a \exp\left(-\frac{m}{n}\right) \frac{(f_a(\epsilon_0)-f_a(\epsilon_0+\delta))^2}{f_a(\epsilon_0)+f_a(\epsilon_0+\delta)}=:O(1)\sum_{a=0}^{\infty}\bar{I}_a,
    \end{align*}
    where for simplicity, denote $\bar{I}_a := \frac{1}{a!}\left(\frac{m}{n}\right)^a\frac{(f_a(\epsilon_0) - f_a(\epsilon_0 + \delta))^2}{f_a(\epsilon_0) + f_a(\epsilon_0 + \delta)}.$
    Then by the mean value theorem $\bar{I}_a \le \delta^2\frac{\max_{y\in [\epsilon_0,\epsilon_0+\delta]}\left(\frac{\partial}{\partial y}f_a(y)\right)^2}{f_a(\epsilon_0)+f_a(\epsilon_0+\delta)}$, whence to bound $\bar{I}_a$ from above, it suffices to bound the denominator of RHS from below and the numerator of RHS from above separately. We next break into 3 cases \snew{depending on the size of} $\frac{m}{n}.$
    
    \textbf{Case 1:} For the sublinear regime, i.e., $\frac{m}{n}\le 1/2,$ we break into 3 cases depending on the value of $a.$
    
     when $a=0,$ applying the mean value theorem 
        \snew{gives} that $|f_0(\epsilon_0 ) - f_0(\epsilon_0 + \delta) |\le  -\frac{m}{n}\delta\frac{2m(\epsilon_0 + \delta)}{n}\exp\left(\frac{m(\epsilon_0 + \delta)}{n}\right)$. Since $f_0(\eps_0)+f_0(\eps_1)=\Omega(1),$ we have that $ \bar{I}_0=O(1)\left(f_0(\epsilon_0) - f_0(\epsilon_0+\delta)\right)^2 = O\left(\frac{m^4}{n^4}\epsilon_0^2\delta^2\right)= O \left(\frac{m^2}{n^2}\epsilon_0^2\delta^2\right).$ 
        
    when $a=1,$ 
            \begin{align*}
                &|f_1(\epsilon_0)-f_1(\epsilon_0+\delta)|\\
                &\le |f_0(\epsilon_0)-f_0(\epsilon_0 + \delta)|\\
                &+\left|\epsilon_0\left(\exp\left(\frac{m\epsilon_0}{n}\right) -\exp\left(-\frac{m\epsilon_0}{n}\right)\right)- (\epsilon_0+\delta)\left(\exp\left(-\frac{m(\epsilon_0+\delta)}{n}\right)+ \exp\left(\frac{m(\epsilon_0+\delta)}{n}\right)\right)\right|\\
                &\le O\left(\frac{m^2}{n^2}\epsilon_0 \delta\right) + \epsilon_0\left|\exp\left(-\frac{m\epsilon_0}{n}\right)-\exp\left(-\frac{m(\epsilon_0+\delta)}{n}\right)\right|+\epsilon_0\left|\exp\left(\frac{m(\epsilon_0+\delta)}{n}\right)-\exp\left(\frac{m\epsilon_0}{n}\right)\right|\\
                &\qquad + \delta\left|\exp\left(\frac{m(\epsilon_0+\delta)}{n}\right)-\exp\left(-\frac{m(\epsilon_0+\delta)}{n}\right)\right|\\
                &=O\left(\frac{m}{n}\epsilon_0 \delta\right).\tag{the mean value theorem}
            \end{align*}
            Combining with the fact that $f_1(\eps_0)+f_1(\eps_1) =\Omega(1)$, we have that $\bar{I}_1= O(1)\left(f_1(\epsilon_0) - f_1(\epsilon_0+\delta)\right)^2 = O\left(\frac{m^2}{n^2}\epsilon^2\delta^2\right).$
            
            when $a \ge 2,$
            \begin{align*}
            \frac{\partial}{\partial y} f_a(y) = &-\frac{m}{n}\exp\left(-\frac{my}{n}\right)(1+y)^a + a(1+y)^{a-1}\exp\left(-\frac{my}{n}\right)\\ 
            &+\frac{m}{n}\exp\left(\frac{my}{n}\right)(1-y)^a - a(1-y)^{a-1}\exp\left(\frac{my}{n}\right).
            \end{align*}
            Before bounding $\left|\frac{\partial}{\partial y} f_a(y)\right|,$ we introduce a technical claim that is helpful in the rest of the proof to \snew{show} the monotonicity of specific families of functions.
            \begin{claim}
             For $a,b,s,d,x\in \R$, when $s+dx\ge 0,$ if $dk\ge(\le,resp.) b(s+dx)$ then $\exp(a-bx)(s+dx)^k$ is nondecreasing(nonincreasing, resp.) \snew{as a function of $x$}. 
            \label{exp()(1+x)nondec}
            \end{claim}
            \begin{proof}[Proof of Claim \ref{exp()(1+x)nondec}]
                $\frac{\partial}{\partial x}\left(\exp(a-bx)(s+dx)^k\right)= \exp(a-bx)(s+dx)^{k-1}[dk-b(s+dx)],$ then if $dk\ge b(s+dx)$, we have that $\frac{\partial}{\partial x}\left(\exp(a-bx)(s+dx)^k\right)\ge 0.$ Similar argument applies when $dk\le b(s+dx).$
            \end{proof}
            
            When $y\in [\epsilon_0,\epsilon_0+\delta]$, by Claim \ref{exp()(1+x)nondec}, since $a-1\ge 1\ge (1+y)\frac{m}{n}$,
            \begin{align*}
                \left|\frac{\partial}{\partial y} f_a(y)\right| &\le \exp\left(-\frac{my}{n}\right)\left[\frac{m}{n}\left((1+y)^a-(1-y)^a\right) + a \left((1+y)^{a-1}-(1-y)^{a-1}\right)\right]\\
                &\le\frac{m}{n}a(1+y)^{a-1}2y+a(a-1)(1+y)^{a-2}2y=O\left(2^aa^2y\right).
            \end{align*}
            Thus, we have that $ \max_{y\in [\epsilon_0,\epsilon_0+\delta]}\left(\frac{\partial}{\partial y}f_a(y)\right)^2=\max_{y\in [\epsilon_0,\epsilon_0+\delta]}\left(O\left(2^{2a}a^4y^2\right)\right)=O\left(4^aa^4\epsilon^2\right)$. 
            Since $f_a(\epsilon_0)+f_a(\epsilon_0+\delta)=\Omega(1)$,
            \begin{align*}
            \sum_{a=2}^{\infty} \bar{I}_a&\le O(\delta^2\epsilon^2)\sum_{a=2}^{\infty}\frac{4^aa^4}{a!}\left(\frac{m}{n}\right)^a .
            \end{align*}
            Since $\sum_{a=2}^{\infty}\frac{4^aa^4}{a!}$ is a converging series, it can be bounded by $O(1). $ Therefore,
            $ \sum_{a=2}^{\infty}\bar{I}_a\le O(\delta^2\epsilon^2)\sum_{a=2}^{\infty}\left(\frac{m}{n}\right)^a=O\left(\frac{m^2}{n^2}\delta^2\epsilon^2\right).$
        
        In conclusion, from the above three cases, $\bar{I}= O\left(\frac{m^2}{n^2}\delta^2\epsilon^2\right).$

        \textbf{Case 2:} For the superlinear regime when $n/2\le m \le o\left(\frac{n}{\epsilon^2\log^2n}\right),$ we start by noticing that when $a$ deviates far enough from $\frac{m}{n}$, the sum of all such $\bar{I}_a$ is negligible. More specifically,
        let $\lambda = \frac{m(1-\epsilon-\delta)}{n}$, and $c>0$ be a constant such that $\left.\exp\left(-\frac{x^2}{2(\lambda+x)}\right)\right|_{x=c \log n\sqrt{m/n} }\le \frac{1}{n^2}$ then by \Cref{lemma:poisson-concentration},  
        \begin{align*}
            \sum_{\substack{a\ge \lfloor\lambda + c\log n \sqrt{m/n}\rfloor\\ a\le \lceil\lambda - c\log n \sqrt{m/n}\rceil}} \frac{(\Pr[T_1=a|X=0] - \Pr[T_1=a|X=1])^2}{\Pr[T_1=a|X=0] + \Pr[T_1=a|X=1]} &= o\lp(\frac{1}{n}\rp).
        \end{align*}
        Therefore, to compute $\bar{I}$, it suffices to consider $\bar{I}_a$ when $a\in \left[\lambda - c\log n \sqrt{m/n},\lambda + c\log n \sqrt{m/n}\right].$
        Instead of directly bounding $|f_a(\epsilon_0)-f_a(\epsilon_1)|,$ we separate $f_a(y)$ into parts. 
        \snew{In particular, by the Taylor expansion of $\exp(x)$}, we have that
        \begin{align*}
            f_a(y) &= \exp\left(-\frac{m}{n}y + a\log (1+y)\right)+\exp\left(\frac{m}{n}y + a\log (1-y)\right)\\
            & = 2 + a\log (1-y^2)  + \sum_{i=2}^{\infty}\frac{1}{i!}\left(\left(-\frac{m}{n}y + a\log (1+y)\right)^i + \left(\frac{m}{n}y + a\log (1-y)\right)^i\right).
        \end{align*}
         \snew{Define} $g_a(y):= a\log (1-y^2)$ and $h_a(y):= \sum_{i=2}^{\infty}\frac{1}{i!}\left(-\frac{m}{n}y + a\log (1+y)\right)^i$. \snew{Then it follows that} $f_a(y) = 2+ g_x(y) + h_a(y) + h_a(-y).$ 

         On one hand, %
         \begin{align}
         |g_a(\epsilon_0)-g_a(\epsilon_1)|&= \left|a\log\left(\frac{1-\epsilon_0^2}{1-\epsilon_1^2}\right)\right|\le a\left|1-\frac{1-\epsilon_0^2}{1-\epsilon_1^2}\right|\nonumber
         \tag{for $x\ge 1, |\log (x)|\le x-1$}\\
         &=\frac{a}{1-\epsilon_1^2}(\epsilon_0+\epsilon_1) |\epsilon_0-\epsilon_1|\nonumber\\
         &=O\left(\frac{m}{n}\epsilon \delta \log n\right).
         \label{geps_bd}
         \end{align}
         On the other hand, 
         \begin{align*}
             \frac{\partial}{\partial y} h_a(y) &= \left(-\frac{m}{n} + \frac{a}{1+y}\right) \sum_{i=2}^{\infty} \frac{1}{(i-1)!}\left(-\frac{m}{n}y + a \log (1+y)\right)^{i-1}\\
             & = \left(-\frac{m}{n} + \frac{a}{1+y}\right) \left(\exp\left(-\frac{m}{n}y\right)(1+y)^x -1 \right).
         \end{align*}
         \snew{When} $y\in [\epsilon_0, \epsilon_1]$, %
         \begin{align*}
             \left|-\frac{m}{n} + \frac{a}{1+y}\right| &\le \left|-\frac{m}{n}+\frac{1}{1+y} \left(\frac{m}{n}(1-\epsilon_0-\delta)+c\log n\sqrt{m/n}\right)\right|\\
             & \le \left|\frac{-y-\epsilon_0-\delta}{1+y}\frac{m}{n}\right|+ \frac{c}{1+y}\log n \sqrt{m/n} \\
             &= O(1)\left(\frac{m}{n}\epsilon_0 +\log n\sqrt{m/n}\right )
             = O\lp(\log n \sqrt{m/n}\rp),
             \end{align*}
             where the last equality follows from the fact that $\sqrt{\frac{m}{n}}=o\left(\frac{\log n}{\epsilon }\right) $ implies that $ \log n \sqrt{m/n}\gg \frac{m}{n}\epsilon$.
             For $\left|\exp\left(-\frac{m}{n}y\right)(1+y)^a-1\right|$, from Claim \ref{exp()(1+x)nondec}, since $a>m/n(1+y)$, $\exp\left(-\frac{my}{n}\right)(1+y)^a$ nondecreasing and takes value 1 when $y=0$. Therefore, we can remove absolute value directly, \snew{which gives that}
             \begin{align*}
                 \left|\exp\left(-\frac{m}{n}y\right)(1+y)^a-1\right| &\le \left(\sum_{i=0}^{\infty}\frac{y^i}{i!}\right)^{-m/n}(1+y)^a-1\le (1+y)^{-m/n}(1+y)^x-1\\
                 &\le 2(1+y)^{\log n\sqrt{m/n}}-1\le  \sum_{k=1}^{\lceil\log n\sqrt{m/n}\rceil}\binom{\lceil\log n\sqrt{m/n}\rceil}{k} y^k \\
                 &\le O(1)\sum_{k=1}^{\lceil\log n\sqrt{m/n}\rceil}\left(\frac{\epsilon e\log n\sqrt{m/n}}{k}\right)^k.
             \end{align*} 
             To show that $O\left(\epsilon \log n \sqrt{m/n}\right)$ dominates the last term, it suffices to show that $\frac{\epsilon e\log n\sqrt{m/n}}{k} = o(1)$ i.e. $\epsilon = o\left(\frac{1}{\log n \sqrt{m/n}}\right) $, which is equivalent to showing that $ \sqrt{m/n} = o\left(\frac{1}{\log n \epsilon}\right) .$ This is true if and only if $ m = o\left(\frac{n}{\epsilon^2\log^2n }\right)$ as assumed in the premise. 
             
             Therefore, we have that $\max_{y\in [\epsilon_0,\epsilon_1]}\left|\frac{\partial}{\partial y} h_a(y)\right| = O\left(\epsilon \log^2 n \frac{m}{n}\right).$ By the mean value theorem, 
             \snew{we have that}
             \begin{equation}|h_a(\epsilon_0) - h_a(\epsilon_1)|= O\left(\epsilon\delta \log^2 n \frac{m}{n}\right), 
             \label{heps_bd}
             \end{equation}
             \snew{and}
            \begin{equation}|h_a(-\epsilon_0) - h_a(-\epsilon_1)|= O\left(\epsilon\delta \log^2 n \frac{m}{n}\right) .
            \label{h-eps_bd}
            \end{equation}
            Combining \Cref{geps_bd},\Cref{heps_bd},and \Cref{h-eps_bd}, we have that 
            \begin{equation}
                |f_a(\epsilon_0) - f_a(\epsilon_1)| = O\left(\epsilon\delta \log^2 n \frac{m}{n}\right).
                \label{num_ub_unif}
            \end{equation}
            We now consider bounding from below the denominator $f_a(\epsilon_0) + f_a(\epsilon_1)$. Recall that from \Cref{fexpform}, \snew{we have that}
            \begin{align*}
                f_a(y)&= \exp\left(-\frac{my}{n}\right)(1+y)^a + \exp\left(\frac{my}{n}\right)(1-y)^a.
            \end{align*}
            Since $1+x\ge \exp(x-x^2)$, we have that $(1+y)^a \ge \exp(ay-ay^2)$. \snew{This implies that}
            \begin{align*}
                f_a(\epsilon_0) + f_a(\epsilon_1)&\ge \Omega(1)\exp\left((\epsilon+\delta)\left(-\frac{m}{n} +a\right) -a(\epsilon+\delta)^2\right)\\
                &=\Omega(1)\exp\left(-\frac{m}{n}(\epsilon+\delta)^2 -(\epsilon+\delta)c\log n\sqrt{m/n}-a(\epsilon+\delta)^2\right) \\
                &= \Omega(1)\exp\left(-\frac{m}{n}\epsilon^2-\epsilon \log n\sqrt{m/n}-\epsilon^2\log n \sqrt{m/n}\right).
            \end{align*}
            Since $\frac{m}{n} = o\left(\frac{1}{\epsilon^2\log^2n }\right),$ we have that 
            \begin{equation}
                f_a(\epsilon_0) + f_a(\epsilon_1)\ge\Omega(1) \frac{1}{\exp(o(1))} = \Omega(1).
                \label{denom_lb_unif}
            \end{equation}
             Hence, by \Cref{num_ub_unif} and \Cref{denom_lb_unif},
        \begin{align*}
            \bar{I} &= O(1)\sum_{a=0}^{\infty}\frac{1}{a!}\left(\frac{m}{n}\right)^a \exp\left(-\frac{m}{n}\right) \frac{(f_a(\epsilon_0)-f_a(\epsilon_0+\delta))^2}{f_a(\epsilon_0)+f_a(\epsilon_0+\delta)}\\
            &= O\left(\epsilon^2\delta^2 \log^4 n \frac{m^2}{n^2}\right)\sum_{\substack{a\ge \lfloor\lambda + c\log n \sqrt{m/n}\rfloor\\ a\le \lceil\lambda - c\log n \sqrt{m/n}\rceil\\ a\in \Z}}\frac{1}{a!}\left(\frac{m}{n}\right)^a \exp\left(-\frac{m}{n}\right) + o\left(\frac{1}{n}\right).
            \end{align*}
            Since 
            $\sum_{a =\lfloor\lambda + c\log n \sqrt{m/n}\rfloor}^{\lceil\lambda - c\log n \sqrt{m/n}\rceil}\frac{1}{a!}\left(\frac{m}{n}\right)^a <\sum_{a=0}^{\infty}\frac{1}{a!}\left(\frac{m}{n}\right)^a=\exp(m/n)$, this term cancels out with the 
            \snew{succeeding}
            $\exp(-m/n)$ term.
            Thus
            \[
            \bar{I}\le O\left(\epsilon^2\delta^2 \log^4 n \frac{m^2}{n^2}\right)+o\left(\frac{1}{n}\right)
            \]
            as desired.

    The above 2 cases conclude the proof of \Cref{MI_unif}.
\end{proof}

The second step of the framework is as follows.

\begin{lemma}[Lipschitzness of Expected  Acceptance Probability]
Assume that $m = \tilde{o}(\sqrt{n}\epsilon^{-2}\rho^{-1})$.
Let $\mathcal{A}$ be a deterministic tester that takes a sample-count vector over $[n]$ as input.
Let $\epsilon_0<\epsilon_1\in[0,\epsilon]$ be such that $\epsilon_0-\epsilon_1\le\epsilon \rho$. Then it holds that
\[
|\Ep_{ \distribution \sim \mathcal M_{\epsilon_0} }[ \text{Acc}_m(\distribution,\mathcal{A})]-\Ep_{ \distribution \sim \mathcal M_{\epsilon_1} }[ \text{Acc}_m(\distribution,\mathcal{A})]|<0.1 \, ,
\]
\label{lem:lipchi_apf}
\snew{where $\text{Acc}_m$ is the acceptance probability function defined as $\text{Acc}_m(\distribution,\mathcal{A}):=\Pr_{T\sim \PoiS(m,\distribution)}[\mathcal{A}(T)=\text{Accept}].$ }
\end{lemma}

\begin{proof}[Proof of Lemma \ref{lem:lipchi_apf}]
    \snew{Assume for the sake of contradiction that}
    $|\Ep_{ \distribution \sim \mathcal M_{\epsilon_0} }[ \text{Acc}_m(\distribution,\mathcal{A})]-\Ep_{ \distribution \sim \mathcal M_{\epsilon_1} }[ \text{Acc}_m(\distribution,\mathcal{A})]|\ge 0.1$. 
    Let $X$ be an unbiased random bit, and $Y$ be the random variable defined as follows: let $\distribution\sim \mathcal{M}_{\epsilon_X}$, $T\sim \PoiS(m,\distribution)$, then $Y=1$ if $\mathcal{A}(T)$ accept, $Y=0$ otherwise. \snew{It follows from the definition and the assumption}
    that $\Pr[Y=1|X=0]=\Ep_{ \distribution \sim \mathcal M_{\epsilon_0} }[ \text{Acc}_m(\distribution,\mathcal{A})]$ and $\Pr[Y=1|X=1]=\Ep_{ \distribution \sim \mathcal M_{\epsilon_1} }[ \text{Acc}_m(\distribution,\mathcal{A})]$,
    \snew{which implies a mutual information bound of $I(X:T) \geq I(X:Y) = \Omega(1)$.}
    This clearly contradicts the result from Lemma \ref{MI_unif}, and hence concludes the proof of \Cref{lem:lipchi_apf}.
\end{proof}

We are ready to show the main result of this subsection.

\begin{proof}[Proof of Lemma \ref{lem:avg-acceptance-prob-unif}]
    Since $\mathcal{A}$ is $0.1$-correct w.r.t. $\mathcal{M}_{0}$ and $\mathcal{M}_{\eps}$, we have that $\Ep_{ \distribution \sim \mathcal M_{0} }[ \text{Acc}_m(\distribution,\mathcal{A})]\ge 0.9$ and $\Ep_{ \distribution \sim \mathcal M_{\epsilon} }[ \text{Acc}_m(\distribution,\mathcal{A})]<0.1$. Furthermore, since $\Ep_{ \distribution \sim \mathcal M_{\xi} }[ \text{Acc}_m(\distribution,\mathcal{A})]$ is a polynomial in $\xi$, it is continuous in $\xi$. Hence, by \snew{the} mean value theorem, there exists $\xi^*\in (\0,\epsilon)$ such that $\Ep_{ \distribution \sim \mathcal M_{\xi^*} }[ \text{Acc}_m(\distribution,\mathcal{A})]=1/2.$ It follows immediately from \ref{lem:lipchi_apf} that $\forall \xi \in [\xi^*-\rho \epsilon,\xi^*+\rho \epsilon ]$ we have that 
    \begin{align*}
    \Ep_{ \distribution \sim \mathcal M_{\xi} }[ \text{Acc}_m(\distribution,\mathcal{A})]&\in \left(\Ep_{ \distribution \sim \mathcal M_{\xi^*} }[ \text{Acc}_m(\distribution,\mathcal{A})]-0.1, \Ep_{ \distribution \sim \mathcal M_{\xi^*} }[ \text{Acc}_m(\distribution,\mathcal{A})]+0.1\right)\\
    &= (0.4,0.6)\subset (1/3,2/3).
    \end{align*}
    In conclusion, if we uniformly at randomly select a $\xi\in[0,\epsilon]$, then once it falls in interval $[\xi^*-\rho \epsilon,\xi^*+\rho \epsilon ]$ of length $2\rho\epsilon$, \snew{which happens with probability $2 \rho \eps$,}
    we have that $\Ep_{ \distribution \sim \mathcal M_{\xi} }[ \text{Acc}_m(\distribution,\mathcal{A})]\in  (1/3,2/3)$ as desired.
\end{proof}

\section{Proof of \texorpdfstring{\Cref{lem:avg-acceptance-prob-closeness}}{}}

\label{app:pf-lem:avg-acceptance-prob-closeness}
The argument again uses information theory and is similar to the proof of \Cref{lem:avg-acceptance-prob-unif}. 
We follow a framework similar to that in \Cref{app:pf-lem:avg-acceptance-prob-unif}, where the main difference is that we work with a different hard instance. We therefore present the needed lemmas without restating the outline. 
\begin{lemma}[Mutual Information Bound for Closeness Testing Hard Instance]
    Let $m<n/2$, $\epsilon_0<\epsilon_1\in[0,\epsilon] $ be such that $\epsilon_1-\epsilon_0 < \epsilon \rho$, $X$ be an unbiased random bit, $(\tilde{\p},\tilde{\q})\sim \mathcal{N}_{\epsilon_X}$ be defined as in \Cref{def:closeness-hard-instance}, $T\sim \PoiS(m,\tilde{\p}\oplus\tilde{\q})$ as in Proposition \ref{prop:closeness-test-poisson-lower-bound} where $\left(T^{\tilde{\p}}_1,T^{\tilde{\p}}_2,\cdots,T^{\tilde{\p}}_n,T^{\tilde{\q}}_1,T^{\tilde{\q}}_2,\cdots,T^{\tilde{\q}}_n\right)=T\in \R^{2n}$ where $T^{\tilde{\p}}_i$ counts the occurrences of element $i$ sampled from $\tilde{\p}/|\tilde{\p}\|_1$, $T^{\tilde{\q}}_i$ counts the occurrences of element $i$ sampled from $\tilde{\q}/|\tilde{\q}\|_1$. Then
    \[
        I\left(X:T^{\tilde{\p}}_1,\cdots,T^{\tilde{\p}}_{n},T^{\tilde{\q}}_1,\cdots, T^{\tilde{\q}}_n \right)=O\left(\frac{m^3}{n^2}\epsilon^4\rho^2\right).
    \]
    \label{MI_close}
\end{lemma}
\begin{proof}[Proof of Lemma \ref{MI_close}]
    Denote $\delta := \epsilon_1-\epsilon_0 = O(\epsilon \rho)$. Since $(T_i,N_i)'s$ are conditionally independent on $X, $ we have that 
    \[
        I\left(X:T^{\tilde{\p}}_1,\cdots,T^{\tilde{\p}}_{n},T^{\tilde{\q}}_1,\cdots, T^{\tilde{\q}}_n \right) \le \sum_{i=1}^nI\left(X:T^{\tilde{\p}}_i,T^{\tilde{\q}}_{i}\right)=:nI\left(X:T^{\tilde{\p}}_1,T^{\tilde{\q}}_{1}\right).
    \] 
    Therefore, it suffices to show that $I\left(X:T^{\tilde{\p}}_1,T^{\tilde{\q}}_1\right)=O\left(\frac{m^3}{n^3}\epsilon^2\delta^2\right).$ We first note that $\Pr\left[T^{\tilde{\p}}_1=a,T^{\tilde{\q}}_1=b|X=0\right]=\Theta(1)\Pr\left[T^{\tilde{\p}}_1=a,T^{\tilde{\q}}_1=b|X=1\right]$ since $\delta = o(\epsilon).$ Therefore, by Claim \ref{MI_asymp} it suffices to show that $\bar{I} := \sum_{a} \frac{\left(\Pr\left[T^{\tilde{\p}}_1=a,T^{\tilde{\q}}_1=b|X=0\right] - \Pr\left[T^{\tilde{\p}}_1=a,T^{\tilde{\q}}_1=b|X=1\right]\right)^2}{\Pr\left[T^{\tilde{\p}}_1=a,T^{\tilde{\q}}_1=b|X=0\right] + \Pr\left[T^{\tilde{\p}}_1=a,T^{\tilde{\q}}_1=b|X=1\right]} =O\left(\frac{m^3}{n^3}\epsilon^2\delta^2\right)$. We next expand $\Pr\left[T^{\tilde{\p}}_1=a,T^{\tilde{\q}}_1=b|X=0\right].$
    \begin{align*}
        &\Pr\left[T^{\tilde{\p}}_1=a,T^{\tilde{\q}}_1=b|X=0\right] \\
        &= \frac{m}{n}\frac{1}{a!}(1-\epsilon)^a \exp(-(1-\epsilon))\frac{1}{b!} (1-\epsilon)^b\exp(-(1-\epsilon))\\
        & \quad +\frac{n-m}{2n}\frac{1}{a!}\left(\frac{m(2\epsilon+\epsilon_0)}{2(n-m)}\right)^a\exp\left(-\frac{m(2\epsilon + \epsilon_0)}{2(n-m)}\right)\frac{1}{b!}\left(\frac{m(2\epsilon-\epsilon_0)}{2(n-m)}\right)^b\exp\left(-\frac{m(2\epsilon - \epsilon_0)}{2(n-m)}\right)\\
        & \quad +\frac{n-m}{2n}\frac{1}{b!}\left(\frac{m(2\epsilon+\epsilon_0)}{2(n-m)}\right)^b\exp\left(-\frac{m(2\epsilon + \epsilon_0)}{2(n-m)}\right)\frac{1}{a!}\left(\frac{m(2\epsilon-\epsilon_0)}{2(n-m)}\right)^a\exp\left(-\frac{m(2\epsilon - \epsilon_0)}{2(n-m)}\right)\\
        & = \frac{1}{a!b!}\left(\frac{m}{n}(1-\epsilon)^{a+b}\exp(-2(1-\epsilon)) + \frac{n-m}{2n}\left(\frac{m(2\epsilon+\epsilon_0)}{2(n-m)}\right)^a\left(\frac{m(2\epsilon-\epsilon_0)}{2(n-m)}\right)^b\exp\left(\frac{2m\epsilon}{n-m}\right)\right.\\
        &\qquad\qquad  \left.+  \frac{n-m}{2n}\left(\frac{m(2\epsilon+\epsilon_0)}{2(n-m)}\right)^b\left(\frac{m(2\epsilon-\epsilon_0)}{2(n-m)}\right)^a\exp\left(\frac{2m\epsilon}{n-m}\right)\right)=:f_{a,b}(\epsilon_0) \;.
    \end{align*}
    Then \snew{it follows that} 
    \[
        \bar{I}=O(1)\sum_{a,b\in \Z_{\ge 0}}\frac{1}{a!b!}\frac{(f_{a,b}(\epsilon_0) - f_{a,b}(\epsilon_1))^2}{f_{a,b}(\epsilon_0)+ f_{a,b}(\epsilon_1)}=:O(1)\sum_{a,b\in \Z_{\ge 0}}\bar{I}_{a,b} \;,
    \]
    where we denote $\bar{I}_{a,b}:=\frac{1}{a!b!}\frac{(f_{a,b}(\epsilon_0) - f_{a,b}(\epsilon_1))^2}{f_{a,b}(\epsilon_0)+ f_{a,b}(\epsilon_1)}$. To bound $\bar{I}_{a,b},$ we break into three cases regarding the value of $a$.
    
    When $a+b = 0,$ viz. $a=b=0$, $\Pr\left[T^{\tilde{\p}}_1=0,T^{\tilde{\q}}_1=0|X=0\right] =\Pr\left[T^{\tilde{\p}}_1=0,T^{\tilde{\q}}_1=0|X=1\right]= \frac{m}{n}\exp(-2(1-\epsilon)) + \frac{n-m}{n}\exp\left(\frac{2m\epsilon}{n-m}\right)$. Therefore, $ \bar{I}_{0,0} =0.$
    
    When $a+b =1,$ without loss of generality $a=0, b=1$ (by the symmetry between $a$ and $b$). We have that 
        \begin{align*}
            &\Pr\left[T^{\tilde{\p}}_1=0,T^{\tilde{\q}}_1=1|X=0\right]= \Pr\left[T^{\tilde{\p}}_1=0,T^{\tilde{\q}}_1=1|X=1\right]=\\
            &=\frac{m}{n}(1-\epsilon)\exp(-2(1-\epsilon)) + \frac{n-m}{2n}\left(\frac{m(2\epsilon-\epsilon_0)}{2(n-m)}\right)\exp\left(\frac{2m\epsilon}{n-m}\right)\\
            &\quad+ \frac{n-m}{2n}\left(\frac{m(2\epsilon+\epsilon_0)}{2(n-m)}\right)\exp\left(\frac{2m\epsilon}{n-m}\right)\\
            &= \frac{m}{n}(1-\epsilon)\exp(-2(1-\epsilon)) + \frac{\epsilon m}{n}\exp\left(-\frac{2m\epsilon}{n-m}\right) \;.
        \end{align*} 
        Thus, $\bar{I}_{0,1}=  \bar{I}_{1,0} =0.$
        
        When $a+b > 1$,  we can assume without loss of generality $a\le b$. Then the denominator term satisfies $f_{a,b}(\epsilon_0)+f_{a,b}(\epsilon_1) = \Omega\left(\frac{m}{n} (1-\epsilon)^{a+b}\right)$. On the other hand, consider the numerator. 
        When $y\in [\epsilon_0 , \epsilon_1],$
        \begin{align*}
            &\left|\frac{\partial}{\partial y} f_{a,b}(y)\right|= \frac{n-m}{2n}\exp\left(-\frac{2m\epsilon}{n-m}\right)\cdot \\
            &\left|\frac{m}{2(n-m)}\left(a\left(\frac{m(2\epsilon + y)}{2(n-m)}\right)^{a-1}\left(\frac{m(2\epsilon - y)}{2(n-m)}\right)^{b} -b \left(\frac{m(2\epsilon + y)}{2(n-m)}\right)^{a}\left(\frac{m(2\epsilon - y)}{2(n-m)}\right)^{b-1} \right)\right.\\
            & \left.+ \frac{m}{2(n-m)}\left(b\left(\frac{m(2\epsilon + y)}{2(n-m)}\right)^{b-1}\left(\frac{m(2\epsilon - y)}{2(n-m)}\right)^{a} -a \left(\frac{m(2\epsilon + y)}{2(n-m)}\right)^{b}\left(\frac{m(2\epsilon - y)}{2(n-m)}\right)^{a-1}\right) \right|\\
            &= \frac{m}{n}\frac{m}{2(n-m)}^{a+b-1}\left|(2\epsilon + y)^{a-1} (2\epsilon-y)^{b-1}[2\epsilon(a-b)-y(a+b)]\right. \\
            &\qquad\qquad\qquad\qquad\qquad+\left.(2\epsilon + y)^{b-1} (2\epsilon-y)^{a-1}[2\epsilon(b-a)-y(a+b)]\right|.\\
            & \text{(Let $c>1$ be such that $n-m= \frac{1}{c}n$.) }\\
            &= O(c^{a+b}) \left(\frac{m}{n}\right)^{a+b}\left[2\epsilon(b-a)\left|(2\epsilon + y)^{b-1} (2\epsilon-y)^{a-1}- (2\epsilon + y)^{a-1} (2\epsilon-y)^{b-1} \right|\right.\\
            &\qquad\qquad\qquad\qquad\qquad \left.+ y(a+b)\left|(2\epsilon + y)^{b-1} (2\epsilon-y)^{a-1}+ (2\epsilon + y)^{a-1} (2\epsilon-y)^{b-1} \right|\right]\\
            &\le O(c^{a+b})\left(\frac{m}{n}\right)^{a+b} \epsilon (a+b) \left[(2\epsilon + y)^{b-1} (2\epsilon-y)^{a-1}+ (2\epsilon + y)^{a-1} (2\epsilon-y)^{b-1} \right]\\
            &=O\left((4c)^{a+b}\left(\frac{m}{n}\right)^{a+b}\epsilon^{a+b-1}(a+b)\right).
        \end{align*}
        This implies that $\forall a+b \ge 2, \bar{I}_{a,b}= O\left(\frac{1}{a!b!}\left(\frac{(4c)^2}{1-\epsilon}\right)^{a+b}\left(\frac{m}{n}\right)^{2a+2b-1}\epsilon^{2(a+b-1)}(a+b)^2\delta^2\right)$. Then 
        \begin{align*}
            \sum_{\substack{a,b\in \Z_{\ge 0}\\a+b\ge 2\\a\le b}} \bar{I}_{a,b} &= O\left(\sum_{a=0}^{\infty}\frac{1}{a!}\left(\frac{(4c)^2}{1-\epsilon}\right)^a\sum_{\substack{b\ge a\\ b\ge 2-a}}\frac{1}{b!}\left(\frac{(4c)^2}{1-\epsilon}\right)^{b}\left(\frac{m}{n}\right)^{2a+2b-1}\epsilon^{2(a+b-1)}(a+b)^2\delta^2\right) \;,
        \end{align*}
        where $\sum_{\substack{b\ge a\\ b\ge 2-a}}\frac{1}{b!}\left(\frac{(4c)^2}{1-\epsilon}\right)^{b}\le \exp((4c)^2/(1-\epsilon))$ and the sum of infinite geometric series is dominated by the first term when common ratio $<1$, we have that
        \begin{align*}
            \sum_{\substack{a,b\in \Z_{\ge 0}\\a+b\ge 2\\a\le b}} \bar{I}_{a,b} &= O\left(\sum_{a=0}^{\infty}\frac{1}{a!}\left(\frac{(4c)^2}{1-\epsilon}\right)^{a} \left(\frac{m}{n}\right)^3 \epsilon^2\delta^2\right)= O\left(\left(\frac{m}{n}\right)^3 \epsilon^2\delta^2\right) = \sum_{\substack{a,b\in \Z_{\ge 0}\\a+b\ge 2\\a\ge b}} \bar{I}_{a,b}
        \end{align*}
        as desired. 
        
    In conclusion, from the above three cases, we have that 
    \[
    \bar{I} \le \bar{I}_{0,0}+ \bar{I}_{0,1}+ \bar{I}_{1,0}+ \sum_{\substack{a,b\in \Z_{\ge 0}\\a+b\ge 2\\a\ge b}} \bar{I}_{a,b} + \sum_{\substack{a,b\in \Z_{\ge 0}\\a+b\ge 2\\a\le b}} \bar{I}_{a,b} = O\left(\left(\frac{m}{n}\right)^3 \epsilon^2\delta^2\right),
    \]
    which concludes the proof of \Cref{MI_close}.
\end{proof}

\begin{lemma}[Lipchitzness of Expected Acceptance Probability]
Assume that $m = o(n^{2/3} \eps^{-4/3} \rho^{-2/3} + n )$, and $\mathcal{A}$ be a deterministic tester takes a sample-count vector over $[n]$ as input and returns 1 if accept 0 otherwise, and recall the acceptance probability function is defined via $\text{Acc}_m( \p , \q, \innerAlg)
:= \Pr_{ T \sim \PoiS(m, \p \oplus \q) } \lp[ \innerAlg(T) =\text{Accept}\rp].$ Let $\epsilon_0<\epsilon_1\in[0,\epsilon]$ be such that $\epsilon_1-\epsilon_0\le\epsilon \rho$. Then it holds that
\[
|\Ep_{ (\p,\q) \sim \mathcal{N}_{\epsilon_0} }[ \text{Acc}_m( \p , \q, \innerAlg)]-\Ep_{ (\p,\q)) \sim \mathcal{N}_{\epsilon_1} }[ \text{Acc}_m( \p , \q, \innerAlg)]|<0.1 \;.
\]
\label{lem:lipchi_apf_close}
\end{lemma}

\begin{proof}[Proof of Lemma \ref{lem:lipchi_apf_close}]
    Assume the opposite $|\Ep_{ (\p,\q) \sim \mathcal{N}_{\epsilon_0} }[ \text{Acc}_m( \p , \q, \innerAlg)]-\Ep_{ (\p,\q)) \sim \mathcal{N}_{\epsilon_1} }[ \text{Acc}_m( \p , \q, \innerAlg)]|\ge 0.1$. Then, let $X$ be an unbiased random bit, and $Y$ be the random variable defined as follows: let $(\p,\q)\sim \mathcal{N}_{\epsilon_X}$, $T \sim \PoiS(m, \p \oplus \q)$, then $Y=1$ if $\mathcal{A}(T)$ accept, $Y=0$ otherwise. From the definition, we notice that $\Pr[Y=1|X=0]=\Ep_{ (\p,\q) \sim \mathcal{N}_{\epsilon_0} }[ \text{Acc}_m( \p , \q, \innerAlg)]$ and $\Pr[Y=1|X=1]=\Ep_{ (\p,\q) \sim \mathcal{N}_{\epsilon_1} }[ \text{Acc}_m( \p , \q, \innerAlg)]$, which implies a mutual information bound of $I(X:T)=I(X:Y) = \Omega(1).$ This contradicts with the result from Lemma \ref{MI_close}.
\end{proof}

We are now ready to prove \Cref{lem:avg-acceptance-prob-closeness}.

\begin{proof}[Proof of Lemma \ref{lem:avg-acceptance-prob-closeness}]
    Since $\mathcal{A}$ is $0.1$-correct w.r.t. $\mathcal{N}_{0}$ and $\mathcal{N}_{\eps}$, we have that $\Ep_{ (\p,\q) \sim \mathcal{N}_{0} }[ \text{Acc}_m( \p , \q, \innerAlg)]\ge 0.9$ and $\Ep_{ (\p,\q) \sim \mathcal{N}_{\epsilon} }[ \text{Acc}_m( \p , \q, \innerAlg)]<0.1$. Furthermore, since $\Ep_{ (\p,\q) \sim \mathcal{N}_{\xi} }[ \text{Acc}_m( \p , \q, \innerAlg)]$ is a polynomial in $\xi$, it is continuous in $\xi$, then by mean value theorem there exists $\xi^*\in (\0,\epsilon)$ such that $\Ep_{ (\p,\q) \sim \mathcal{N}_{\xi^*} }[ \text{Acc}_m( \p , \q, \innerAlg)]=1/2.$ Immediately following from Lemma \ref{lem:lipchi_apf_close} that $\forall \xi \in [\xi^*-\rho \epsilon,\xi^*+\rho \epsilon ]$ we have that 
    \begin{align*}
    \Ep_{ (\p,\q) \sim \mathcal{N}_{\xi} }[ \text{Acc}_m( \p , \q, \innerAlg)]&\in \left(\Ep_{ (\p,\q) \sim \mathcal{N}_{\xi^*} }[ \text{Acc}_m( \p , \q, \innerAlg)]-0.1, \Ep_{ (\p,\q) \sim \mathcal{N}_{\xi^*} }[ \text{Acc}_m( \p , \q, \innerAlg)]+0.1\right)\\
    &= (0.4,0.6)\subset (1/3,2/3).
    \end{align*}
    In conclusion, if we uniformly at random select a $\xi\in[0,\epsilon]$, then once it falls in interval $[\xi^*-\rho \epsilon,\xi^*+\rho \epsilon ]$ of length $2\rho\epsilon$ we have that $\Ep_{ (\p,\q) \sim \mathcal{N}_{\xi} }[ \text{Acc}_m( \p , \q, \innerAlg)]\in  (1/3,2/3)$ as desired.
\end{proof}

Conditioned on some $\xi$ satisfying the probabilistic condition in \Cref{lem:avg-acceptance-prob-closeness}, we then proceed to show that the acceptance probability $\text{Acc}_m( \p , \q, \innerAlg)$ concentrates around the expected acceptance probability $\Ep_{ (\p, \q) \sim \mathcal N_{\xi} }
\lp[ \text{Acc}_m( \p , \q, \innerAlg) \rp]$.

\end{document}